\documentclass{article} 
\usepackage{iclr2027_conference,times}

\usepackage{amsmath,amsfonts,bm}

\def\eqref#1{equation~\ref{#1}}

\def\1{\bm{1}}

\DeclareMathAlphabet{\mathsfit}{\encodingdefault}{\sfdefault}{m}{sl}
\SetMathAlphabet{\mathsfit}{bold}{\encodingdefault}{\sfdefault}{bx}{n}

\usepackage{xcolor}
\definecolor{citec}{HTML}{2a66cc}
\definecolor{refc}{HTML}{2a66cc}
\definecolor{urlc}{HTML}{2a66cc}
\definecolor{enp}{HTML}{0000f0}
\usepackage[colorlinks,
            linkcolor=refc,
            anchorcolor=refc,
            urlcolor=urlc,
            citecolor=citec]{hyperref}
\usepackage{url}
\usepackage{wasysym}
\usepackage{multicol}
\usepackage{latexsym}
\usepackage{booktabs}
\usepackage{enumitem}
\usepackage{amssymb}
\usepackage{amsmath}
\usepackage{subfig}
\usepackage{graphicx}
\usepackage{pifont}
\usepackage{multirow}
\usepackage[ruled,linesnumbered]{algorithm2e}
\usepackage{colortbl}
\usepackage{orcidlink}
\usepackage{rotating}
\usepackage[inkscapelatex=false]{svg}
\usepackage{diagbox}
\usepackage{wrapfig}
\usepackage{mathtools}
\usepackage{amsthm}
\usepackage{tcolorbox}
\usepackage{makecell}
\usepackage{fontawesome5}
\usepackage{longtable}
\usepackage{microtype}

\iclrfinaltrue

\title{Rethinking Contextualization by \\Reinterpreting Attention Head Channels}

\author{Hakaze Cho\orcidlink{0000-0002-7127-1954}${}^{1,2}$\hfill \phantom{1} Haolin Yang\orcidlink{0009-0000-5904-3054}${}^{3}$\phantom{1111}\hfill \phantom{1.11} Zhun Sun$^{2}$\\
\textbf{Naoya Inoue}${}^{4,1}$\hfill \phantom{11.11111}\textbf{Benjamin Heinzerling}${}^{1,2}$\phantom{11111}\hfill \textbf{Kentaro Inui}${}^{1,2,5}$\\
${}^{1}$RIKEN \phantom{1} ${}^{2}$Tohoku University \phantom{1} ${}^{3}$New York University \phantom{1} ${}^{4}$JAIST \phantom{1} ${}^{5}$MBZUAI \\ \faGithub~{TBA} \phantom{11} {\small\texttt{yufeng.zhao@riken.jp}}}

\begin{document}

\maketitle

\vspace{-\baselineskip}
\begin{abstract}
Contextualization, the core operation of language modeling, transmits information across words to build sentence-specific word representations. Prior works mainly study contextualization, focusing on individual words and attention heads as a growing discrete dictionary, lacking a global view of their general behavior. Therefore, we propose a general principle: \textbf{Globally}, we find and estimate that different words carry different amounts of information, and less-informative words tend to absorb more contextual information. \textbf{Specifically}, these low-information words do not absorb contextual words uniformly, and finer-grained selectivity enables more precise routing to promote information transmission between matched words. \textbf{Moreover}, to find what mechanism causes such processing, we reinterpret attention heads as channels gated by their singular vectors and find that: \textbf{(1)} these singular vectors point to the hidden states of more informative words, allowing such words to write their information to others more strongly to act as information sources, and vice versa; and \textbf{(2)} these singular vectors can be viewed equally as hidden state features, enabling automated interpretation of attention heads beyond prior heuristic head discovery, also embedding heads into a continuous space rather than treating them as discrete, independent dictionary entries.
\end{abstract}

\vspace{-\baselineskip}
\section{Introduction}

Contextualization overrides single words' original representations with semantics sourced from context, which is a core operation for computational language modeling~\citep{Peters2018DeepCW, Liu2019LinguisticKA} and human language understanding~\citep{doi:10.1126/science.7350657, Hagoort2004IntegrationOW, Lau2008ACN, Heilbron2020AHO, Broderick2017ElectrophysiologicalCO}. In modern Transformer \textbf{L}anguage \textbf{M}odels (LMs), contextualization is implemented by attention-based calculation, making both words (or tokens) and attention heads central objects for understanding the principle of contextualization.

As to be discussed in~\S\ref{sec:preparation}, previous works have mainly investigated contextualization between specific words~\citep{Abnar2020QuantifyingAF, Clark2019WhatDB, wang2023label, hendel2023context, chen2024sepllm} and specific attention heads~\citep{qiu2025eliciting, olsson2022context, Wang2022InterpretabilityIT, chen2024unveiling, yu2025correcting, Cho2024RevisitingIL}, which is based on heuristic methods with observe-conjecture-verify paradigm, then exhaustive or incomplete enumeration of word or token pairs, and specifying and naming attention heads with their functionality in an ever-growing but discrete dictionary~\citep{zheng2024attention}, making it hard for a concise, general principle, which is a central ideal of scientific spirit~\citep{friedman1974explanation, kitcher1981explanatory, rissanen1978modeling, anderson1972more}.

\begin{wrapfigure}[26]{r}{0.4\textwidth}
    \vspace{-0.4\baselineskip}
    \centering
    \includegraphics[width=0.4\textwidth, trim=0 0 13pt 0, clip]{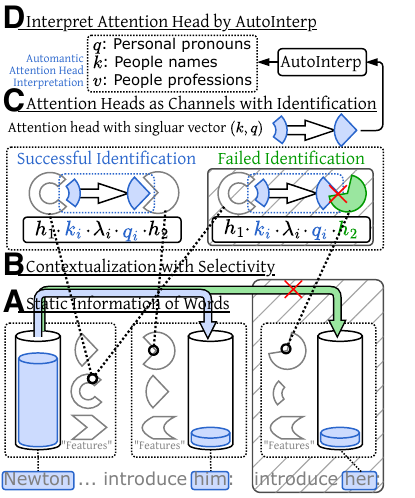}
    \vspace{-1.80\baselineskip}
    \caption{\textbf{(A)} Words initially carry different amounts of static information. \textbf{(B)} During contextualization, information tends to flow from information-rich to information-poor tokens with semantic selectivity. \textbf{(C)} We reinterpret each attention head as an information channel gated by singular vectors.}
    \label{fig:fig1}
\end{wrapfigure}

Therefore, as mentioned in~\S\ref{sec:principle}, in this paper, we re-describe the contextualization process as a general principle (Fig.~\ref{fig:fig1}-B): \textit{Information is sourced from informative words and absorbed into less informative ones with finer-grained selectivity} (as illustrated by the ``Newton'' example in Fig.~\ref{fig:fig1}). Such a principle can be decomposed into 3 basic hypotheses: \textbf{(1) Static Information:} (Fig.~\ref{fig:fig1}-A) words carry various amounts of information before contextualization. \textbf{(2) Orientation by Information Gradient:} (Fig.~\ref{fig:fig1}-B) During contextualization, words with more information tend to serve as stable information sources, and words with less information tend to absorb information from context words. \textbf{(3) Selectivity Overriding Information Gradient:} (Fig.~\ref{fig:fig1}-C) However, information gradients alone cannot explain all contextualization behaviors. For example, ``Newton'' may transmit information more to ``he'' than to ``she'', suggesting that information transmission is selective and effective only between words matched along certain semantic identification.

\newpage

We investigate the principle from two directions. \textbf{(I, \S\ref{sec:phenomenology}) Phenomenology}: we directly observe the three hypotheses from LM's calculations as summarized in Fig.~\ref{fig:method}. In detail: \textbf{(1)} We quantify a word's information by its effect on outputs across a set of semantically related queries, estimating and confirming that information content varies across words. \textbf{(2)} We find that during contextualization, representations of low-information words show greater hidden-state variation across occurrences in different sentences, indicating stronger accumulation of contextual information, whereas high-information words maintain a relatively stable representation. \textbf{(3)} We demonstrate the selectivity of contextualization using some matched token pairs (e.g., ``Newton'' and ``he'') against unmatched token pairs (e.g., ``Newton'' and ``she''), and find that information flows more strongly between the matched ones. 

Also, with the duality of contextualization with the attention mechanism, we provide \textbf{(II, \S\ref{sec:mechanism}) Mechanism} explanation to find what properties of attention head induce the previous principle. As shown in Fig.~\ref{fig:fig1}-C, we view an attention head as an information channel gated by the singular vectors of $W_\text{Q}^\top W_\text{K}$ and $W_\text{O}W_\text{V}$ parameter matrices. Put simply, for information to propagate between two tokens, they must jointly activate a pair (with the same index) of $W_\text{Q}^\top W_\text{K}$'s singular directions and a $W_\text{O}W_\text{V}$'s right singular direction, all with non-negligible singular values. Therefore, these singular vectors effectively characterize the concrete operational modes of an attention head. From this perspective, we make three developments: \textbf{(1) Singular Vectors Point to Directions of Increasing Word Information Amount.} Especially for $W_\text{O}W_\text{V}$'s right singular directions, which control the writing magnitude: hidden states from words with higher information content exhibit larger projections onto these directions, strengthening the writing of their own information into other words. \textbf{(2) Automatic Attention Head Interpretation.} These singular vectors can be regarded as latent feature which are ``listening'' features residing in the hidden states of the corresponding layer, allowing us to apply automatic feature interpretation~\citep{paulo2025automatically} to explain the operations performed by attention heads, which provides a fundamental alternative to the previous heuristic method to discover and interpret attention heads. \textbf{(3) Embed Attention Heads in Vector Space.} Finally, using singular vectors as proxies for attention-head operations yields continuous representations of head functionality, moving beyond previous discrete taxonomies.

\section{Background}
\label{sec:preparation}

\textbf{Word-wise Behavior in Contextualization.} Prior work has investigated contextualization between specific types of words.~\citet{chen2024sepllm} find that separator words collect and compress information from context, and some works scope the investigation to structural tokens in the prompt template~\citep{wang2023label}, or the last-position token~\citep{hendel2023context} in the in-context learning scenario. Also,~\citet{Abnar2020QuantifyingAF} and~\citet{Clark2019WhatDB} investigate the attention scores flowing to the [CLS] or [MASK]. However, the principles governing overall information exchange rather than the aforementioned special words have not yet been systematically investigated.

\textbf{Attention Head-wise Behavior in Contextualization.} Meanwhile, investigating the behavior of individual words almost inevitably links to the behavior of attention heads where the observed behaviors are operationalized~\citep{wu2025retrieval}. Typical works are largely heuristic: these studies typically begin by specifying a set of tasks or an input distribution, then identify attention heads that play critical roles in those tasks through ablation~\citep{ghorbani2020neuron, li2024optimal}, circuit discovery~\citep{sundararajan2017axiomatic, hanna2024have}, or even heuristic inspiration, and finally name these heads according to the task or their behavioral patterns. Works in such a paradigm include induction heads~\citep{olsson2022context, Cho2024RevisitingIL}, operating with prompt template tokens for in-context learning, and repeated token head~\citep{Wang2022InterpretabilityIT} for the repeated words in the inputs. Such works mainly observe input-dependent attention scores with doubtful robustness~\citep{jain2019attention, wiegreffe2019attention, bibal2022attention, bastings-filippova-2020-elephant}, and more critically, finds many types of attention heads and tagged them with discrete names, which are organized into an excessively crude set-style topological structure like classified dictionaries~\citep{zheng2024attention} and therefore block scientific insight (e.g., how different attention heads are related or evolved), and are also fundamentally limited by the polysemy~\citep{gould2024successor} of attention heads.

\section{General Principles of Contextualization}
\label{sec:principle}

\textbf{Why a General Principle?} Science requires summarizing descriptions of individual phenomena with concise principles capturing their macroscopic regularities~\citep{friedman1974explanation, kitcher1981explanatory, rissanen1978modeling} as statistical mechanics~\citep{jaynes1957information}: rather than tracking the dynamics of individual particles, it explains macroscopic or average behavior through collective quantities~\citep{anderson1972more}. Prior work therefore faces the same limitation identified above: case-by-case analyses scale poorly, generalize only weakly beyond the configurations examined, and accumulate into an ever-growing catalogue of token interactions and named attention-head functions rather than a compact theory. 

Analogous to statistical mechanics, instead of enumerating information transmission across innumerable word pairs and heads, we seek a macroscopic principle that captures their shared regularities, as illustrated in Fig.~\ref{fig:fig1}: \textit{Words with more information content tend to be stable information sources, and words with less information tend to be overridden by contextual information; attention heads here serve as gated channels to transmit such information}, which we reword into 3 basic hypotheses:

\begin{enumerate}[topsep=0pt, itemsep=-1pt, leftmargin=15pt]
\item \textbf{Static Information:} As shown in Fig.~\ref{fig:fig1}-A, before any contextualization, words carry various amounts of information, therefore bring various magnitudes of output influence. Naturally, words with more specific semantics carry more information; for example, ``Newton'' refers more directly to a particular entity than the semantically less specific pronoun ``he''.

\item \textbf{Contextualization Oriented by Static Information Gradient:} During contextualization, words with more information tend to keep their information stable, so that they serve as a stronger information source for contextualizing other words, and words with less information tend to absorb information from context words and rewrite their representations.The logic is intuitive: some abstract words, such as ``he'', require more specific contextual semantics potentially supplied through implicit coreference resolution before they can meaningfully contribute to prediction.

\item \textbf{Fine-grained Selectivity:} The two principles above characterize the average behavior of contextualization, while individual word pairs are further modulated by semantic compatibility. For example, when a high-information word such as ``Newton'' co-occurs with lower-information words such as ``he'' and ``she'', it is more likely to transmit information to ``he'' than to ``she''. This fine-grained selectivity connects our unified global principle with the case-by-case phenomena identified by prior empirical analyses, treating those manually characterized interactions as specific instances of a broader rule.
\end{enumerate}

We next validate these principles through LM outputs and hidden states (\S\ref{sec:phenomenology}), and find how parameters of attention heads induce these operations (\S\ref{sec:mechanism}) to provide a new perspective on attention computation.

\section{Experimental Settings}
\label{sec:settings}

\textbf{Models.} We conduct experiments on 10 LMs: \textbf{Encoder-only}: BERT-Base~\citep{devlin2019bert} and XLM-RoBERTa-Large~\citep{conneau2020xlmr}; \textbf{Decoder-only}: Llama 3.2-1B, Llama 3-8B~\citep{grattafiori2024llama}, Llama 2-13B~\citep{touvron2023llama}, Qwen3-8B, Qwen3-14B, Qwen3.6-27B~\citep{yang2025qwen3}, Granite 4.1-30B~\citep{granite2026}, and OLMo 3-32B~\citep{olmo2025olmo}. We show results on Llama 3.2-1B in default, refer to Appendix~\ref{appendix.more_results} for results on other models.

\textbf{Data.} To broadly investigate the contextualization behavior of diverse words, we use GPT-4o-mini to sample around 50 words from each of the 16 categories shown on the vertical axis of Fig.~\ref{fig:static_information}, yielding 800 words in total. Then we sample 150 queries with output candidates about these words for the experiments on Hypothesis 1 (e.g., ``Please judge the gender of [Word]'' with output candidates ``male'' and ``female'', refer to~\S\ref{subsec:static_info}), and 128 templates, each with an empty space to fill the word to be tested in, for the experiments on Hypothesis 2 (Refer to~\S\ref{subsec.h2}). For multi-token words, if necessary, we average the results over all sub-word tokens (dataset details in Appendix~\ref{appendix.dataset}).

\section{Phenomenology: Observing the Contextualization}
\label{sec:phenomenology}

In this section, we directly validate the 3 hypotheses in~\S\ref{sec:principle} by observing the model outputs and hidden states, to provide a direct and empirical confirmation of our proposed principle.

\begin{figure}[t]
    \centering
    \includegraphics[width=\linewidth]{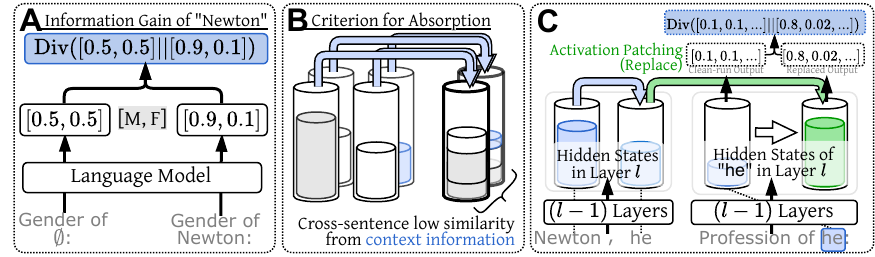}
    \vspace{-1.8\baselineskip}
    \caption{{Methods to characterize contextualization.} \textbf{(A)} Static information is measured by output divergence with the word to be tested input, against the blank background. \textbf{(B)} Contextualization magnitude is measured through hidden-state similarities. \textbf{(C)} Selective transmission is measured by output divergence against the background run and the activation-patched run from contexted inputs.}
    \label{fig:method}
\end{figure}

\subsection{Static Information: Measuring the Information Gain of Tokens}
\label{subsec:static_info}

In this subsection, we mainly validate the first hypothesis, i.e., tokens carry various amounts of information before contextualization, as they bring various magnitudes of influence to the output.

\begin{wrapfigure}[21]{r}{0.5\textwidth}
    \vspace{-1.1\baselineskip}
    \centering
    \includegraphics[width=0.5\textwidth, trim=8.5pt 0 0 0, clip]{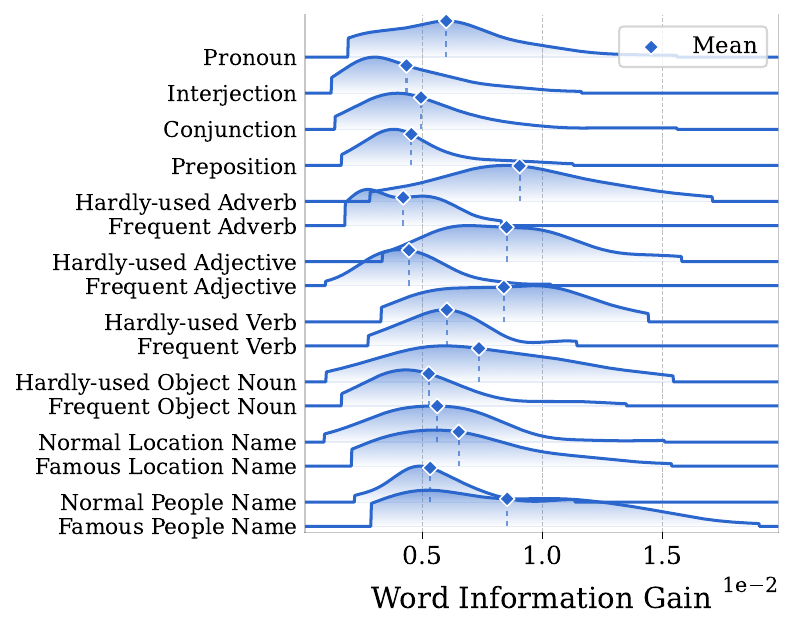}
    \vspace{-2\baselineskip}
    \caption{Static information across word categories. Distributions of word-level information gain for the 16 word categories in Llama 3.2-1B, dots are category means. Words with more specific semantics or lower frequency generally exhibit greater information gain.}
    \label{fig:static_information}
\end{wrapfigure}

\textbf{Experiment Method: Measuring the Information Gain of Words.} \textbf{(Principle)} We can regard a word as ``carrying information'' if and only if that word exerts a non-negligible influence on the model’s output. Therefore, the information carried by a word can be quantified by comparing the model’s output when the word is present against an appropriate blank control condition. \textbf{(Method)} Therefore, as shown in Fig.~\ref{fig:method}-A, we utilize the 150 queries (indexed by $i$) in~\S\ref{sec:settings}, filling all 800 words (indexed by $j$) into the queries, so for each query, we get 800 answer distributions $\{y_{i,j}\}_{j=1}^{800}$ with normalized output probabilities of the output candidates defined by the query. Therefore, we can calculate the information gain of every single ($j$-th) word for one specific ($i$-th) query as: $I_{i,j} = \mathrm{Div}_\text{JS}\left[y_{i,j}\Vert\mathbb{E}_{k=1}^{800}\left[y_{i,k}\right]\right]$, indicating how far the specific word push the output distribution from the averaged distribution (which can be regarded as an blank control condition\footnote{A more direct and simpler implementation can be the divergence of $y_{i,j}$ against output with blank input $y_{i,\emptyset}$, but we use average (``batch calibration~\citep{zhou2024batch}'') here for a better output calibration.}) on one query, so that induces variance to the outputs, i.e., ``exerts its information''. The static information of $j$-th word is then calculated as $\mathbb{E}_{i=1}^{150}[I_{i,j}]$, i.e., the averaged information gain of this token among all the queries (refer to Appendix~\ref{appendix.exp.1} for the details, and the motivation for proposing a new information estimation against the corpus frequency-based previous method~\citep{oyama2023norm}).

\textbf{Static Information of Words.} We show the results on Llama 3.2-1B for words within their categories in Fig.~\ref{fig:static_information}, where we can observe: \textbf{(1)} Different words carry quite different amounts of information, causing varying impacts on the output, which directly confirms our Hypothesis 1. \textbf{(2)} Intuitively, words with more specific semantics (e.g., a famous person's name like ``Newton'' rather than a common name) or words with lower frequencies (see Appendix~\ref{appendix.dataset} for the word frequency) carry more information. Although this aligns with human intuition, we provide empirical evidence for it. We further hypothesize that high-frequency or semantically underspecified words, which carry less information on their own, need to absorb information from context, i.e., contextualized, to acquire more information, so that they can provide enough output contribution; therefore, semantic absorption towards less-informative words may play a prominent role in language modeling. Moreover, in \S\ref{subsec.6.2}, we will show that such information estimation can be captured by the singular vectors of the attention parameters, thereby providing a principled justification and grounding.

\begin{wrapfigure}[18]{r}{0.63\textwidth}
    \vspace{-1.5\baselineskip}
    \centering
    \includegraphics[width=0.63\textwidth]{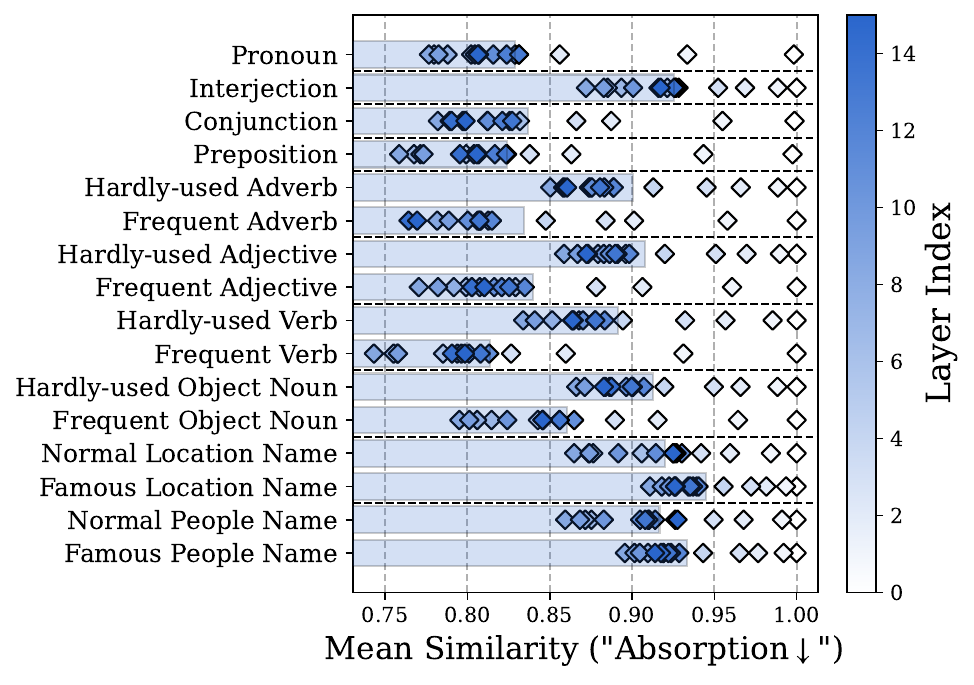}
    \vspace{-2.2\baselineskip}
    \caption{Layer-wise absorption magnitude across word categories. Marker is layer-wise result, bar is mean across layers.}
    \label{fig:donor_and_receptor}
\end{wrapfigure}

\textbf{Effect of Tokenization.} One can doubt that ``famous names'' or ``hardly used'' words may be decomposed into more sub-word tokens, so that potentially\footnote{In fact, there is no clear evidence indicating that a greater number of input tokens necessarily results in a stronger input influence, nor supporting the claim that they yield a detectable increase in information gain.} induce more effect on the feed-forward calculation, showing more information gain. However, as we show in Appendix~\ref{appendix.token_length_control}, when controlling the token length (e.g., by considering only words of 2 tokens), the same conclusions remain robust. This does not rule out a broader role of tokenization: allocating more subword units to certain words may itself benefit language modeling~\citep{gigant2026decoupling, pagnoni-etal-2025-byte, wolleb-etal-2023-assessing}. However, tokenization length alone does not account for the information-gain pattern observed here.


\subsection{Information Transmission Oriented by Static Information Gradient}
\label{subsec.h2}

\begin{wrapfigure}[17]{r}{0.35\textwidth}
    \vspace{-1.2\baselineskip}
    \centering
    \includegraphics[width=0.35\textwidth]{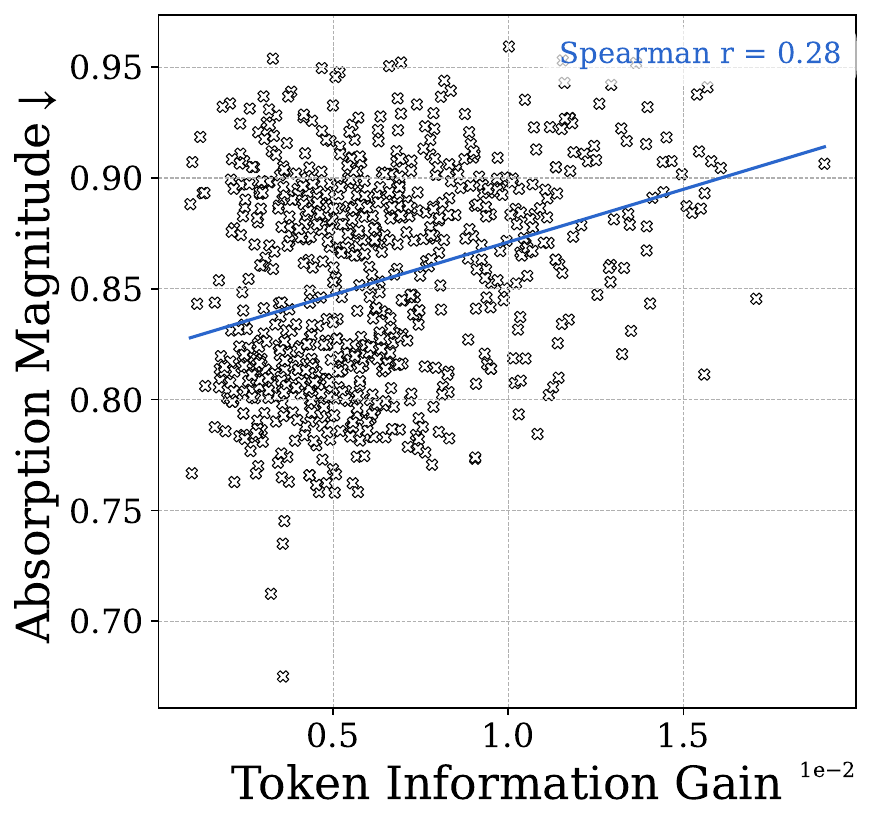}
    \vspace{-1.8\baselineskip}
    \caption{Correlation of static information with information absorption magnitude (negative) in Layer 9 of Llama 3.2-1B. All layer results are in Appendix~\ref{appendix.h2_full_layer}.}
    \label{fig:donor_and_receptor_correlation}
\end{wrapfigure}

Following the intuition that ``words with less information need to be overridden by the context information'', we investigate the LM's hidden states and find that words with less static information tend to absorb more context information.


\textbf{Experiment Method.} \textbf{(Principle)} As shown in Fig.~\ref{fig:method}-B, to investigate how much a word is rewritten by contextual information, we utilize a simple criterion based on the following principle: If a word is absorbing (or being rewritten by) context information, its representation (i.e., LM's hidden states) should vary substantially across occurrences of this same word in various sentences. \textbf{(Method)} Based on the principle, we fill each of the 800 words into all the same 128 templates\footnote{It can be imagined that the 128 templates cannot always be suitable for all the words to form correct sentences, so we repeat this experiment on only grammatically correct instances in Appendix~\ref{appendix.specific_template}.}, yielding 128 sentences per word. We then extract the hidden state of each word from a (each) layer in each sentence and compute the cosine similarity among its 128 contextualized representations as a negative measure of absorption magnitude, i.e., the greater the measurement, the weaker the information absorption.


\textbf{Information-poor Words Tend to Absorb, and Vise Versa.} Following the aforementioned methods, we measure the magnitude of information absorption with averaged results inside each word category shown in Fig.~\ref{fig:donor_and_receptor}, and a more direct correlation with static information amount is shown in Fig.~\ref{fig:donor_and_receptor_correlation} (tokenization-length controlled results in Appendix~\ref{appendix.token_length_control}). In these results, the words with higher information conduct weaker contextual information absorption. Such an observation can be viewed as the general picture of contextualization: words with lower static information absorb more contextual information from other tokens, whereas words with high static information serve as sources\footnote{Here we can only hint at this clue: if low-information words collected contextual information uniformly, high-information words with more stable or purer representations would contribute more. In practice, however, this collection is intrinsically biased toward high-information words, as we will establish in~\S\ref{subsec.6.2}.} carrying the information to be absorbed. The results for other layers are shown in Appendix~\ref{appendix.h2_full_layer}.


\subsection{Selectivity Overriding Information Gradient}

So far, we have shown the global contextualization principles. However, we still need to consider specific cases that deviate from the averaged story, i.e., the absorption is not uniform. For example, as shown in Fig.~\ref{fig:fig1}, when the informative word ``Newton'' co-occurs with less informative words ``he'' and ``she'', to which of the two would the information carried by ``Newton'' be more likely to be transmitted? Intuitively, the information is likely to be transmitted to ``he''. In this section, we confirm such an intuition: some fine-grained selectivity based on the semantic identity modulates the contextualization between two specific words, causing the variance among the global pattern.

    

\textbf{Experiment Method.} To verify the selectivity between two words, as shown in Fig.~\ref{fig:method}-C, to test the magnitude of information modification from word \texttt{A} (e.g., ``Newton'') to word \texttt{B} (e.g., ``he''), \textbf{(1)} we put them in a sequence like ``\texttt{A}, \texttt{B}'' as a simple implementation to induce the contextualization, and save the hidden states of word \texttt{B} here. \textbf{(2)} Then, to test how the information of the contextualized \texttt{B} differs from plain \texttt{B}, we conduct an activation patching~\citep{heimersheim2024use}-styled method as shown in Fig.~\ref{fig:method}-C. In detail, similar to~\S\ref{subsec:static_info}, we save the output distribution of \texttt{B} in the normal setting on many queries, and replace the hidden state of \texttt{B} with the previously saved contextualized one, to produce another set of output distributions. The JS-divergence averaged among queries between both distributions is the contextualization magnitude between \texttt{A} and \texttt{B}.

\begin{figure}[t]
    \centering
    \includegraphics[height=0.293\linewidth]{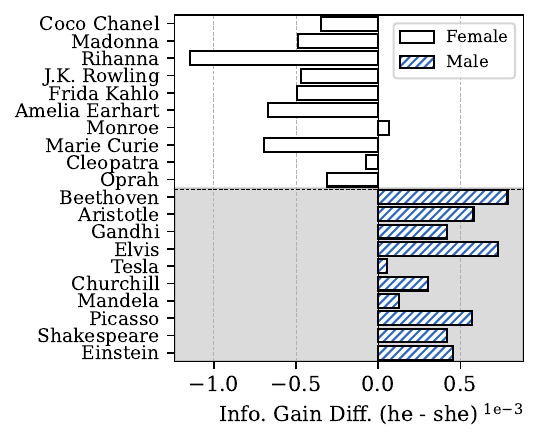}\hfill
    \includegraphics[height=0.293\linewidth]{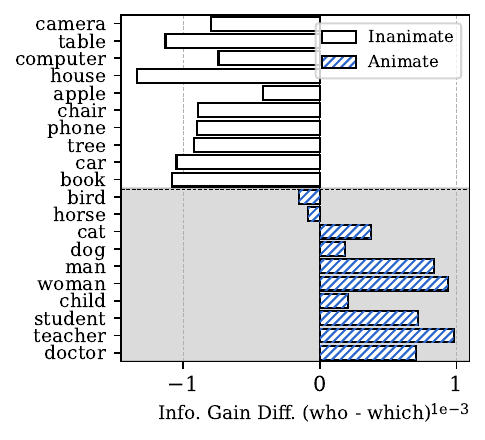}\hfill
    \includegraphics[height=0.293\linewidth]{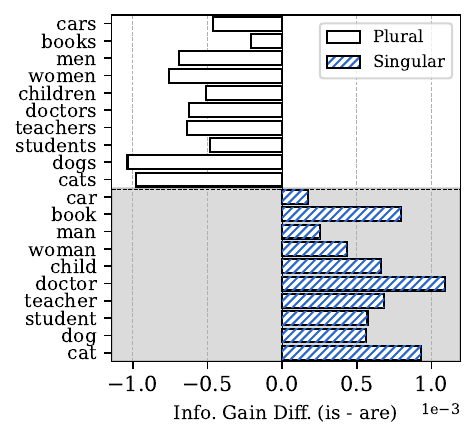}\\ \vspace{-0.4\baselineskip}
    \hfill
    \includegraphics[height=0.303\linewidth]{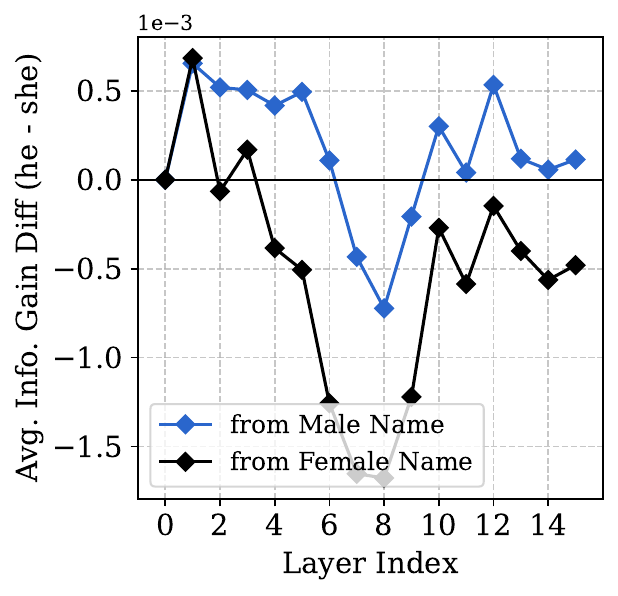}
    \includegraphics[height=0.303\linewidth]{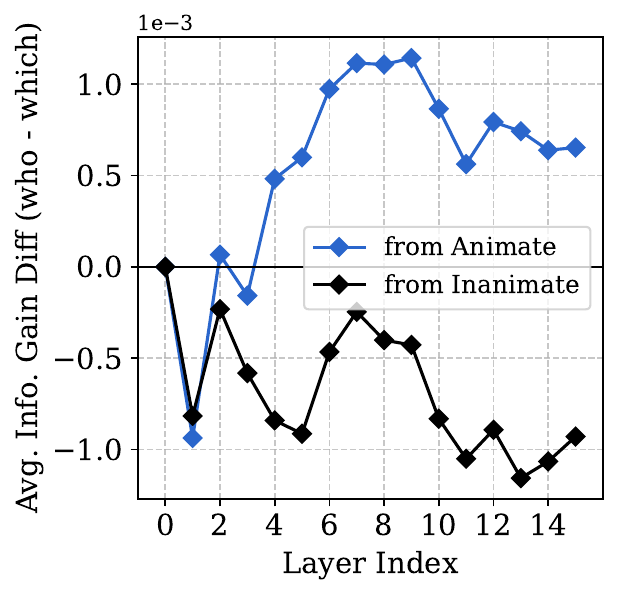}
    \includegraphics[height=0.303\linewidth]{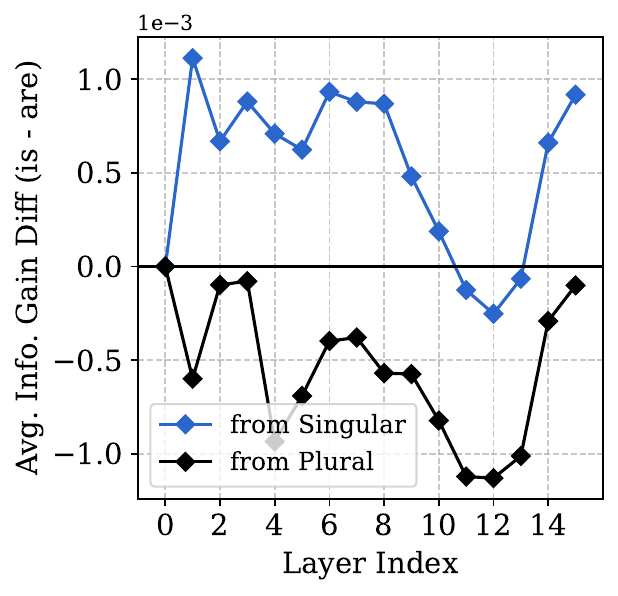}\hfill\\
    \vspace{-1\baselineskip}
    \caption{Contextualization selectivity on (\textbf{Left}) male and female names to ``he'' and ``she'', (\textbf{Middle}) animate and inanimate nouns to ``who'' and ``which'', (\textbf{Right}) plural and singular nouns to ``is'' and ``are''. \textbf{Upper}: Examples on Llama 3.2-1B Layer 5; \textbf{Lower}: Statistics on all layers and similar inputs.}
    \label{fig:selectivity_stat}
    \vspace{-0.5\baselineskip}
\end{figure}

\textbf{Selectivity on Information Transmission.} We choose 3 scenarios to demonstrate such selectivity as shown in Fig.~\ref{fig:selectivity_stat}. For example, with activation patching on layer 5 of Llama 3.2-1B, we can observe that the animate noun (like ``man'', ``student'') tends to modify the representation of ``who'' more than ``which'', and vise versa. Such observations and complete statistics in the lower figures suggest a finer-grained selectivity overriding the global principle, i.e., low-information words do not absorb information from all words uniformly; instead, they perform selectivity focusing on the words with ``identified semantics'' sourced from coreference resolution or semantic / syntactic rules, which reveals additional properties of attention-head computation, we explore in the next section.

\section{Mechanism: Analyzing the Channel of Attention Head}
\label{sec:mechanism}


In this section, to explain the above phenomena, we reinterpret attention heads as information channels controlled by singular vectors of their parameter matrices (Fig.~\ref{fig:fig1}-C). In these channels, information can be transmitted between two tokens when they jointly activate a pair of left and right singular vectors of $W_\text{Q}^\top W_\text{K}$ and a right singular vector of $W_\text{O}W_\text{V}$, with all corresponding singular values sufficiently large. Using this perspective, we \textbf{(1)} provide a deeper mechanistic interpretation for contextualization, \textbf{(2)} enable automatic interpretation of attention heads through AutoInterp~\citep{paulo2025automatically}, and \textbf{(3)} embed discrete attention heads into a continuous space.



\begin{figure}[t]
    \vspace{-1.3\baselineskip}
    \centering
    \includegraphics[width=1\textwidth]{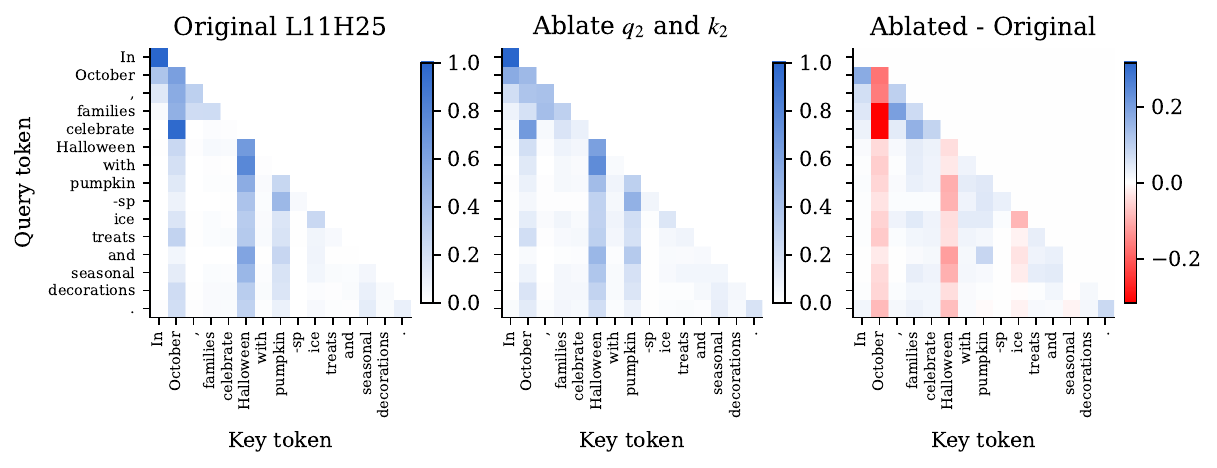}
    \vspace{-2.2\baselineskip}
    \caption{Attention scores (attention sink removed) of one head from Llama 3.2-1B (interpretation in Table~\ref{tab:interp_example}). When we ablate the $q_2$ and $k_2$ features, the attention scores change as interpreted.}
    \label{fig:ablation_head}
    \vspace{-0.7\baselineskip}
\end{figure}

\begin{table}[t]
    \centering
    \caption{{Examples of automatically extracted interpretations by AutoInterp of attention head singular vectors in Llama 3.2-1B.} The F1 score of random prediction is $0.143$.}
    \vspace{-0.8\baselineskip}
    \label{tab:interp_example}
    \resizebox{\linewidth}{!}{
    \begin{tabular}{crlcc}
    \toprule
       \makecell*{\textbf{Attn.}\\ \textbf{Head}}  &  & \textbf{Interpretation} & \makecell*{\textbf{\small Singl.}\\ \textbf{Value}} & \makecell*{\textbf{\small Score}\\ \textbf{(F1)}} \\ \midrule
       \multirow{3}{*}{\makecell*{Layer 2 \\ Head 11}} & $-q_1$ & ``references to specific \textbf{years} or \textbf{numbers} in historical and descriptive contexts'' & \multirow{2}{*}{4.80} & 0.96 \\ 
         & $-k_1$ & ``the substring '\textbf{pre}' and '\textbf{post}' in various legal and procedural contexts'' & & 1.00 \\
         & $-v_1$ & ``phrases indicating the \textbf{start}, effective \textbf{date}, or \textbf{duration} of time-related events or actions'' & 0.48 & 1.00 \\
       \midrule
       
       \multirow{6}{*}{\makecell*{Layer 11 \\ Head 25}} 
       & $q_2$ & ``words related to celebrating cultural or religious \textbf{festivals}, particularly Christmas and similar events'' & \multirow{2}{*}{5.05} & 0.96 \\ 
         & $k_2$ & ``\textbf{months holidays} and \textbf{seasonal events} mentioned in the text'' &  & 1.00 \\
         & $-v_1$ & ``references to \textbf{Christmas} and related holiday concepts'' & 0.76 & 0.93 \\
         & $-v_2$ & ``the concepts of pumpkin spice and \textbf{Halloween celebrations}'' & 0.63 & 0.96 \\ \cmidrule(l){2-5} 

        & $q_1$ & ``the phrase 'from [place]' indicating origins or affiliations with specific locations'' & \multirow{2}{*}{4.87} & 0.85 \\
        & $k_1$ & ``the substring 'Inc', file extensions, and locations indicated with specific abbreviations like NY and TX'' &  & 0.67 \\ 
        \midrule
        
        
        \multirow{3}{*}{\makecell*{Layer 6 \\ Head 29}} & $q_3$ & ``pronouns indicating possession and \textbf{reference to individuals} in various contexts'' & \multirow{2}{*}{4.92} & 0.93 \\
        & $k_3$ & ``the \textbf{word `he'} in various contexts, indicating masculine subjects and their actions or statements'' & & 1.00 \\
        & $v_1$ & ``pronouns and related terms reflecting \textbf{human perspective and reactions}'' & 0.42 & 0.93 \\
    \bottomrule
    \end{tabular}}
    \vspace{-0.7\baselineskip}
\end{table}

\subsection{Multi-head Attention as Gated Information Channel}


\textbf{Revisiting Attention Calculation by Singular Vectors.} By singular vector decomposition~\citep{merullo2024talking}, the self-attention on hidden state $H\in\mathbb{R}^{d\times n}$ can be:
\begin{equation}
\begin{aligned}
    &\text{HeadOutput} = W_\text{O}W_\text{V}H\mathrm{softmax}(H^\top W_\text{Q}^\top W_\text{K}H/d'^{1/2})^\top\text{,} \qquad\text{where} \\ 
    &W_\text{Q}^\top W_\text{K} = \sum\nolimits_{i=1}^{d_h} \lambda_iq_i k_i^\top\text{,} \qquad W_\text{O}W_\text{V} = \sum\nolimits_{i=1}^{d_h} \omega_io_i v_i^\top\text{,}
\end{aligned}
\label{eq:attention}
\end{equation}
where $q_i, k_i, o_i, v_i\in\mathbb{R}^{d}$ are the correstponding left and right singular vectors of $W_\text{Q}^\top W_\text{K}, W_\text{O}W_\text{V}\in\mathbb{R}^{d\times d}$; $\lambda_i, \omega_i\in\mathbb{R}^+_0$ are the singular values. According to Eq.~\ref{eq:attention}, consider the hidden states $h_1$ and $h_2$ produced by two tokens at the current layer, the writing magnitude from $h_2$ to $h_1$ is controlled by two quantities: $\sum_i\lambda_i(q_i^\top h_1)(k_i^\top h_2)$ and $\left\Vert\sum_j\omega_j(v_j^\top h_2)o_j \right\Vert=\sqrt{\sum_j\omega_j^2(v_j^\top h_2)^2}$ (the equality holds since $o_j$s are a set of orthonormal bases). That is, as shown in Fig.~\ref{fig:fig1}-C: $(q_i,k_i)$ pairwise serves as a Key-Lock mechanism, which determines whether a score $\lambda_i$ can be added to the attention score (before $\mathrm{Softmax}$) from key $h_2$ to query $h_1$, so it can then write a vector with norm $w_j\vert v_j^\top h_2\vert$ to direction $o_j$. The $o_j$s are unit vectors that determine the directions in which information is written into $h_1$, but not the magnitude of the written vector. This suggests that the top singular vectors of attention parameters act as gates and descriptors of attention-head behavior.

\subsection{Word Information Can be Self-sniffed by Attention Heads}
\label{subsec.6.2}

\begin{wrapfigure}[16]{r}{0.375\textwidth}
    \vspace{-1.6\baselineskip}
    \centering
    \includegraphics[width=0.375\textwidth]{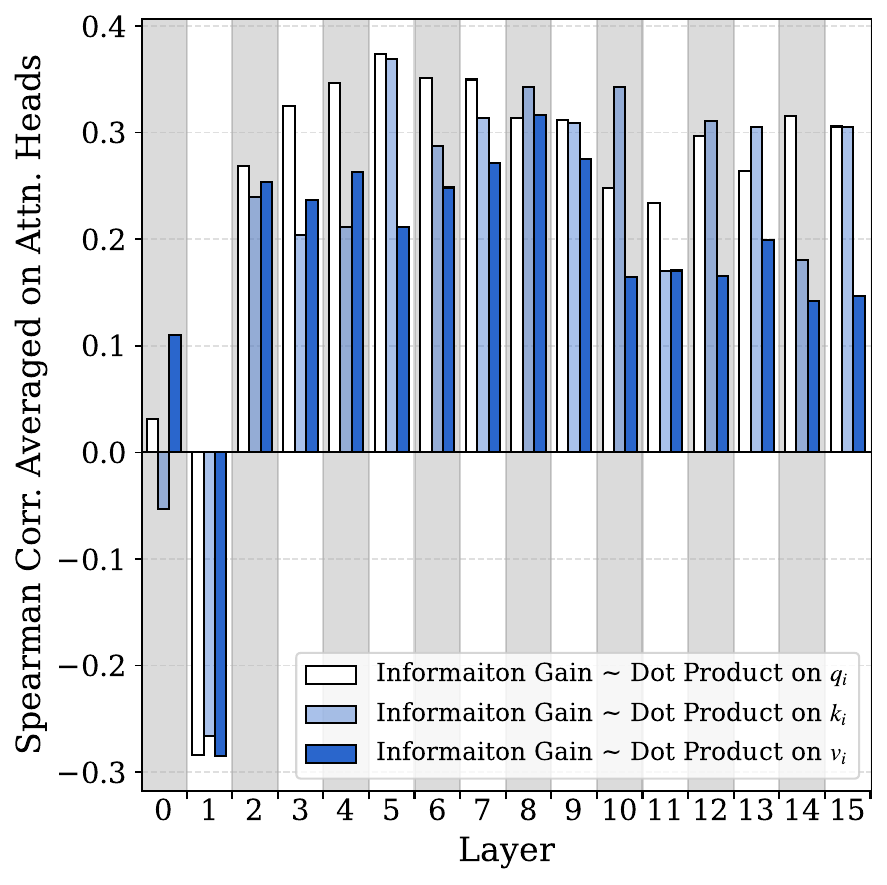}
    \vspace{-2.1\baselineskip}
    \caption{Correlation: length of words' hidden states projected to top $q,k,v$ and words' information.}
    \label{fig:projection}
\end{wrapfigure}

\textbf{Explaining the Contextualization Tendency on Singular Vectors.} The preceding results identify low-information words as the primary recipients of contextual information, but do not directly reveal which words act as the sources of that information. Reinterpreting attention heads as gated information channels provides a way to address this. Because information routing and writing depend on the alignment of token hidden states with the head's singular directions, we can ask which words most strongly activate the source-side $k_i$ and $v_i$ directions, thereby identifying the tokens most likely to contribute information. Therefore, for each layer and attention head, we project each word's hidden state onto the top three $q$, $k$, and $v$ singular vectors and summarize the projection magnitude by a singular-value-weighted score, e.g., $\sum_{i=1}^{3}\omega_i|h^\top v_i|$ for the $v$ directions (details in Appendix B.2). We then compute, separately for each head, the Spearman correlation across words between this projection score and the static-information measure from~\S\ref{subsec:static_info}, and aggregate these head-wise correlations within each layer (Fig.~8). We find that higher-information words align more strongly with these directions, especially $k$ and $v$, suggesting that they more readily establish source-side routing and contribute larger write magnitudes. Moreover, this provides a mechanistic link between the empirical information measure in~\S\ref{subsec:static_info} and the singular directions governing attention computation, rather than leaving it contrived and rootless.

\subsection{Automantic Explaination for Attention Heads}
\label{subsec.head_semantic}


\begin{wrapfigure}[11]{r}{0.26\textwidth}
    \vspace{-1.3\baselineskip}
    \centering
    \includegraphics[width=0.26\textwidth]{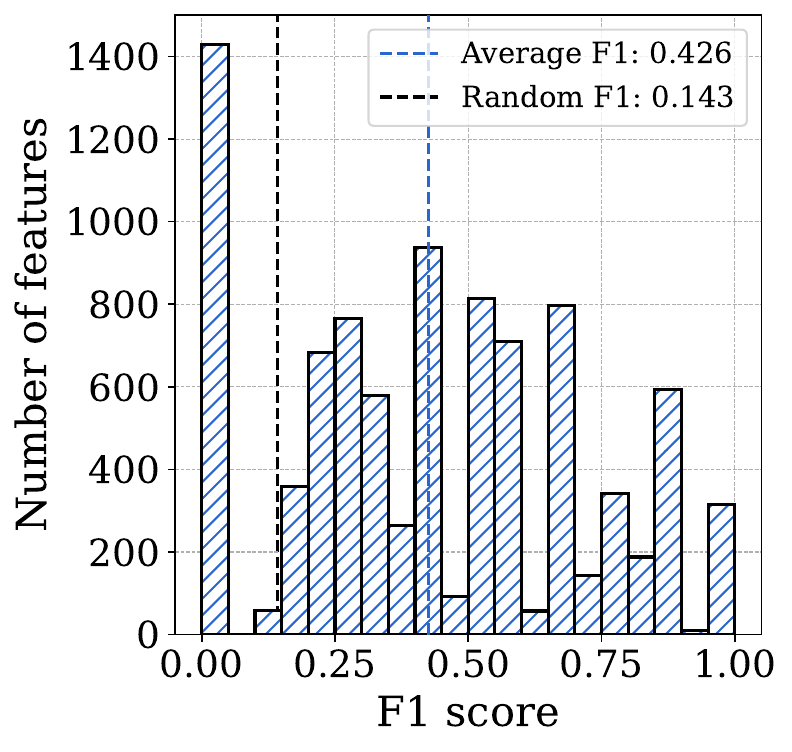}
    \vspace{-2.1\baselineskip}
    \caption{Histogram of AutoInterp F1 scores.}
    \label{fig:interp_score}
\end{wrapfigure}

\textbf{Automantic Interp Singular Vectors for Attention Heads.} Note that each $q,k,v$ singular vector extracted from attention heads corresponds to a direction for ``listening'' potential features~\citep{franco2026singular} from the hidden state of the current layer. We can therefore \textit{treat these singular vectors as (potential) features in the current layer} and apply automatic feature interpretation methods (e.g., AutoInterp) originally used for sparse autoencoder features~\citep{paulo2025automatically, bills2023language} to assign interpretations to these singular vectors, thereby revealing what semantics the attention heads are waiting for, then matching, and writing. \textbf{The main advantages of such an approach are}: \textbf{(1) Non-heurisitc}: as mentioned in~\S\ref{sec:preparation}, instead of guessing head functions from some observations, we can interpret them automatically and at scale. Then, the interpretation quality is no longer tied to mysterious heuristic inspiration, but mainly to the AutoInterp technique, whose improvement is already an inevitable demand of mechanistic interpretability~\citep{paulo2025automatically}. \textbf{(2) Assign Multiple Interpretations to A Head.} Some works~\citep{gould2024successor} show that an attention head can serve multiple distinct operations according to the inputs, and our method allows different singular vectors of one head to receive different interpretations (we show it in Table~\ref{tab:interp_example}), while previous methods assign each head one name and therefore fail to capture this polyfunctionality.


\begin{wrapfigure}[9]{r}{0.26\textwidth}
    \vspace{-1.4\baselineskip}
    \centering
    \includegraphics[width=\linewidth]{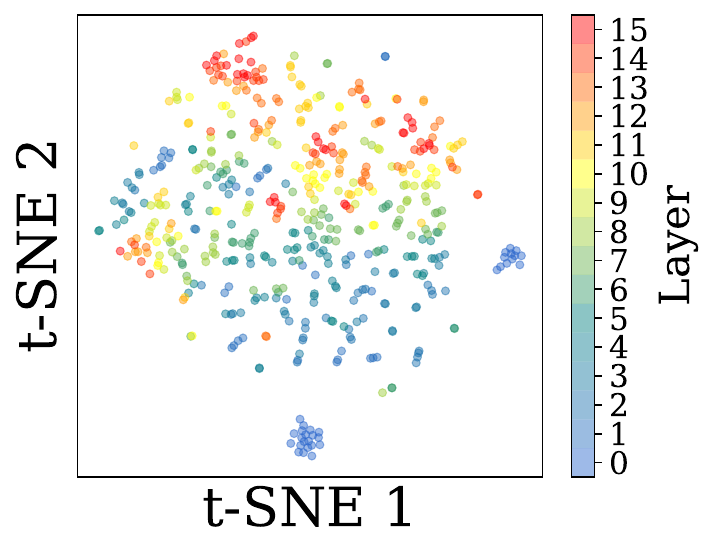}
    \vspace{-2\baselineskip}
    \caption{t-SNE of head embedding.}
    \label{fig:attention_head_tsne}
\end{wrapfigure}

\textbf{Singular Vectors Provide Human-understandable Attention Head Interpretation.} We conduct AutoInterp~\citep{paulo2025automatically} on the top-3 singular vectors of all the heads in Llama 3.2-1B, with some examples shown in Table~\ref{tab:interp_example} and AutoInterp F1 (compares the sequences ``predicted to activate the feature'' from each feature’s explanation, against ones ``actually activate'') in Fig.~\ref{fig:interp_score} (details and all interpretation in Appendix~\ref{appendix.exp.autointerp} and~\ref{appendix.more_results}), where: \textbf{(1) AutoInterp Scores:} Although we might be desire to argue that these singular vectors are meaningful and ``worth interpreting'' because the F1 scores are high, this inference is invalid: prior work shows that AutoInterp can generate plausible explanations even for arbitrary random features~\citep{heap2026automated, korznikov2026sanity}, which clearly have no value to interpret. In contrast, the singular vectors are intrinsically significant directions in the attention parameter spaces, and our following ablation experiment further establishes their functional relevance. Interestingly, singular vectors yield many zero-score features, possibly because they capture more abstract or functional semantics which can not be interpreted by literal-based AutoInterp~\citep{cho2025binary}. As discussed in our Limitations, interpreting such features depends on advances in AutoInterp tools, and is out of scope of this work. \textbf{(2) Human-understandable Attention Operation Occurs.} For example, the $q_2$ and $k_2$ singular vectors in Layer 11 Head 25 tell an interesting story, as also shown in the attention scores in Fig.~\ref{fig:ablation_head}-Left: this head produces high attention scores on (\textbf{queries}) holiday-related words, such as ``pumpkin'', and (\textbf{keys}) words associated with seasonal events, such as ``Halloween''. It then copies Halloween-related semantics to the later ``pumpkin'' token, overriding the ``pumpkin'' with ``Halloween pumpkin'' over ``ordinary vegetables''. Also, as a causality experiment, we set (``ablate'') the singular value $\lambda_2\coloneqq0$, then the attention score for the previous ``pumpkin'' operation is damaged (Fig.~\ref{fig:ablation_head}, Appendix~\ref{appendix.more_results} for more cases). \textbf{(3) One Attention Head Contains Multiple Interpretations.} The Layer 11 Head 25 also contains some ``position localization'' singular vectors, which is substantially different from the ``Halloween head'' where previous works would stop.







\subsection{Embed Attention Head Sets into Continuous Vector Space} 
\label{subsec.embedding}



\begin{wrapfigure}[13]{r}{0.4\textwidth}
    \vspace{-1.3\baselineskip}
    \includegraphics[width=\linewidth]{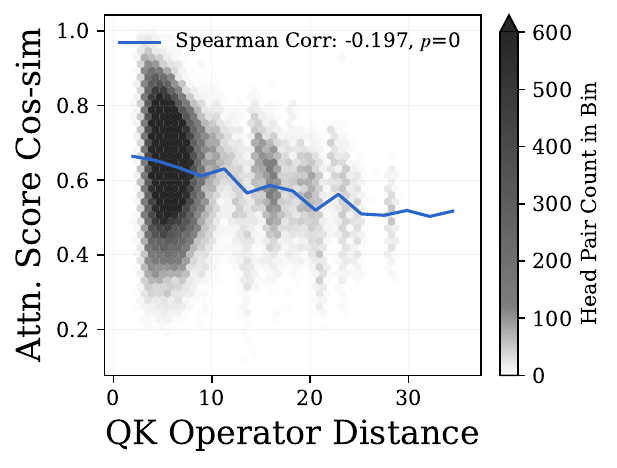}
    \vspace{-2\baselineskip}
    \caption{Correlation between head embedding distance and attention map.}
    \label{fig:score_singluar_corr}
\end{wrapfigure}

\textbf{Continous Attention Head Embedding.} As mentioned in~\S\ref{sec:preparation}, prior work has proposed various taxonomies of attention heads~\citep{zheng2024attention}, which has some scientific and technique limitations. Our singular-vector perspective instead enables a continuous representation: as shown in Fig.~\ref{fig:attention_head_tsne}, we embed each attention head using its $[q_1;k_1;-q_1;-k_1]$ vector (since the singular vectors are unoriented) into a continuous space. Also, we find that the distance between two attention heads ($C_2^{512}$ pairs for Llama 3.2-1B, see Appendix~\ref{appendix.details_embedding} for the distance calculation) is negatively correlated with the similarity of attention scores produced by heads. The correlation is slight, which might be because the input spaces of these heads are not so aligned (layer-alignment left for future work), but still statistically significant.


\section{Conclusion and Discussion}

\textbf{Conclusions.} We revisit contextualization as a principle: informative words are information sources, and less-informative ones are sinks, and validate it through observation of LM outputs and hidden states, then identify its mechanism by reinterpreting attention heads as gated information channels, whose singular vectors are controllers and identifiers of head operations, advancing previous heuristic head discovery and discrete taxonomies into automantic discovery and continuous head embedding.

\textbf{Future Directions Enabled by Our Work.} \textbf{(1) Absorption-Relay Dynamics.} Information richness may itself be dynamic rather than fixed. After absorbing information, some low-information words may become high-information sources and serve as new sources, forming an interesting absorption–relay dynamics that needs further investigation.  \textbf{(2) Optimization Pressure on Contextualization.} Although we hint at the possibility that low-information words may improve model outputs by collecting contextual information, how this contributes to prediction and how training pressure shapes the process remain unclear.  \textbf{(3) Primitive Operations of Language Modeling.} A fundamental question in training dynamics and LLM evaluation is what an LM's primitive operations are. By continuous representations of attention-head operations, our framework enables geometric analysis of meaningful basis directions, which correspond to operations. \textbf{(4) Higher-order Statistical Properties.} Continuing the analogy to statistical mechanics, future work may combine basic contextualization statistics into higher-order quantities that better predict language-modeling quality.

\textbf{Limitations.} \textbf{(1) Purity of Singular-Vector Features.} Unlike SAEs, the attention head singular vectors need not be monosemantic, and may also encode a feature through multiple directions. This concern is whether a single-sentence interpretation should be bijected to a singular vector, and indeed a technical issue for AutoInterp. \textbf{(2) Automatic Feature Interpretation.} We rely on existing AutoInterp methods with known limitations~\citep{bills2023language, paulo2025automatically, tian2025measuring, maher2026multishot}, including potentially spurious LLM-generated explanations and limited generalization. These limitations concern the interpretation method rather than our framework, which provides a method-agnostic interface compatible with improved interpreters. 

\subsection*{AI use statement}
We utilized AI to assist in reviewing language and code implementation, generating the datasets used in the paper, and translating the original manuscript written by the authors in their native language into English. Additionally, AI handled large-scale processing tasks, such as identifying features suitable for illustrative examples and organizing tables of vocabulary listed in the appendix, thereby automating repetitive work.

\subsubsection*{Reproducibility Statement}
The experimental code for this paper will be made publicly available upon acceptance.

\subsubsection*{Acknowledgments}
This work was supported by the SPDR Funding (Number 202601094073), and Programs for Junior Scientists at RIKEN.

\bibliography{iclr2026_conference}
\bibliographystyle{iclr2027_conference}

\clearpage
\appendix
\begin{center}
{\LARGE {\textbf{Appendices}}}
\end{center}
\section{Dataset Details}
\label{appendix.dataset}

\textbf{Sampled Words in Categories.} In~\S\ref{sec:settings}, we sample around 50 words in 16 categories, and we list them, together with their word frequency estimated on the first 500,000 samples of Pile~\citep{gao2020pile} in Table~\ref{tab:dataset-vocabulary}. Also, its structured \texttt{JSON} version has also been attached to the supplementary materials.

\textbf{Queries for Experiment in~\S\ref{subsec:static_info}.} In~\S\ref{sec:settings}, we sample 150 queries for the estimation of word information content, we list them together with their output candidates in Table~\ref{tab:dataset-questions}. Also, its structured \texttt{JSON} version has also been attached to the supplementary materials.

\textbf{Sentence Templates for Experiment in~\S\ref{subsec.h2}} In~\S\ref{sec:settings}, we sample 128 templates for evaluating the contextualization magnitude, we list them in Table~\ref{tab:dataset-templates-any}. Also, its structured \texttt{JSON} version has also been attached to the supplementary materials.
 
\section{Experimental Details}
\label{appendix.exp}

\subsection{Experimental Details in~\S\ref{subsec:static_info}}
\label{appendix.exp.1}

\textbf{How is the normalized answer distribution $y$ calculated?} We adopt a noisy-channel-style approach to calculate the output distribution $y$. For each question template (e.g., ``The gender of [Word]?''), we instantiate the template with a specific word, yielding a query such as ``The gender of Newton:''. We then append each output candidate to the query to construct a set of joint sequences, such as ``The gender of Newton: male'', ``The gender of Newton: female'', and ``The gender of Newton: non-binary''. We compute the LM loss for each sequence, obtaining a vector such as $[3.2, 9.0, 10.0]$, and take the softmax over its negated values as the final output distribution $y$. The batch-calibration~\citep{zhou2024batch} described below further removes loss biases arising from implementation details such as differences in label length.

\textbf{Motivation for proposing a new information estimation method against the previous word frequency-based method.} Prior works~\citep{oyama2023norm} have mainly argued that the amount of information carried by a token is negatively correlated with its frequency. Although our results are also consistent with this tendency, our formulation has a fundamental advantage, which can be illustrated by a simple thought experiment: Suppose there exists a token that has never appeared in the training corpus. Its embedding and unembedding vectors would necessarily remain essentially random, since no gradient has ever updated them. In a high-dimensional hidden space, such random vectors are overwhelmingly likely to be nearly orthogonal to the operational directions of attention heads and other modules, which are typically characterized by singular vectors of their parameter matrices. Consequently, this token would be unlikely to participate meaningfully in the model's computation and would exert only a negligible influence on the output. In contrast, conventional frequency-based formulations would assign such a zero-frequency token an unbounded amount of information, leading to a contradiction. Therefore, we propose a new measurement of empirical information amount.

\subsection{Experimental Details in~\S\ref{subsec.6.2}}
\label{appendix.exp.projection}

To quantify the projection length of a word's hidden state vector $h$ (we choose the last token hidden state for multi-token words) onto a set of singular vectors (the top 3 in our experiments) for subsequent correlation analysis, we compute the singular-value-weighted sum of the projection magnitudes onto these vectors, as (using the $v_i$ singular vectors as an example): $\sum_{i=1}^3 \omega_i\vert h^\top v_i\vert\in\mathbb{R}$, which serves as one of the variables of the correlation calculation in Fig.~\ref{fig:projection}. We use the absolute value since the singular vectors of a matrix are intrinsically unoriented.

\subsection{Experimental Details in~\S\ref{subsec.head_semantic}}
\label{appendix.exp.autointerp}

In this experiment, we treat both the positive and negative directions of each extracted singular vector as separate features to be interpreted (again, note that singular vectors of a matrix are intrinsically unoriented). For example, if one LM has 512 attention heads and we consider the top three singular vectors, then across $q,k,o,v$ we obtain a total of $2 \times 512 \times 3 \times 4$ features. We then inject these directions into the encoder weights of a standard ReLU SAE and use the AutoInterp framework to generate explanations. We scan 5,000,000 tokens from Pile to compute feature activations, selecting the 16 highest-activating examples for explanation generation. For scoring, we use 28 examples in total, including 8 activating examples and 20 randomly sampled non-activating examples.

\subsection{Attention Head Distance Calculation in~\S\ref{subsec.embedding}}
\label{appendix.details_embedding}

In~\S\ref{subsec.embedding}, we utilize the top-1 singular vectors $q_1$ and $k_1$ to calculate the ``operation distance'' of two attention heads (one of which is marked with a superscript prime) as:
\begin{equation}
    \text{OperationDistance} = \Vert q_1\lambda_1k_1^\top-q'_1\lambda'_1k'^{\top}_1\Vert_F,
\end{equation}
where $q_1k_1^\top\in\mathbb{R}^d$. Such a calculation removes the effect of sign while faithfully preserving the distances between singular vectors, since the norm is homogeneous.

The attention distance on the vertical axis is calculated from 256 sentences from the Pile, where, for an attention head pair, we calculate the input-wise attention score matrix cosine similarity and average such cosine similarity over all inputs as the attention score cosine similarity between the two attention heads.

\section{Experiment in~\S\ref{subsec.h2} on Grammatically-correct Inputs}
\label{appendix.specific_template}

One concern is that, in the experiments in~\S\ref{subsec.h2}, we took the Cartesian product of all word categories and all sentence templates. Although this provides proper control over variables (since the filled words share the same position and context), it may generate some grammatically incorrect sentences. We therefore removed all such ungrammatical sentences and rerun the experiments, obtaining the new results shown in Fig.~\ref{fig:more_specific_temp_begin} to~\ref{fig:more_specific_temp_end}, which are aligned with the normal ones.

\section{Token Length Controlling Analysis}
\label{appendix.token_length_control}

\begin{wrapfigure}[18]{r}{0.45\textwidth}
    \vspace{-1.6\baselineskip}
    \centering
    \includegraphics[width=\linewidth]{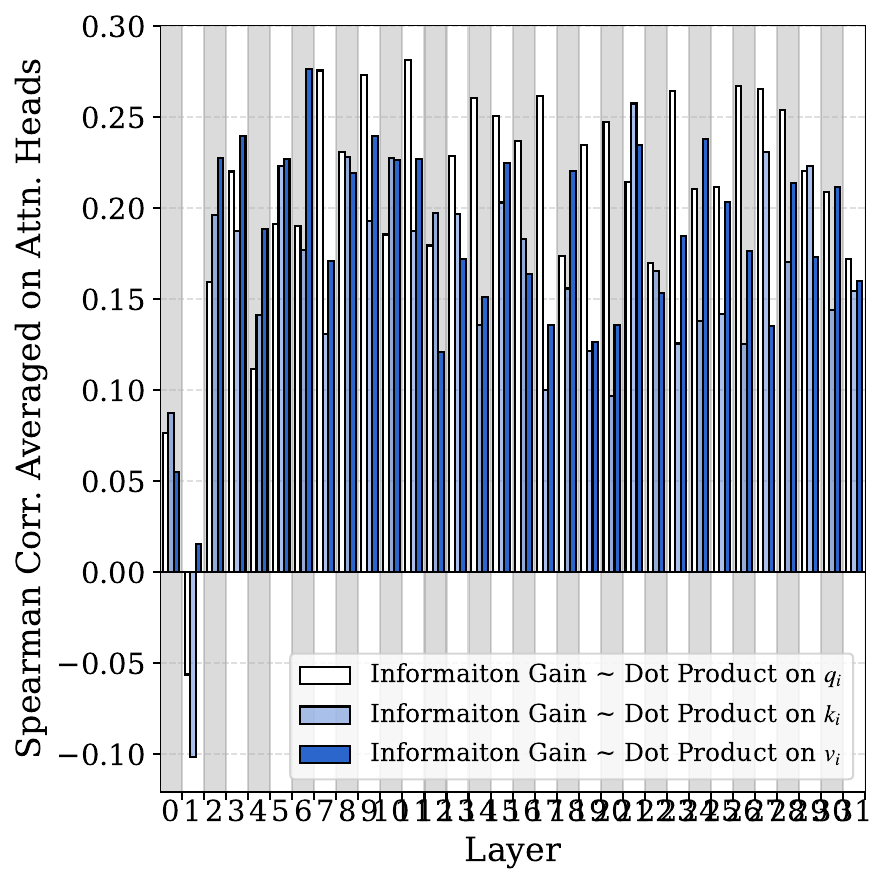}
    \vspace{-2\baselineskip}
    \caption{Experiment result of Fig.~\ref{fig:projection} on Llama 3-8B (only 20\% of heads are calculated).}
    \label{fig:more_fig8}
\end{wrapfigure}

In \S\ref{subsec:static_info}, we suspect that a word's tokenization length may act as a potential confounding factor: rarer words may be split into more tokens, thereby trivially producing stronger output effects and more stable representations. In this section, we rule out this concern. Specifically, we directly prune the vocabulary dataset and retain only words that are tokenized into exactly two tokens (we choose two-token words because this restriction still preserves a substantial number of words in every category). We then reproduce all the experiments under this controlled setting:

\begin{itemize}[topsep=0pt, itemsep=0pt, leftmargin=15pt]
\item \textbf{Fig.~\ref{fig:static_information}:} The augmentated results on 2-token-input for Fig.~\ref{fig:static_information} are shown in Fig.~\ref{fig:more_exp1_len2_begin} to~\ref{fig:more_exp1_len2_end}. The results are aligned with the uncontrolled variations.
\item \textbf{Fig.~\ref{fig:donor_and_receptor}:} The augmentated results on 2-token-input for Fig.~\ref{fig:donor_and_receptor} are shown in Fig.~\ref{fig:more_exp2_len2_begin} to~\ref{fig:more_exp2_len2_end}. The results are aligned with the uncontrolled variations.
\item \textbf{Fig.~\ref{fig:donor_and_receptor_correlation}:} The augmentated results on 2-token-input for Fig.~\ref{fig:donor_and_receptor_correlation} (full-layer correlation version) are shown in Fig.~\ref{fig:more_exp3_len2_begin} to~\ref{fig:more_exp3_len2_end}. The results are aligned with the uncontrolled variations.
\end{itemize}

\section{Augmentated Results}
\label{appendix.more_results}
\label{appendix.h2_full_layer}

\textbf{Full Layer Correlation of Fig.~\ref{fig:donor_and_receptor_correlation}.} We show the full-layer correlation coefficient in Fig.~\ref{fig:more_exp3_begin} to~\ref{fig:more_exp3_end} for all 10 models. Most models and layers exhibit a clear positive correlation between word information gain and absorption negative measurement.

\textbf{Augmentated Results of Fig.~\ref{fig:static_information}.} We show the remaining 9 model results under the same experiment as Fig.~\ref{fig:static_information} in Fig.~\ref{fig:more_exp1_begin} to~\ref{fig:more_exp1_end}. The results are aligned with Fig.~\ref{fig:static_information}.

\begin{wrapfigure}[13]{r}{0.4\textwidth}
    \vspace{-1.4\baselineskip}
    \centering
    \includegraphics[width=\linewidth]{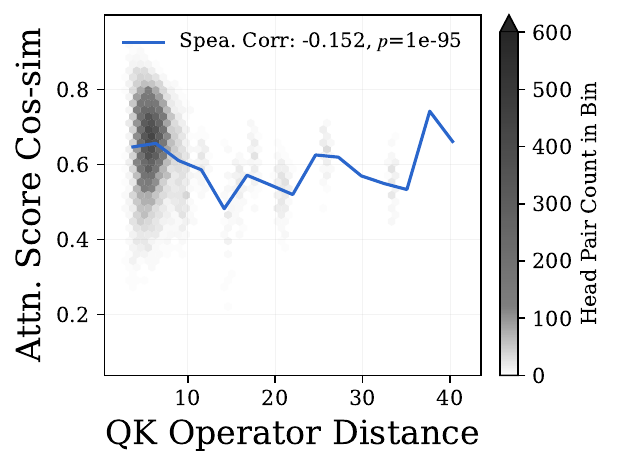}
    \vspace{-2\baselineskip}\caption{Experiment result of Fig.~\ref{fig:score_singluar_corr} on Llama 3-8B (only 20\% of heads are calculated).}
    \label{fig:more_fig9}
\end{wrapfigure}

\textbf{Augmentated Results of Fig.~\ref{fig:donor_and_receptor}.} We show the remaining 9 model results under the same experiment as Fig.~\ref{fig:donor_and_receptor} in Fig.~\ref{fig:more_exp2_begin} to~\ref{fig:more_exp2_end}. The results are aligned with Fig.~\ref{fig:donor_and_receptor}.

\textbf{Augmentated Results of Table~\ref{tab:interp_example}.} We release the interpretation of full attention head features for Llama 3.2-1B, 20\% of random-sampled head features for Qwen 3-8B and Llama 3-8B in the Supplementary Material. Please refer to the OpenReview page of this paper.

\begin{figure}
    \centering
    \includegraphics[width=\linewidth]{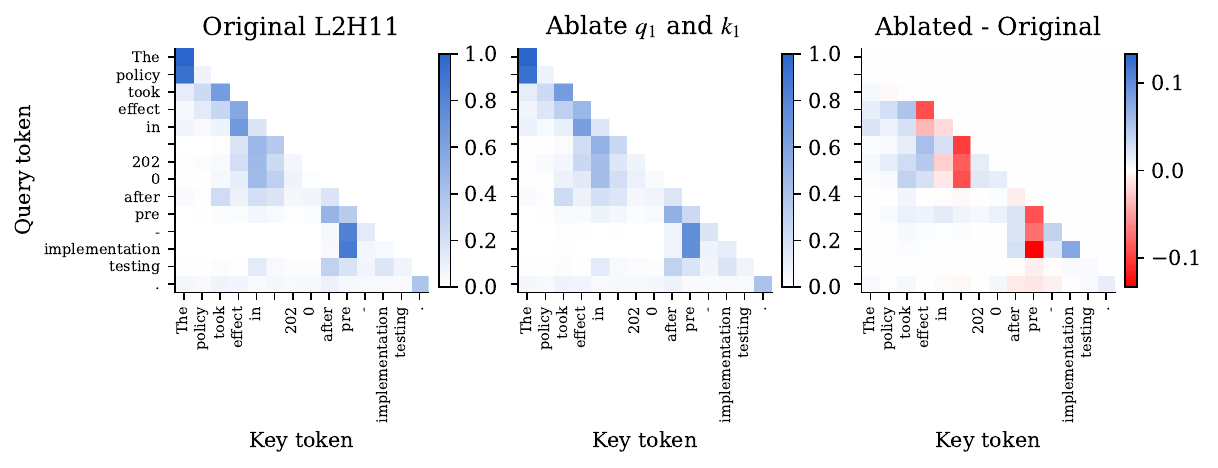}
    \vspace{-2\baselineskip}
    \caption{More attention score visualization like Fig.~\ref{fig:ablation_head} on Llama 3.2-1B Layer 2 Head 11.}
    \label{fig:moreheadvis_start}
    \vspace{-2\baselineskip}
\end{figure}

\textbf{Augmentated Results of Fig.~\ref{fig:ablation_head}.} We visualize more cases like Fig.~\ref{fig:ablation_head} in Fig.~\ref{fig:moreheadvis_start} to~\ref{fig:moreheadvis_end}.

\textbf{Augmentated Results of Fig.~\ref{fig:projection} and~\ref{fig:score_singluar_corr}.} We show the results of Fig.~\ref{fig:projection} and~\ref{fig:score_singluar_corr} on Llama 3-8B (due to computational limitations, only 20\% of the randomly sampled attention heads are calculated) in Fig.~\ref{fig:more_fig8} and Fig.~\ref{fig:more_fig9}.

\clearpage

\begin{figure}
    \centering
    \includegraphics[width=\linewidth]{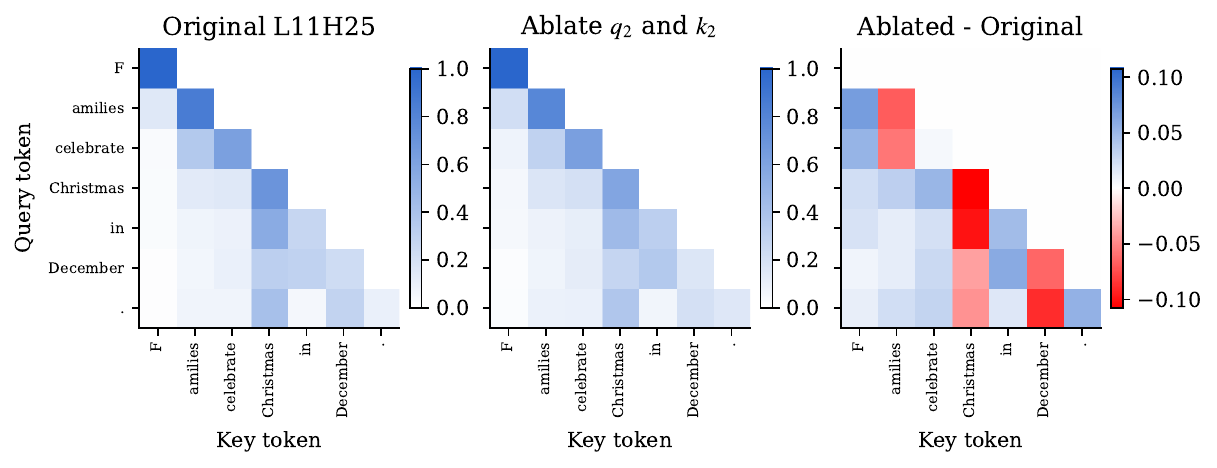}
    \vspace{-2\baselineskip}
    \caption{More attention score visualization like Fig.~\ref{fig:ablation_head} on Llama 3.2-1B Layer 11 Head 25.}
    \vspace{-2\baselineskip}
\end{figure}

\begin{figure}
    \centering
    \includegraphics[width=\linewidth]{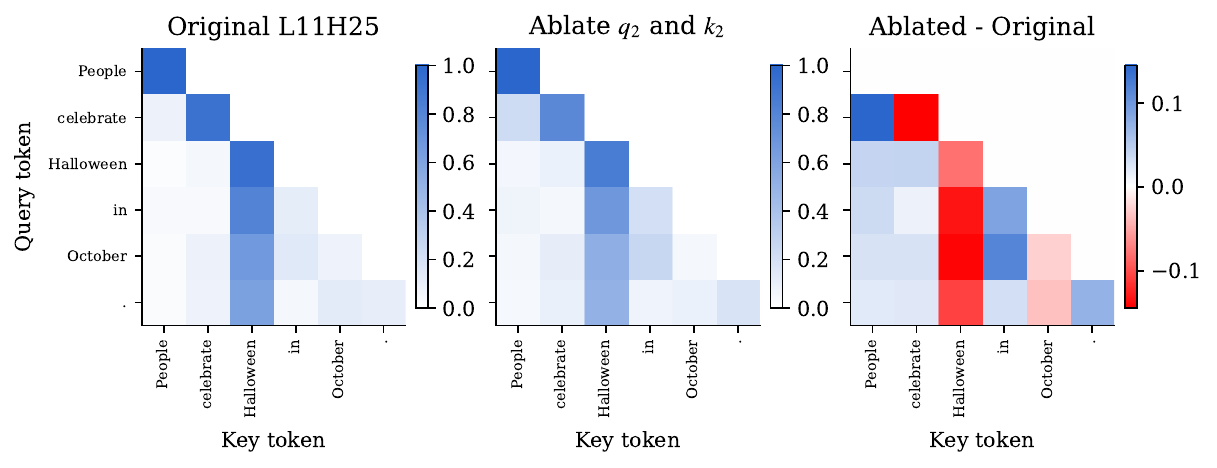}
    \vspace{-2\baselineskip}
    \caption{More attention score visualization like Fig.~\ref{fig:ablation_head} on Llama 3.2-1B Layer 11 Head 25.}
    \vspace{-2\baselineskip}
\end{figure}

\begin{figure}
    \centering
    \includegraphics[width=\linewidth]{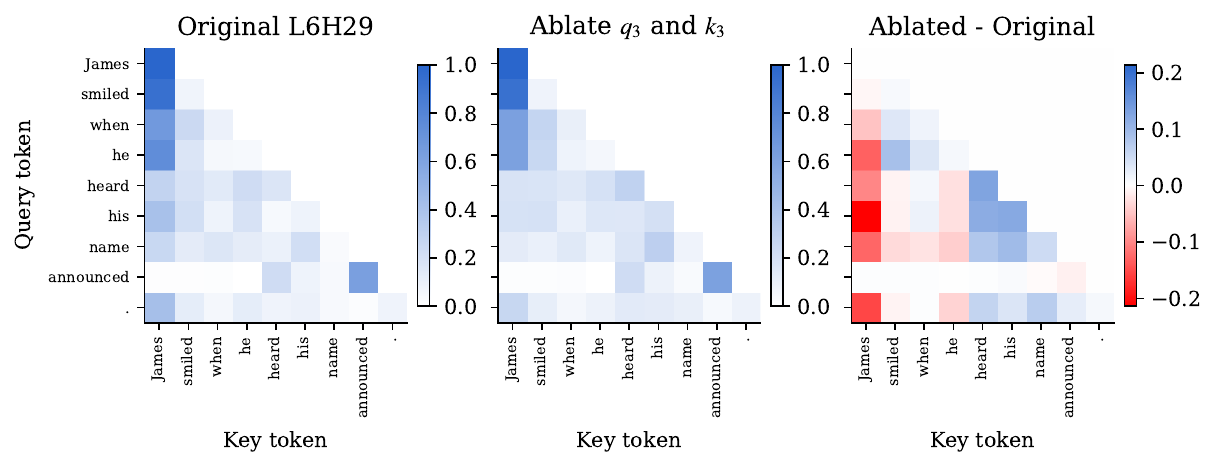}
    \vspace{-2\baselineskip}
    \caption{More attention score visualization like Fig.~\ref{fig:ablation_head} on Llama 3.2-1B Layer 6 Head 29.}
    \vspace{-2\baselineskip}
\end{figure}

\begin{figure}
    \centering
    \includegraphics[width=\linewidth]{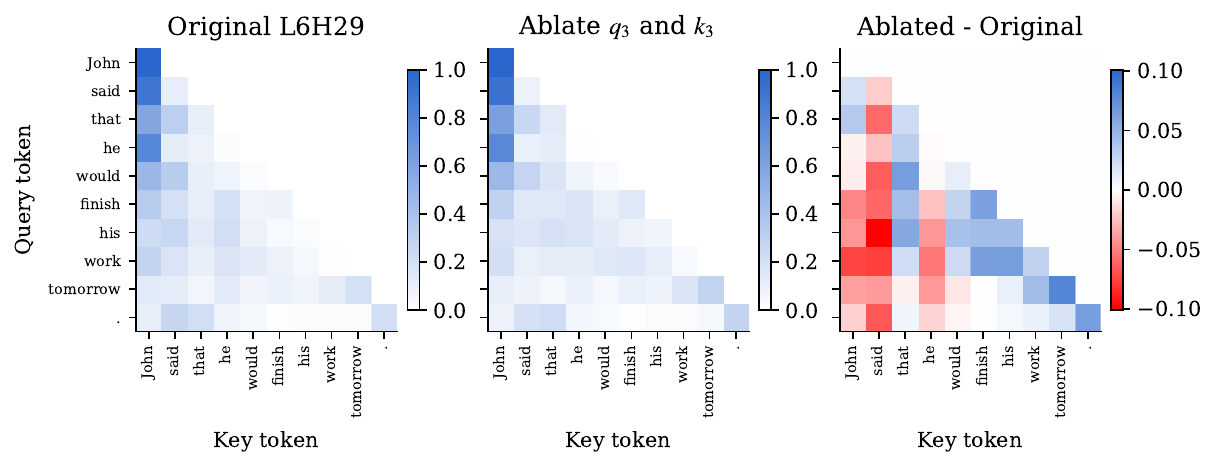}
    \vspace{-2\baselineskip}
    \caption{More attention score visualization like Fig.~\ref{fig:ablation_head} on Llama 3.2-1B Layer 6 Head 29.}
    \vspace{-2\baselineskip}
\end{figure}

\begin{figure}
    \centering
    \includegraphics[width=\linewidth]{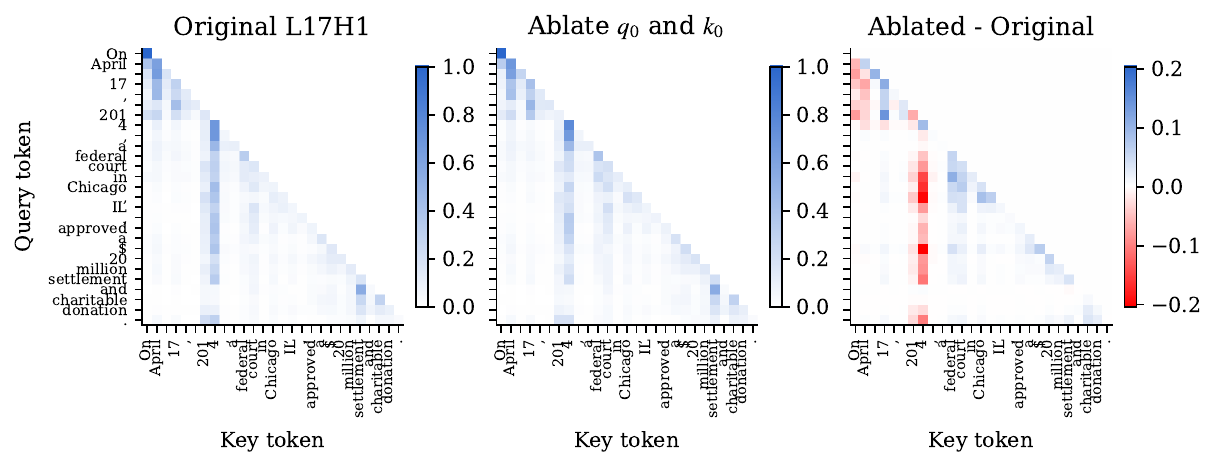}
    \vspace{-2\baselineskip}
    \caption{More attention score visualization like Fig.~\ref{fig:ablation_head} on Llama 3-8B Layer 17 Head 1 with the interpretation of the first set of singular vector: $q_1$: ``specific dates and mentions of events or locations like the Bahamas, court cases, and significant donations'', $k_1$: ``years denoted in a specific format with examples like 2014 and 2009'', $v_1$: ``geographic locations and their corresponding state abbreviations in text''.}
    \label{fig:moreheadvis_end}
\end{figure}
\begin{figure}[t]
    \centering
    \begin{minipage}[t]{0.4\linewidth}
    \includegraphics[width=\linewidth]{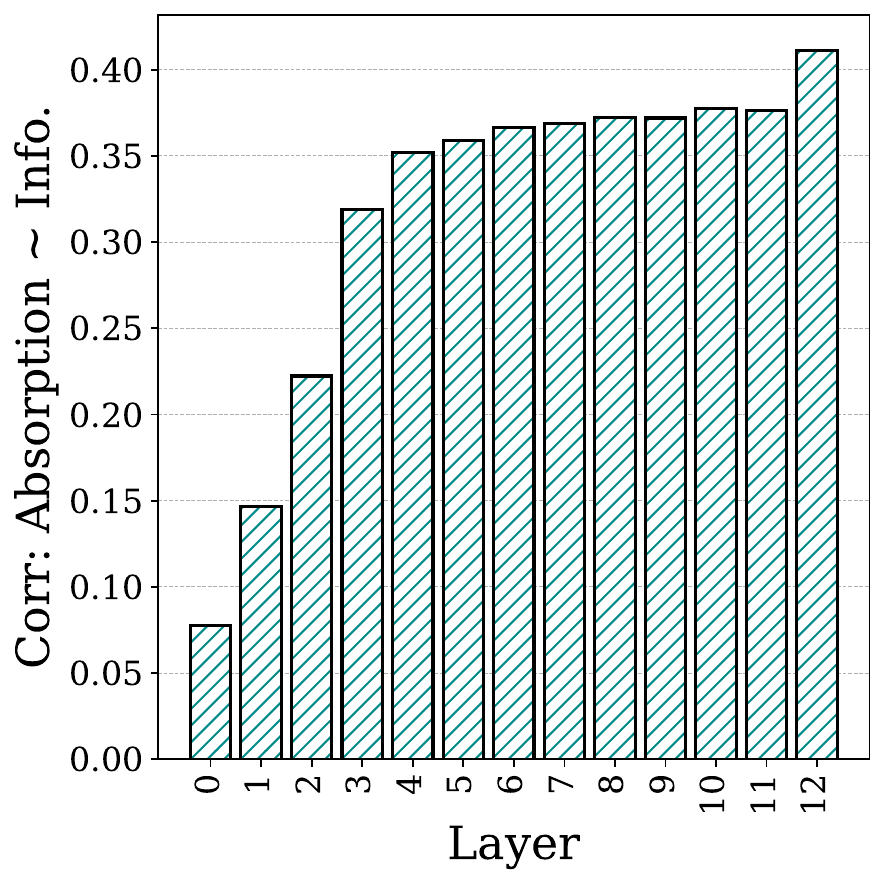}
    \vspace{-2\baselineskip}\caption{Experiment result of Fig.~\ref{fig:static_information} on BERT-Base with grammatically correct inputs.}
    \label{fig:more_specific_temp_begin}
    \end{minipage}\hfill
    \centering
    \begin{minipage}[t]{0.57\linewidth}
    \includegraphics[width=\linewidth]{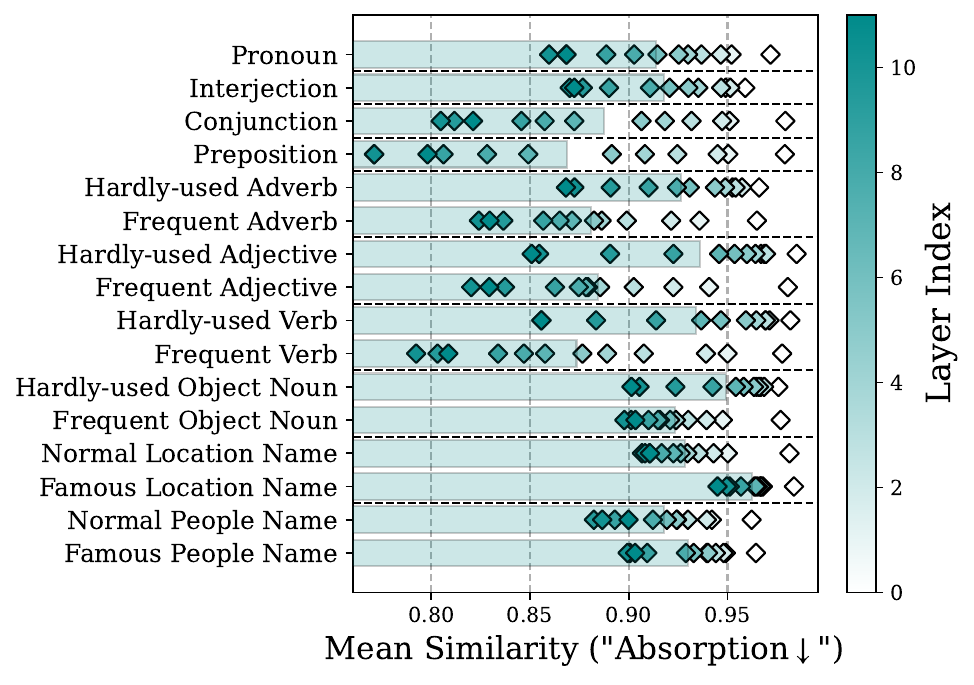}
    \vspace{-2\baselineskip}\caption{Experiment result of Fig.~\ref{fig:donor_and_receptor_correlation} on BERT-Base with grammatically correct inputs.}
    \end{minipage}\hfill
\vspace{-0.6\baselineskip}\end{figure}

\begin{figure}[t]
    \centering
    \begin{minipage}[t]{0.47\linewidth}
    \includegraphics[width=\linewidth]{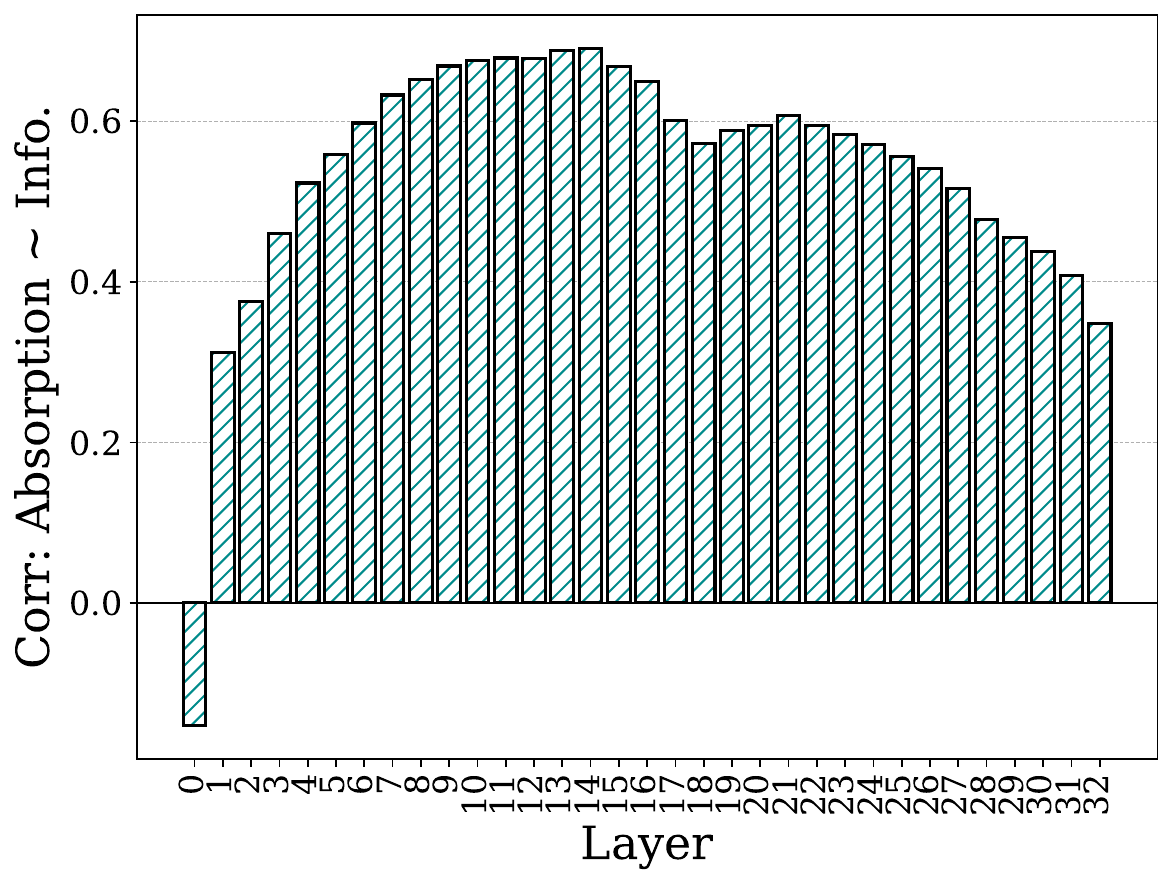}
    \vspace{-2\baselineskip}\caption{Experiment result of Fig.~\ref{fig:static_information} on Llama 3-8B with grammatically correct inputs.}
    \end{minipage}\hfill
    \centering
    \begin{minipage}[t]{0.505\linewidth}
    \includegraphics[width=\linewidth]{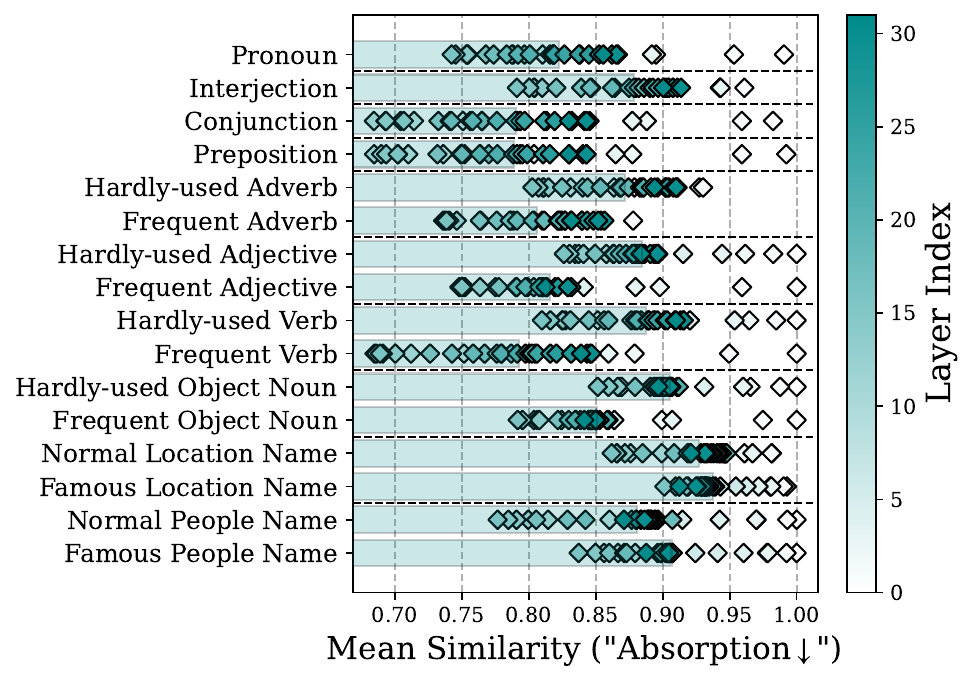}
    \vspace{-2\baselineskip}\caption{Experiment result of Fig.~\ref{fig:donor_and_receptor_correlation} on Llama 3-8B with grammatically correct inputs.}
    \end{minipage}\hfill
\vspace{-0.6\baselineskip}\end{figure}

\begin{figure}[t]
    \centering
    \begin{minipage}[t]{0.4\linewidth}
    \includegraphics[width=\linewidth]{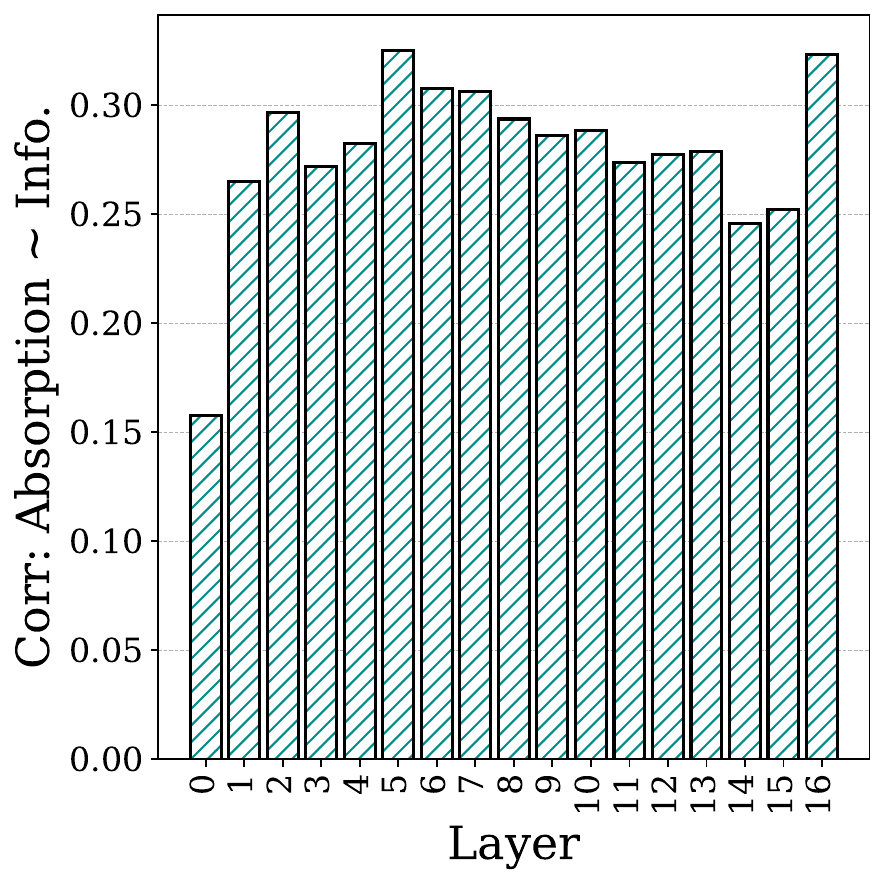}
    \vspace{-2\baselineskip}\caption{Experiment result of Fig.~\ref{fig:static_information} on Llama 3.2-1B with grammatically correct inputs.}
    \end{minipage}\hfill
    \centering
    \begin{minipage}[t]{0.57\linewidth}
    \includegraphics[width=\linewidth]{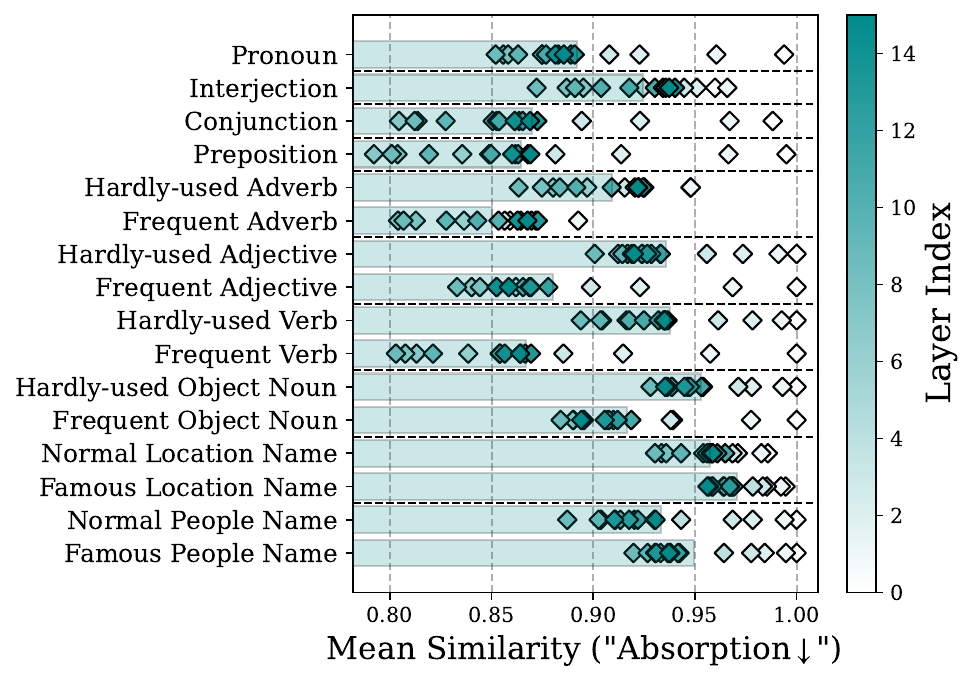}
    \vspace{-2\baselineskip}\caption{Experiment result of Fig.~\ref{fig:donor_and_receptor_correlation} on Llama 3-8B with grammatically correct inputs.}
    \end{minipage}\hfill
\vspace{-0.6\baselineskip}\end{figure}

\begin{figure}[t]
    \centering
    \begin{minipage}[t]{0.50\linewidth}
    \includegraphics[width=\linewidth]{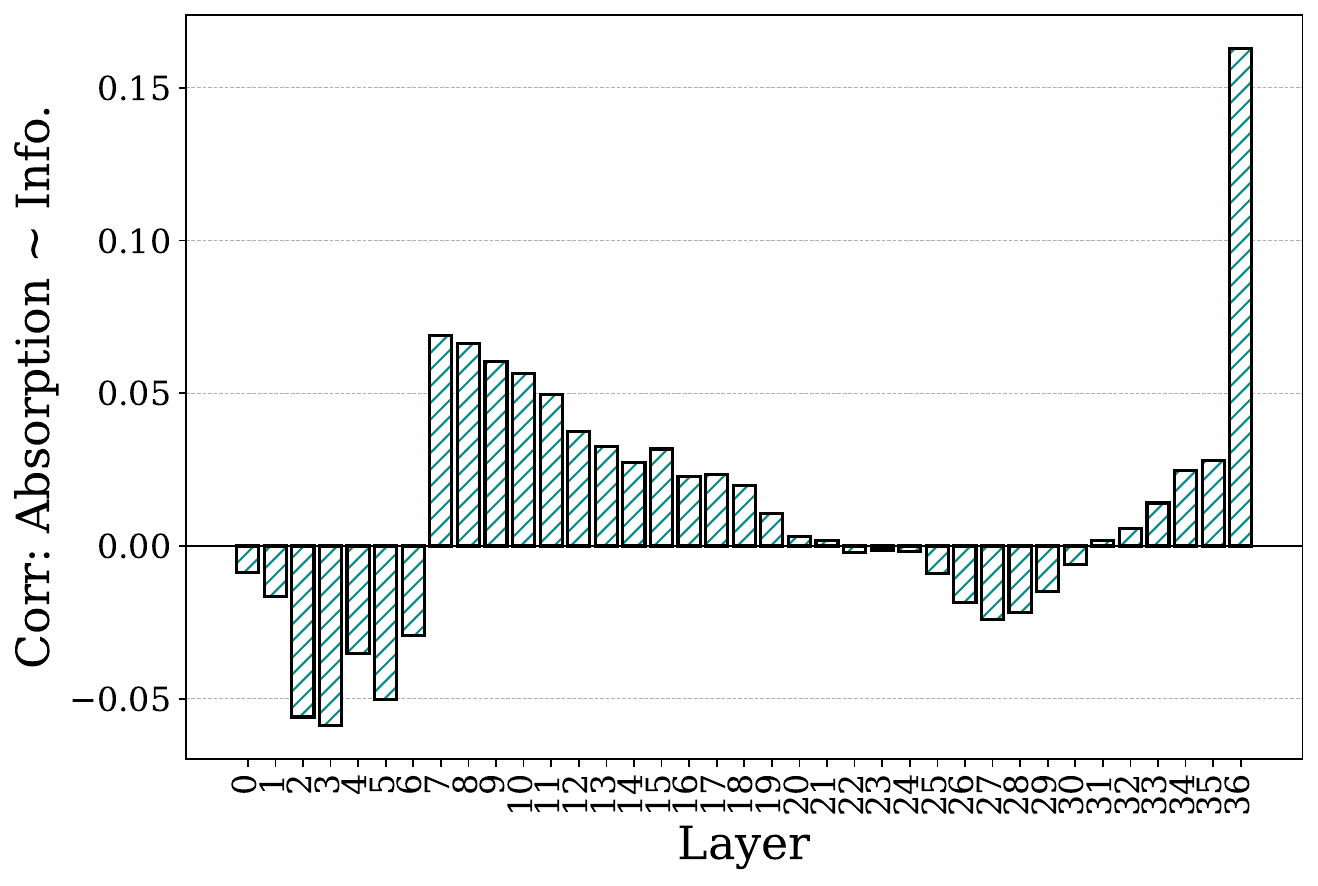}
    \vspace{-2\baselineskip}\caption{Experiment result of Fig.~\ref{fig:static_information} on Qwen 3-8B with grammatically correct inputs.}
    \end{minipage}\hfill
    \centering
    \begin{minipage}[t]{0.48\linewidth}
    \includegraphics[width=\linewidth]{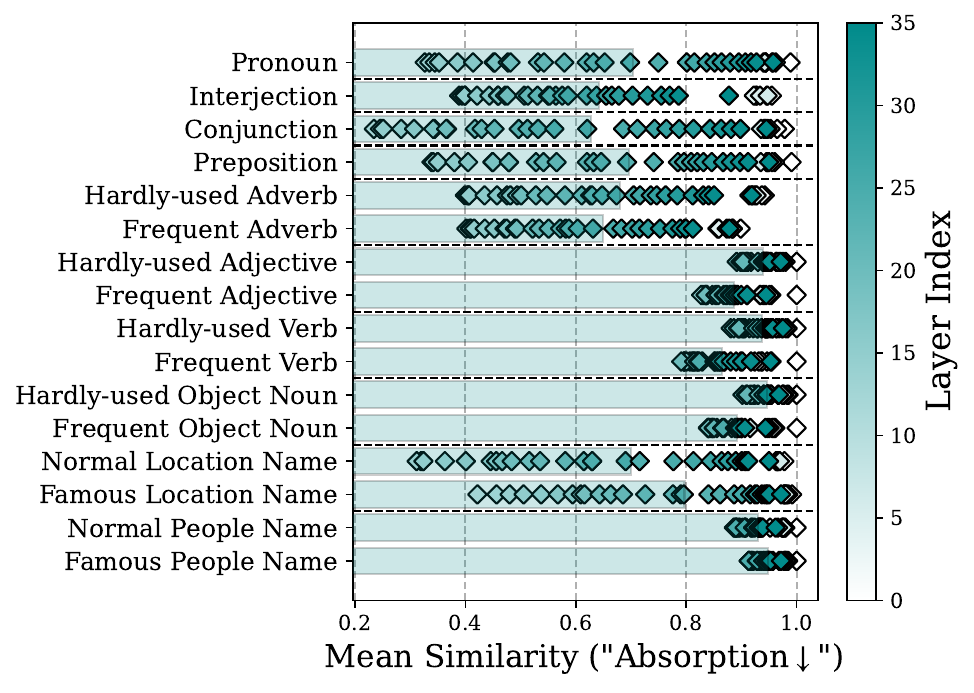}
    \vspace{-2\baselineskip}\caption{Experiment result of Fig.~\ref{fig:donor_and_receptor_correlation} on Qwen 3-8B with grammatically correct inputs.}
    \end{minipage}\hfill
\vspace{-0.6\baselineskip}\end{figure}

\begin{figure}[t]
    \centering
    \begin{minipage}[t]{0.4\linewidth}
    \includegraphics[width=\linewidth]{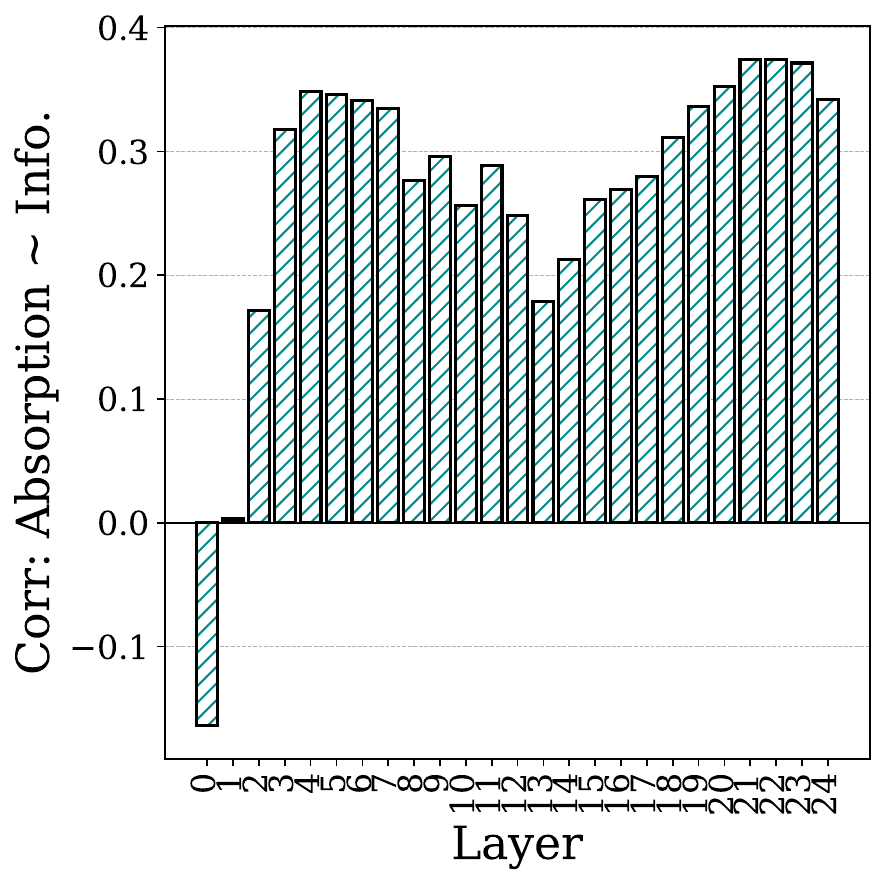}
    \vspace{-2\baselineskip}\caption{Experiment result of Fig.~\ref{fig:static_information} on XLM-RoBERTa-Large with grammatically correct inputs.}
    \end{minipage}\hfill
    \centering
    \begin{minipage}[t]{0.57\linewidth}
    \includegraphics[width=\linewidth]{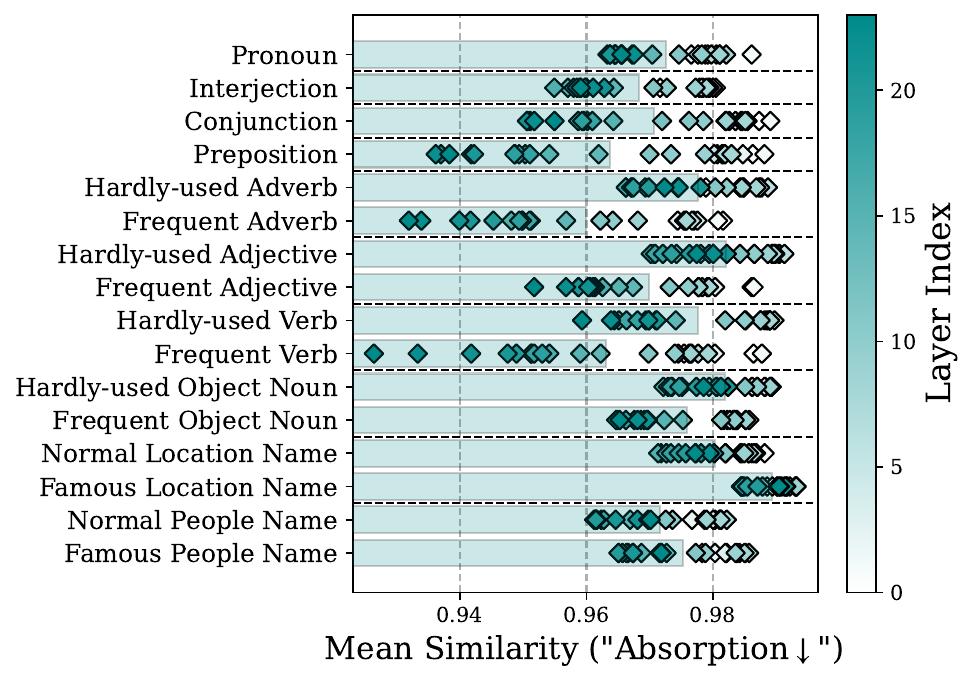}
    \vspace{-2\baselineskip}\caption{Experiment result of Fig.~\ref{fig:donor_and_receptor_correlation} on XLM-RoBERTa-Large with grammatically correct inputs.}
    \label{fig:more_specific_temp_end}
    \end{minipage}\hfill
\vspace{-0.6\baselineskip}\end{figure}
\begin{figure}[t]
    \centering
    \begin{minipage}[t]{0.49\linewidth}
    \includegraphics[width=\linewidth]{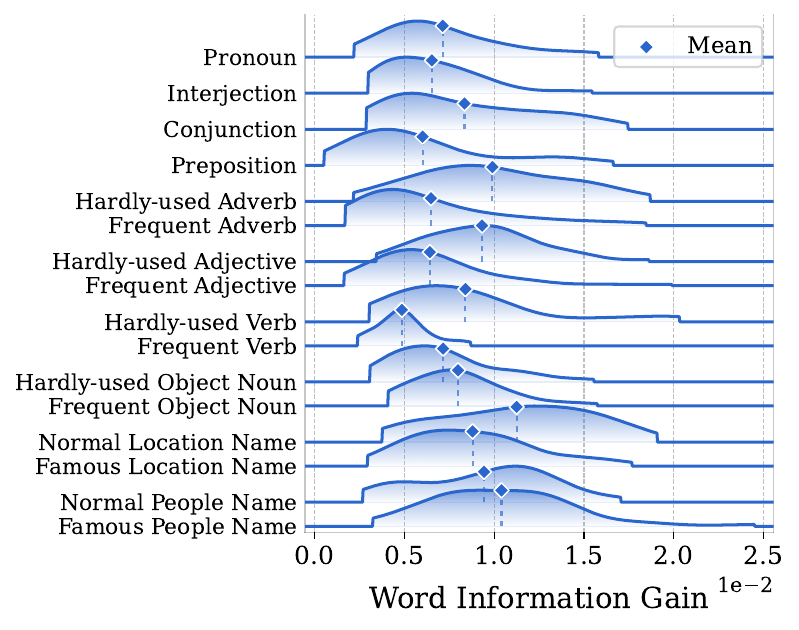}
    \vspace{-2\baselineskip}\caption{Experiment result of Fig.~\ref{fig:static_information} on BERT-Base.}
    \label{fig:more_exp1_begin}
    \end{minipage}\hfill
    \centering
    \begin{minipage}[t]{0.49\linewidth}
    \includegraphics[width=\linewidth]{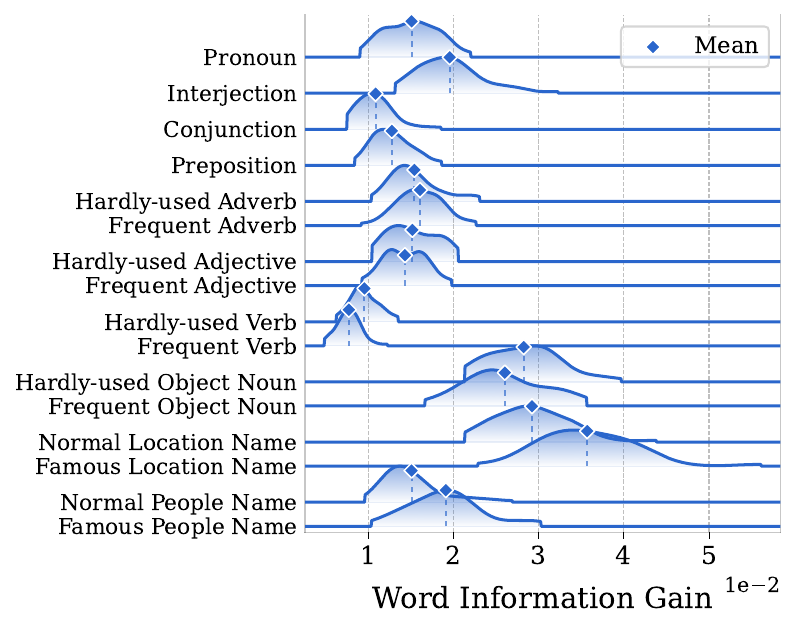}
    \vspace{-2\baselineskip}\caption{Experiment result of Fig.~\ref{fig:static_information} on Llama 2-13B.}
    \end{minipage}\hfill
\vspace{-0.6\baselineskip}\end{figure}

\begin{figure}[t]
    \centering
    \begin{minipage}[t]{0.49\linewidth}
    \includegraphics[width=\linewidth]{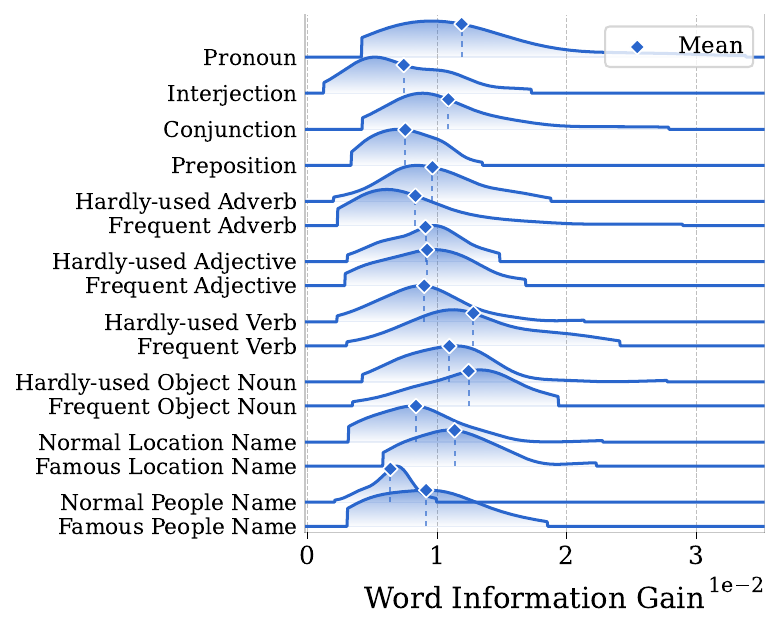}
    \vspace{-2\baselineskip}\caption{Experiment result of Fig.~\ref{fig:static_information} on Granite 4.1-30B.}
    \end{minipage}\hfill
    \begin{minipage}[t]{0.49\linewidth}
    \includegraphics[width=\linewidth]{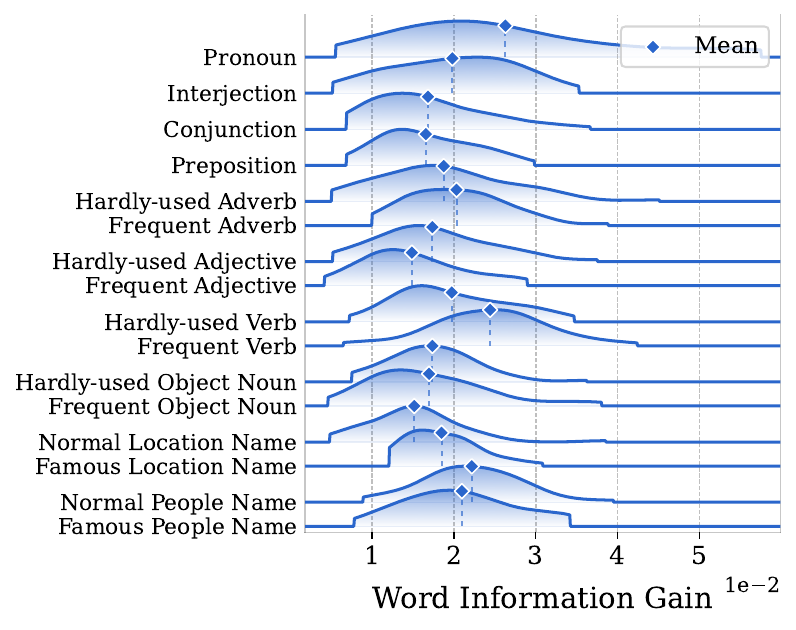}
    \vspace{-2\baselineskip}\caption{Experiment result of Fig.~\ref{fig:static_information} on Llama 3-8B.}
    \end{minipage}
\vspace{-0.6\baselineskip}\end{figure}

\begin{figure}[t]
    \centering
    \begin{minipage}[t]{0.49\linewidth}
    \includegraphics[width=\linewidth]{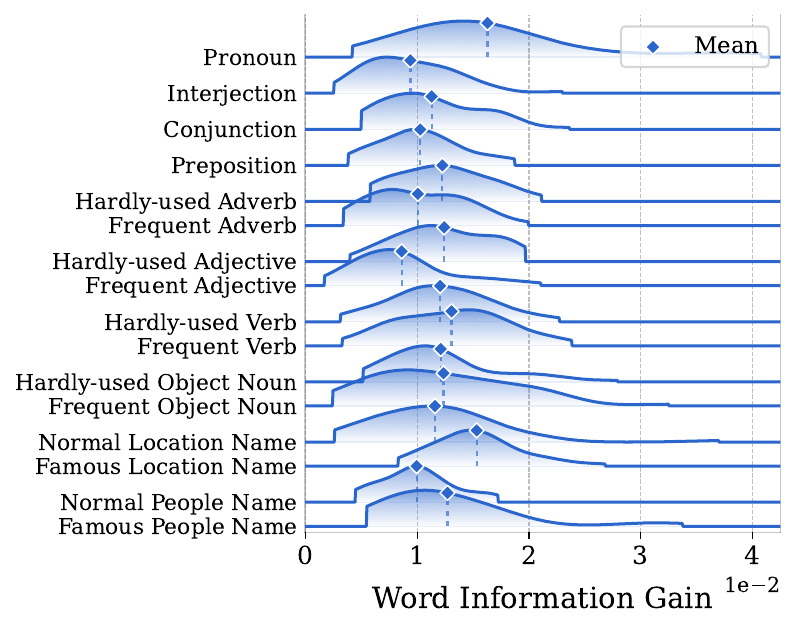}
    \vspace{-2\baselineskip}\caption{Experiment result of Fig.~\ref{fig:static_information} on Olmo 3-32B.}
    \end{minipage}\hfill
    \begin{minipage}[t]{0.49\linewidth}
    \includegraphics[width=\linewidth]{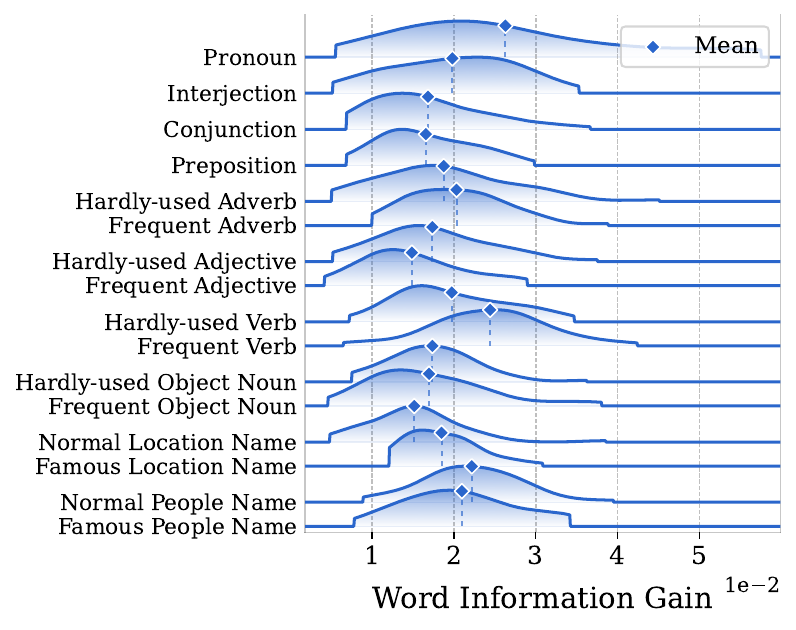}
    \vspace{-2\baselineskip}\caption{Experiment result of Fig.~\ref{fig:static_information} on Qwen 3-8B.}
    \end{minipage}
\vspace{-0.6\baselineskip}\end{figure}

\begin{figure}[t]
    \centering
    \begin{minipage}[t]{0.49\linewidth}
    \includegraphics[width=\linewidth]{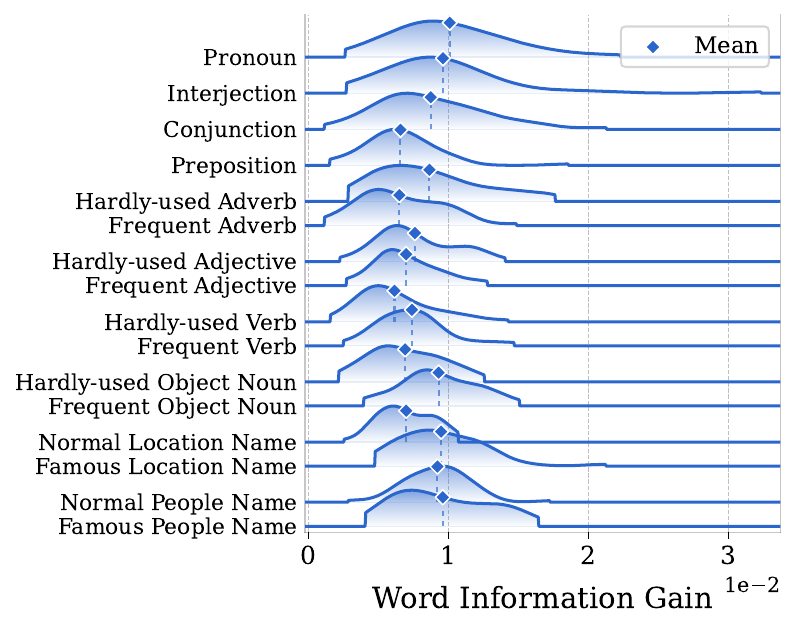}
    \vspace{-2\baselineskip}\caption{Experiment result of Fig.~\ref{fig:static_information} on Qwen 3-14B.}
    \end{minipage}\hfill
    \begin{minipage}[t]{0.49\linewidth}
    \includegraphics[width=\linewidth]{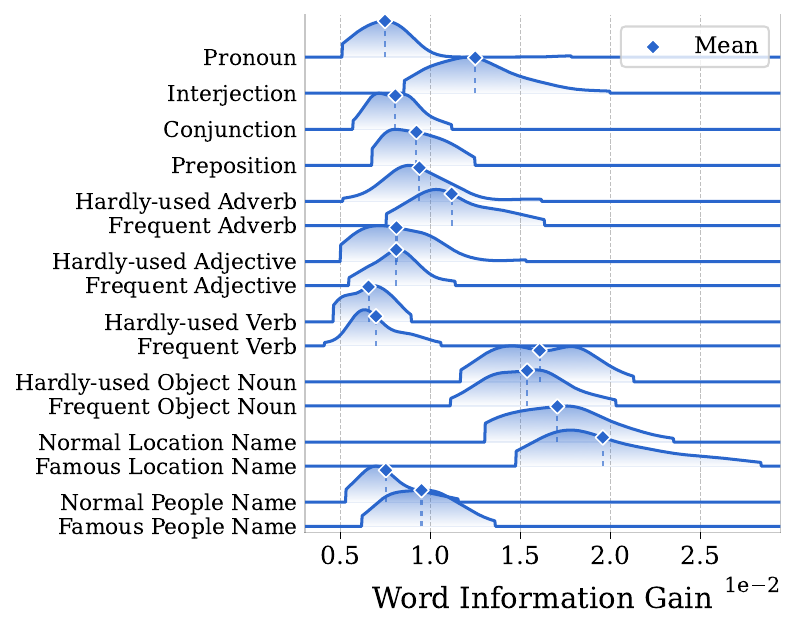}
    \vspace{-2\baselineskip}\caption{Experiment result of Fig.~\ref{fig:static_information} on Qwen 3.6-27B.}
    \end{minipage}
\vspace{-0.6\baselineskip}\end{figure}

\begin{figure}[t]
    \centering
    \begin{minipage}[t]{0.44\linewidth}
    \includegraphics[width=\linewidth]{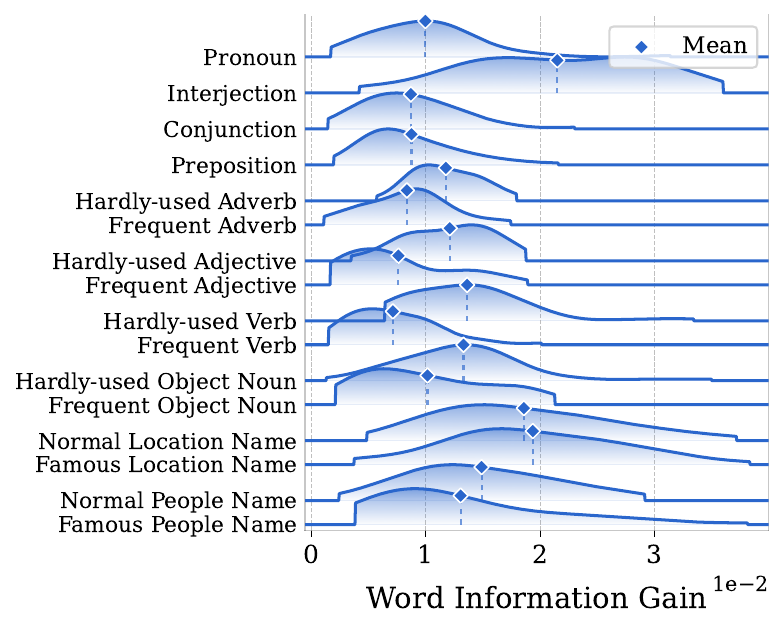}
    \vspace{-2\baselineskip}\caption{Experiment result of Fig.~\ref{fig:static_information} on XLM-RoBERTa-Large.}
    \label{fig:more_exp1_end}
    \end{minipage}\hfill
    \begin{minipage}[t]{0.52\linewidth}
    \includegraphics[width=\linewidth]{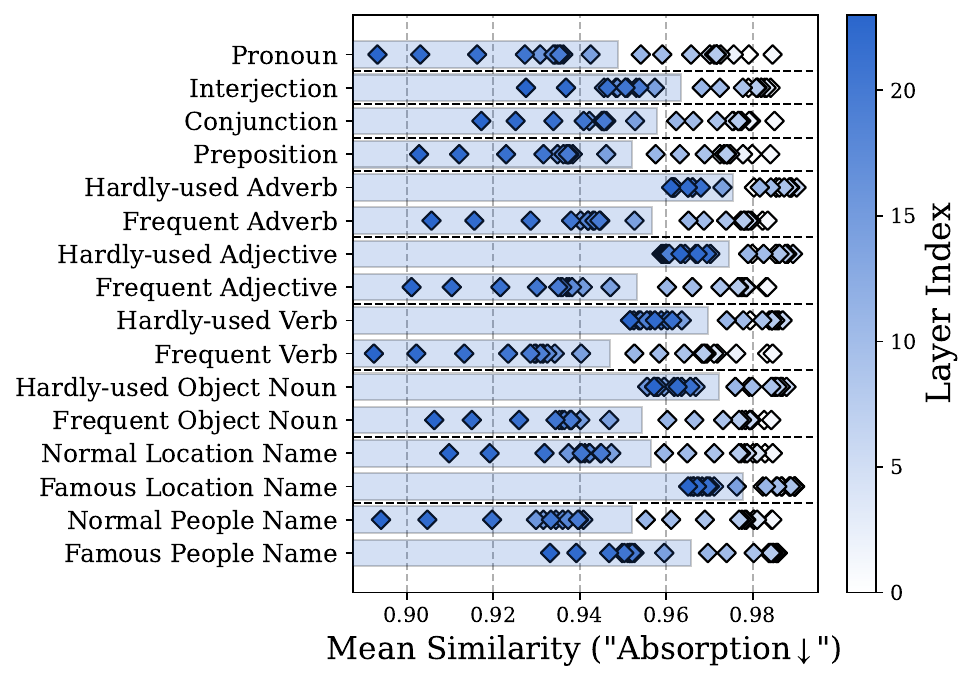}
    \vspace{-2\baselineskip}\caption{Experiment result of Fig.~\ref{fig:donor_and_receptor} on XLM-RoBERTa-Large.}
    \label{fig:more_exp2_begin}
    \end{minipage}
\vspace{-0.6\baselineskip}\end{figure}
\begin{figure}[t]
    \centering
    \begin{minipage}[t]{0.49\linewidth}
    \includegraphics[width=\linewidth]{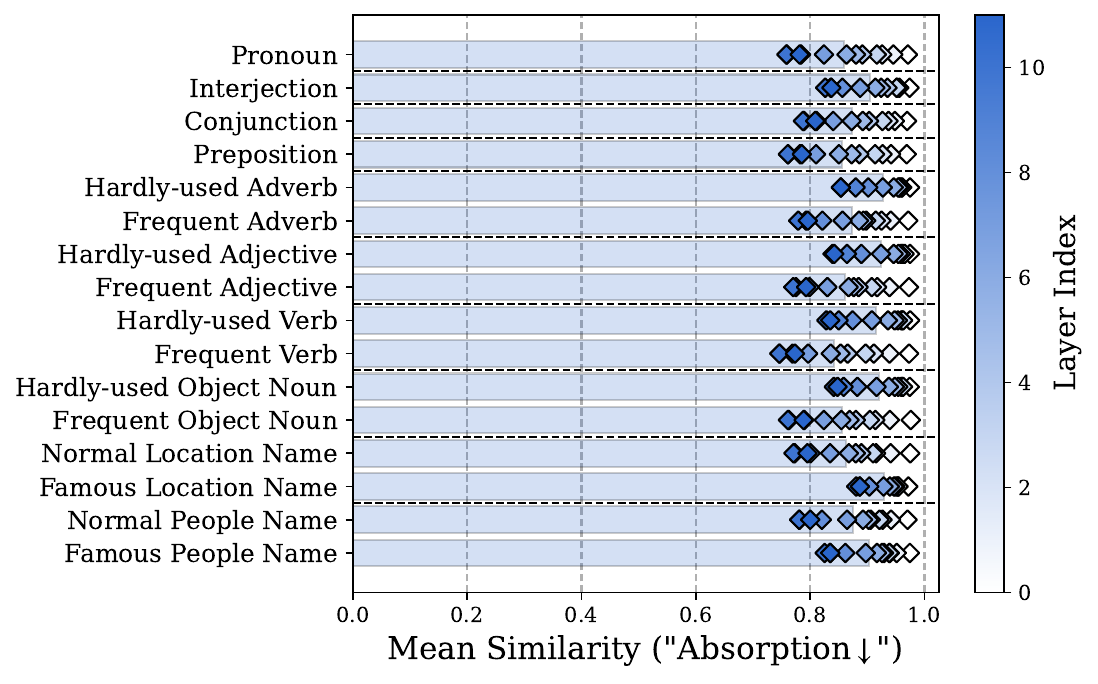}
    \vspace{-2\baselineskip}\caption{Experiment result of Fig.~\ref{fig:donor_and_receptor} on BERT-Base.}
    \end{minipage}\hfill
    \centering
    \begin{minipage}[t]{0.49\linewidth}
    \includegraphics[width=\linewidth]{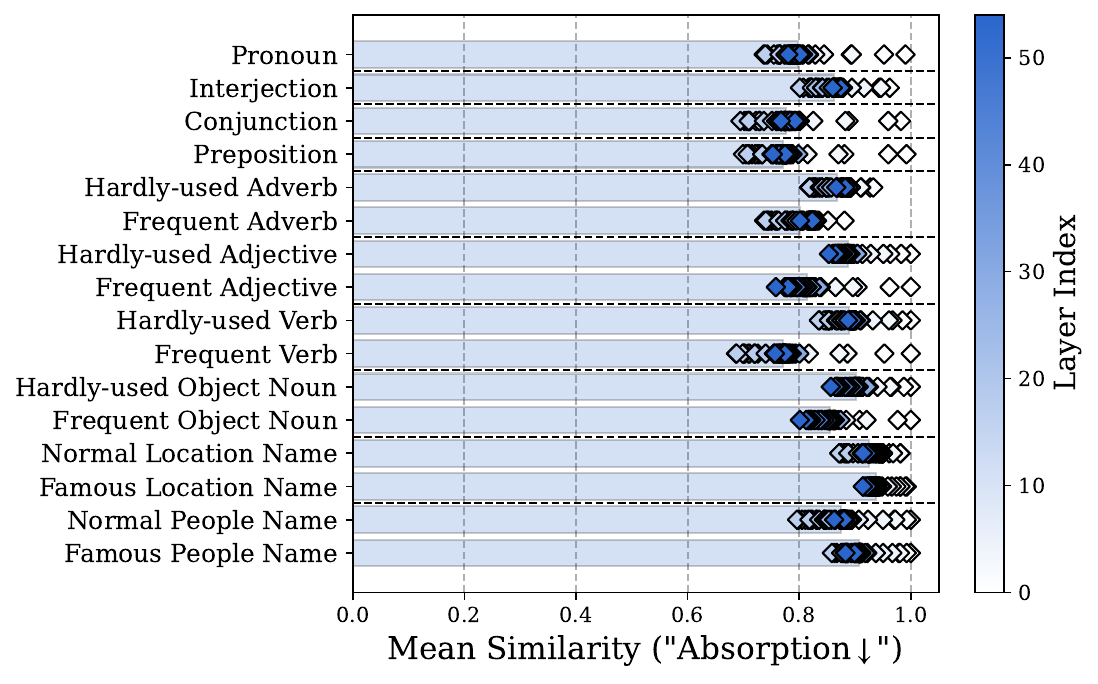}
    \vspace{-2\baselineskip}\caption{Experiment result of Fig.~\ref{fig:donor_and_receptor} on Llama 2-13B.}
    \end{minipage}\hfill
\vspace{-0.6\baselineskip}\end{figure}

\begin{figure}[t]
    \centering
    \begin{minipage}[t]{0.52\linewidth}
    \includegraphics[width=\linewidth]{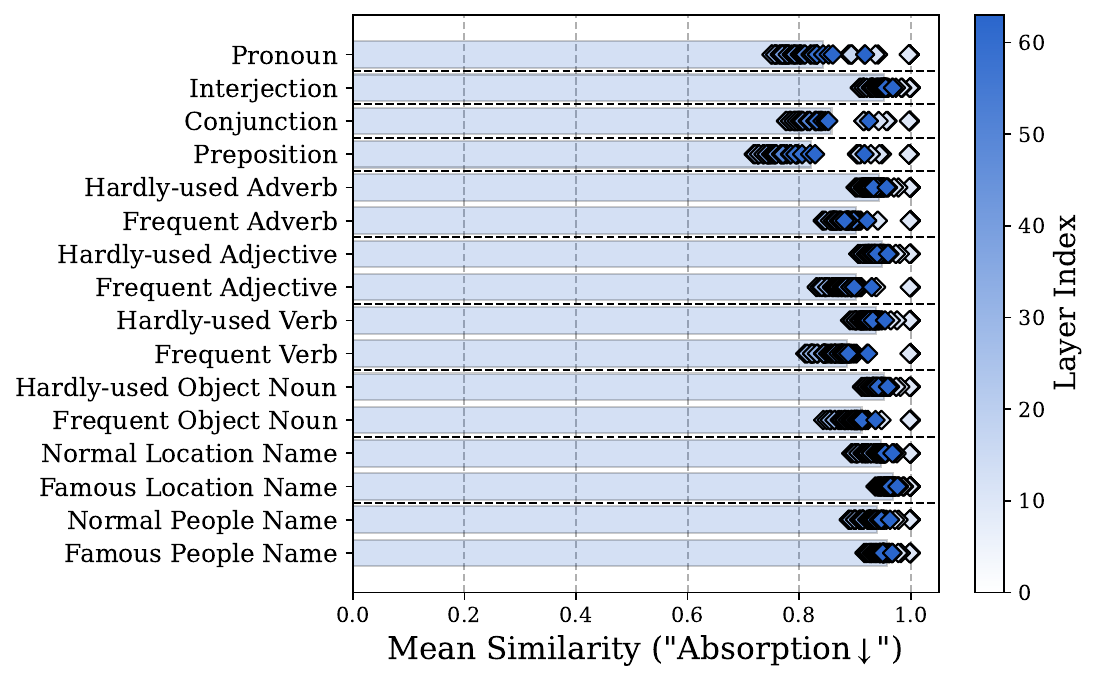}
    \vspace{-2\baselineskip}\caption{Experiment result of Fig.~\ref{fig:donor_and_receptor} on Granite 4.1-30B.}
    \end{minipage}\hfill
    \begin{minipage}[t]{0.46\linewidth}
    \includegraphics[width=\linewidth]{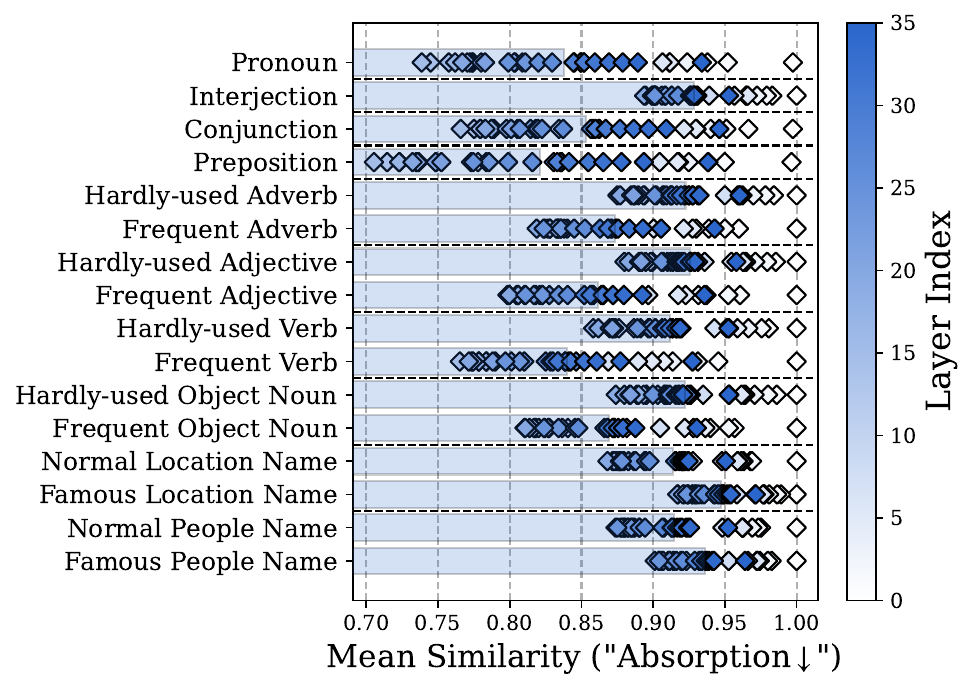}
    \vspace{-2\baselineskip}\caption{Experiment result of Fig.~\ref{fig:donor_and_receptor} on Llama 3-8B.}
    \end{minipage}
\vspace{-0.6\baselineskip}\end{figure}

\begin{figure}[t]
    \centering
    \begin{minipage}[t]{0.49\linewidth}
    \includegraphics[width=\linewidth]{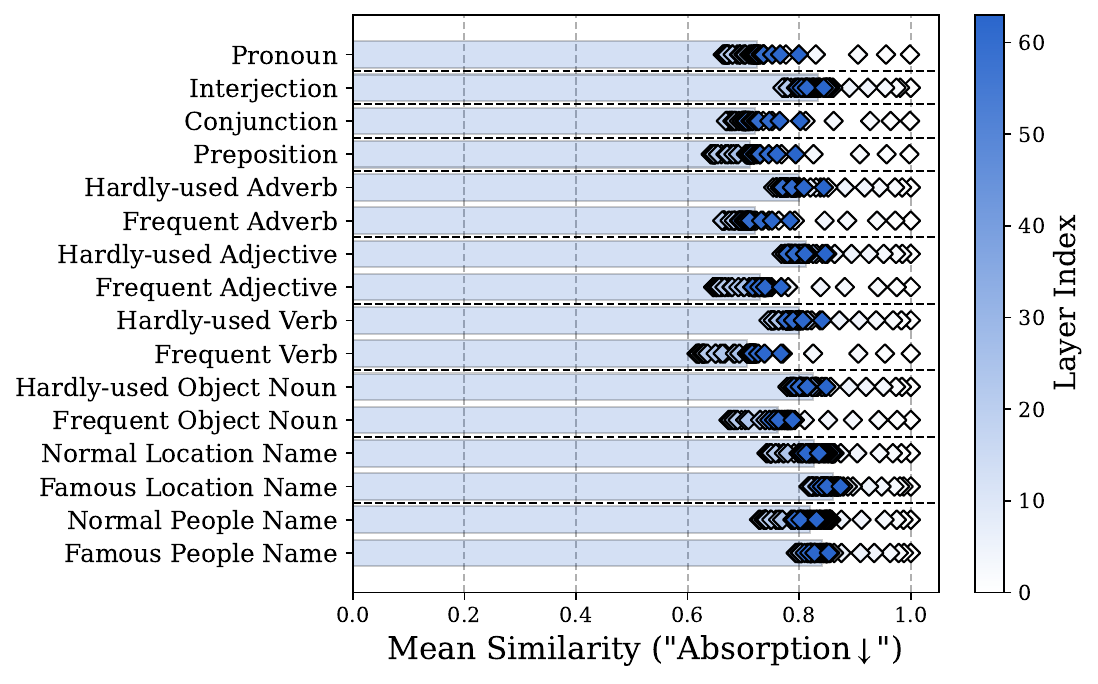}
    \vspace{-2\baselineskip}\caption{Experiment result of Fig.~\ref{fig:donor_and_receptor} on Olmo 3-32B.}
    \end{minipage}\hfill
    \begin{minipage}[t]{0.49\linewidth}
    \includegraphics[width=\linewidth]{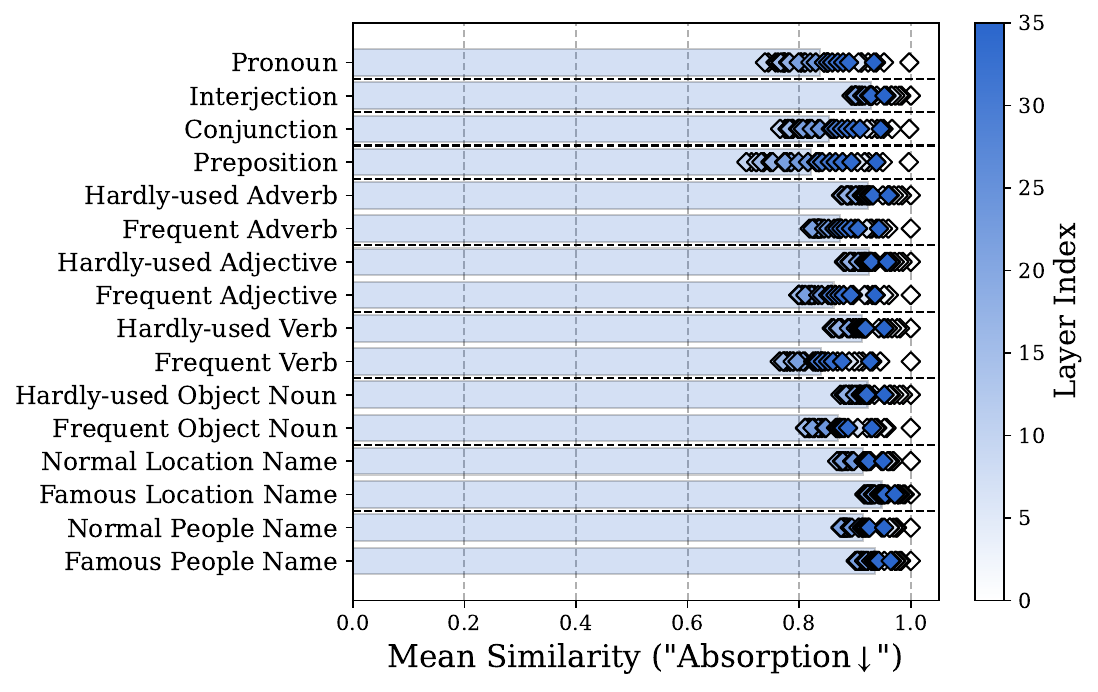}
    \vspace{-2\baselineskip}\caption{Experiment result of Fig.~\ref{fig:donor_and_receptor} on Qwen 3-8B.}
    \end{minipage}
\vspace{-0.6\baselineskip}\end{figure}

\begin{figure}[t]
    \centering
    \begin{minipage}[t]{0.49\linewidth}
    \includegraphics[width=\linewidth]{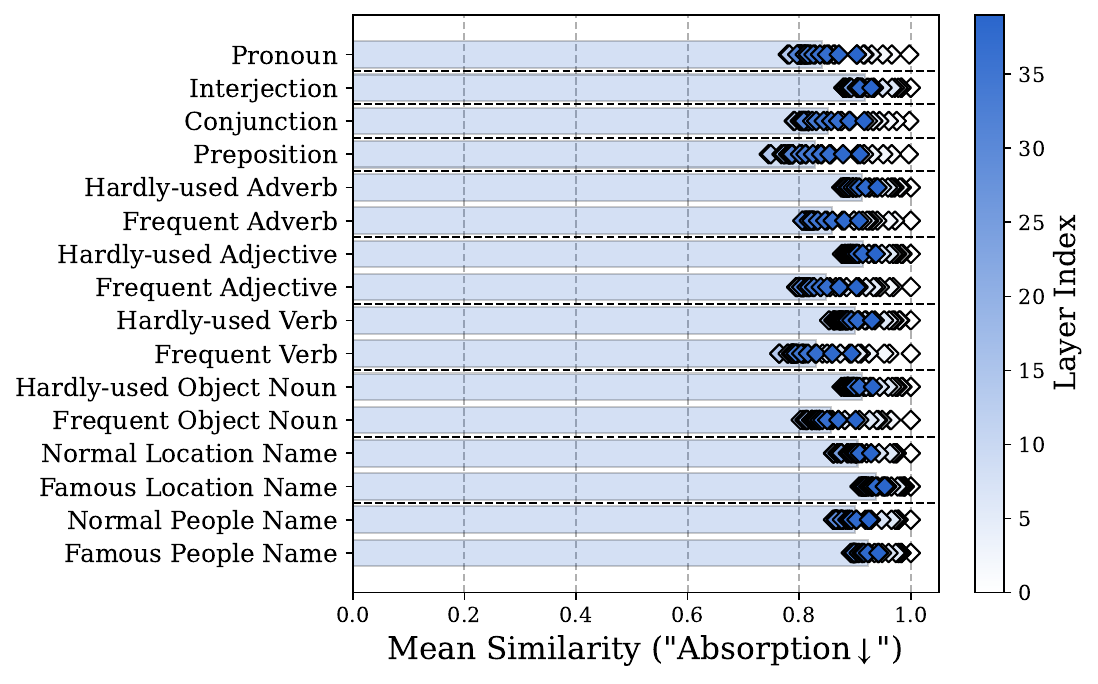}
    \vspace{-2\baselineskip}\caption{Experiment result of Fig.~\ref{fig:donor_and_receptor} on Qwen 3-14B.}
    \end{minipage}\hfill
    \begin{minipage}[t]{0.49\linewidth}
    \includegraphics[width=\linewidth]{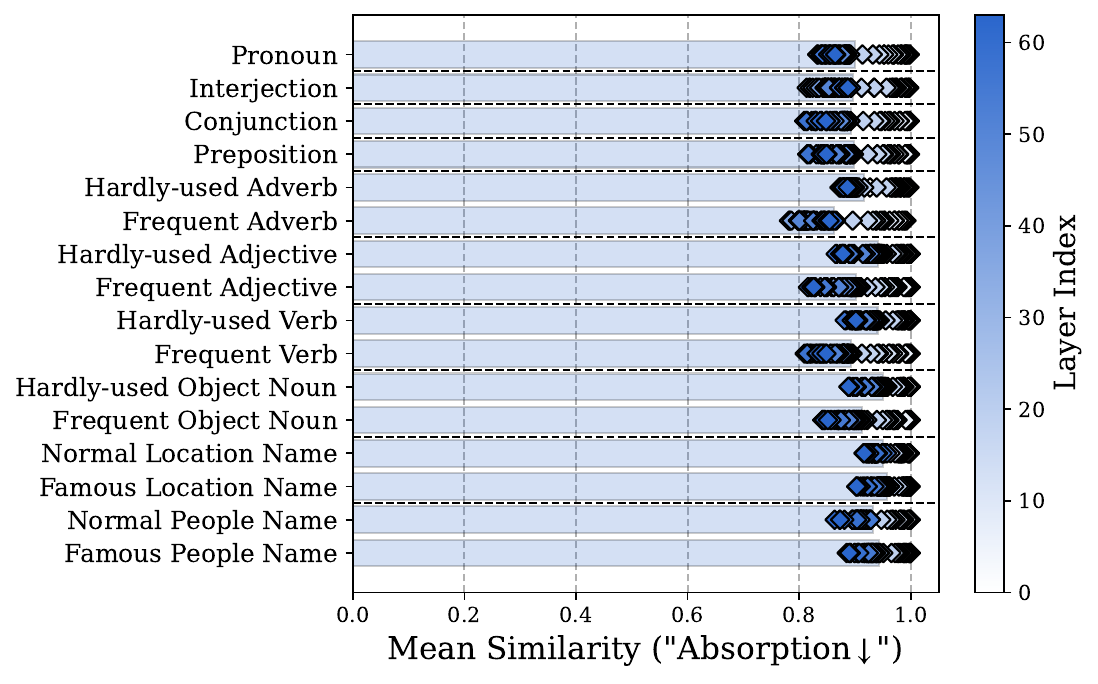}
    \vspace{-2\baselineskip}\caption{Experiment result of Fig.~\ref{fig:donor_and_receptor} on Qwen 3.6-27B.}
    \label{fig:more_exp2_end}
    \end{minipage}
\vspace{-0.6\baselineskip}\end{figure}
\begin{figure}[t]
    \centering
    \begin{minipage}[t]{0.42\linewidth}
    \includegraphics[width=\linewidth]{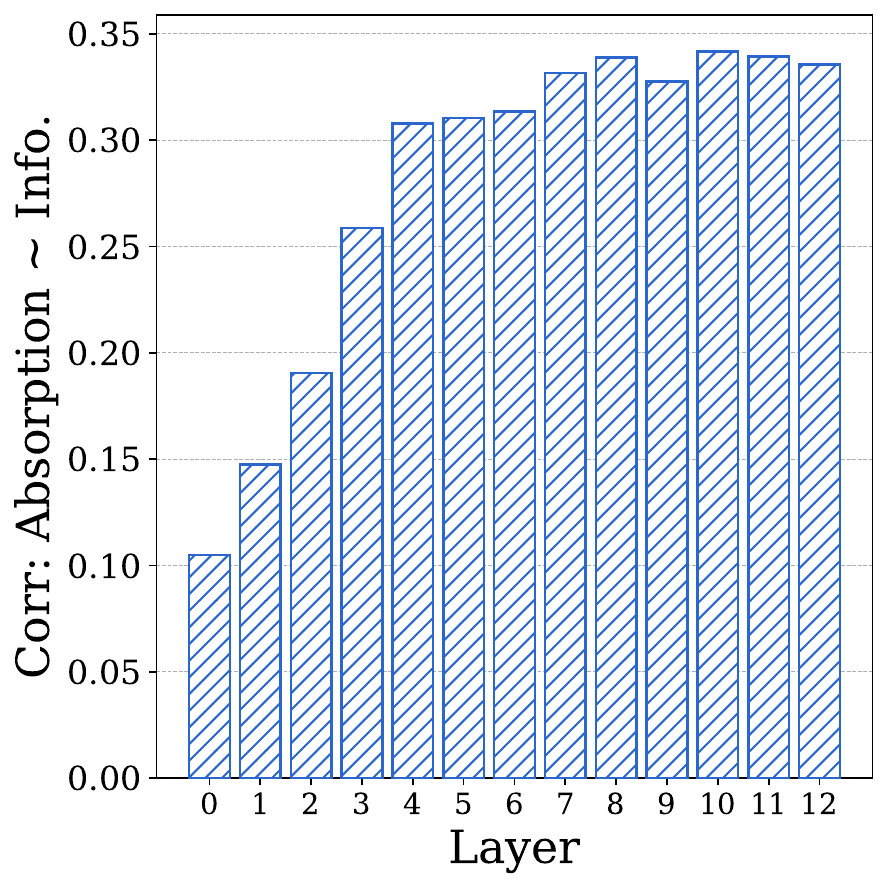}
    \vspace{-2\baselineskip}\caption{Full layer correlation (absorption is a negative measurement) of Fig.~\ref{fig:donor_and_receptor_correlation} on BERT-Base.}
    \label{fig:more_exp3_begin}
    \end{minipage}\hfill
    \begin{minipage}[t]{0.55\linewidth}
    \includegraphics[width=\linewidth]{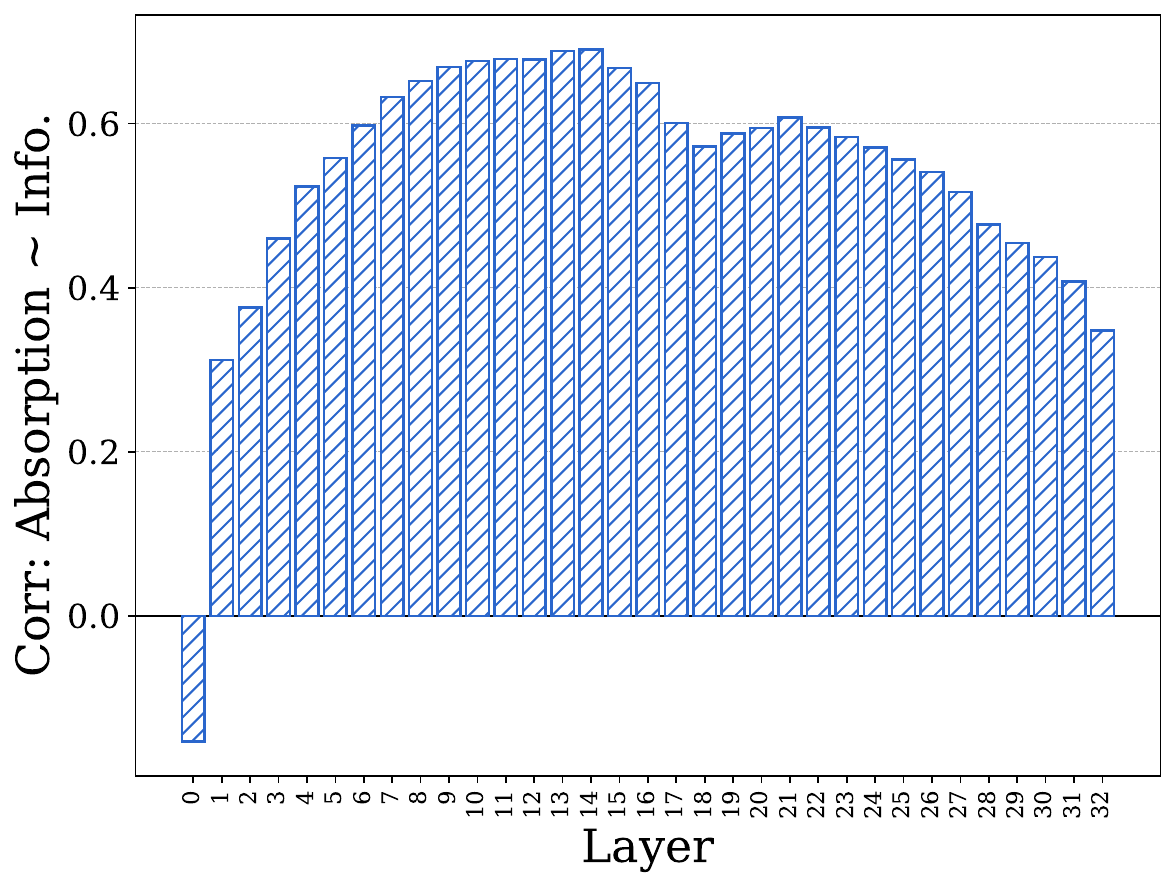}
    \vspace{-2\baselineskip}\caption{Full layer correlation (absorption is a negative measurement) of Fig.~\ref{fig:donor_and_receptor_correlation} on Llama 3-8B.}
    \end{minipage}
\vspace{-0.6\baselineskip}\end{figure}

\begin{figure}[t]
    \centering
    \includegraphics[width=\linewidth]{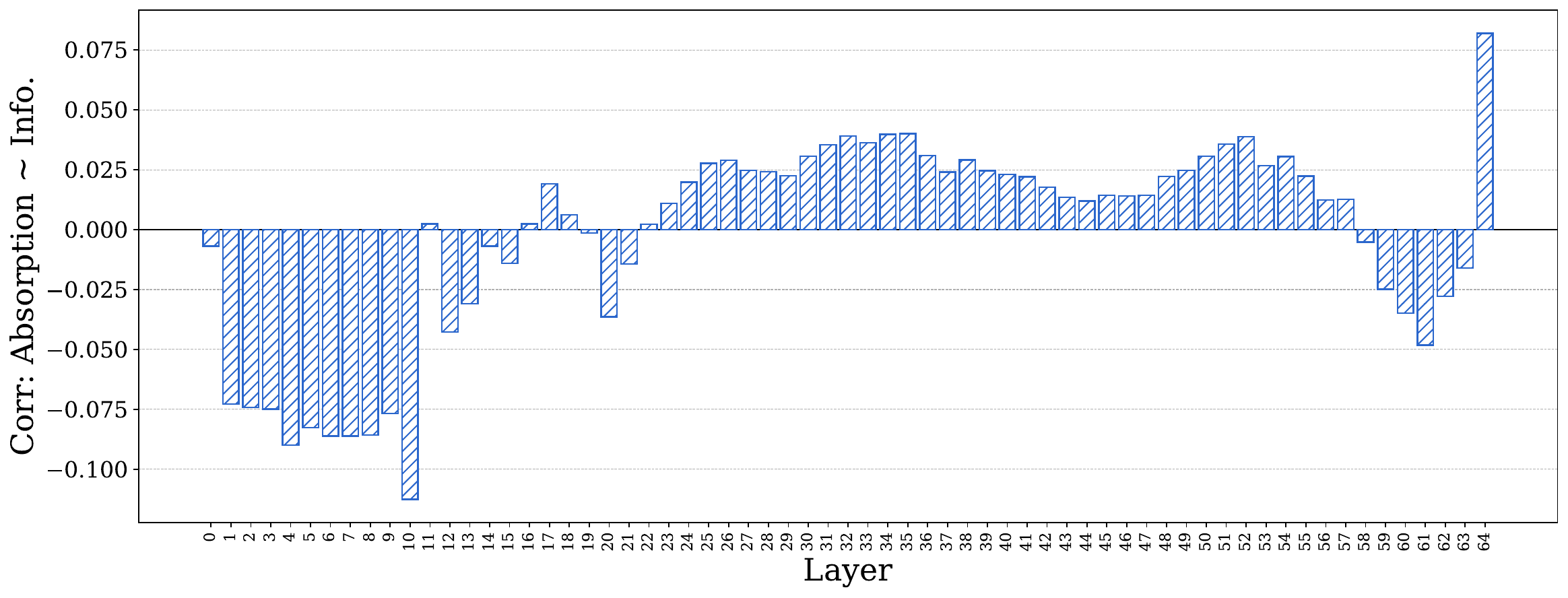}
    \vspace{-2\baselineskip}\caption{Full layer correlation (absorption is a negative measurement) of Fig.~\ref{fig:donor_and_receptor_correlation} on Granite 4.1-30B.}
\vspace{-0.6\baselineskip}\end{figure}

\begin{figure}[t]
    \centering
    \includegraphics[width=\linewidth]{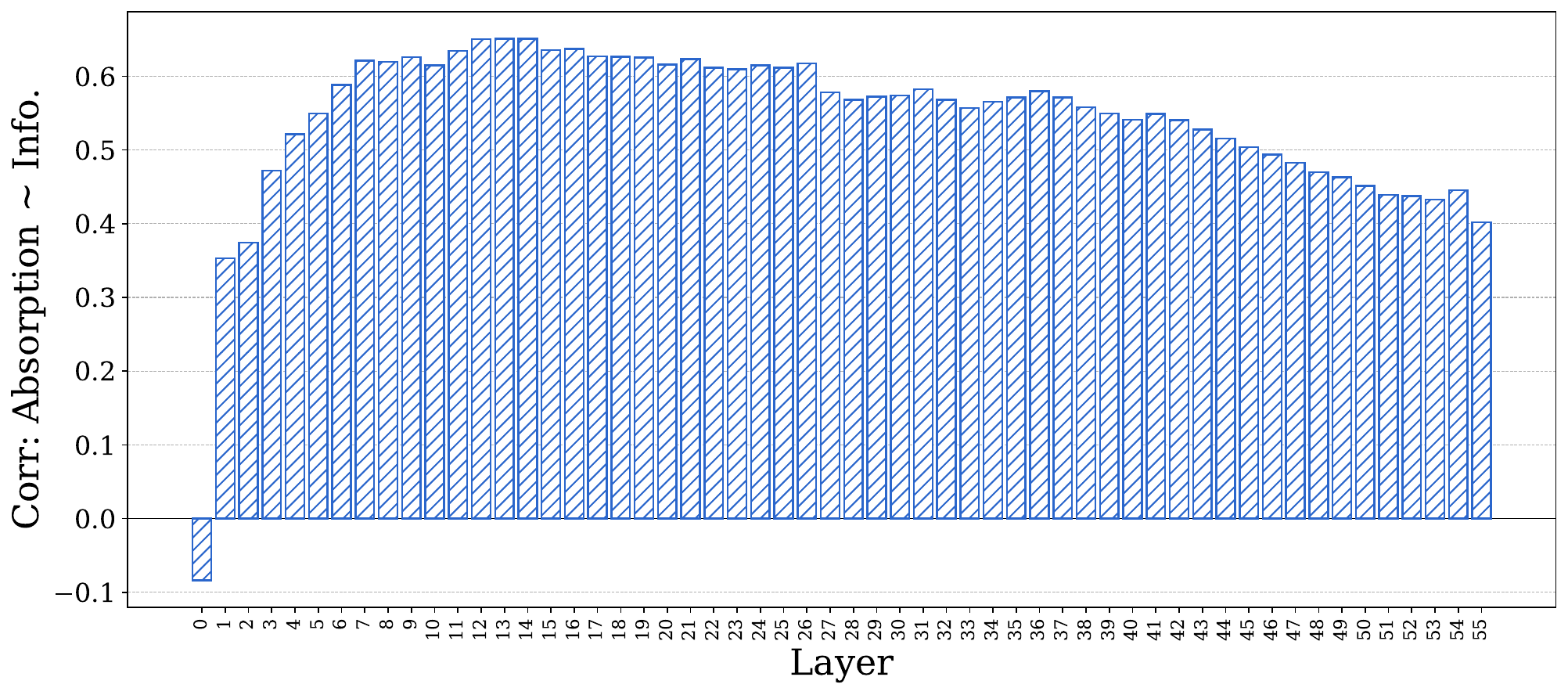}
    \vspace{-2\baselineskip}\caption{Full layer correlation (absorption is a negative measurement) of Fig.~\ref{fig:donor_and_receptor_correlation} on Llama 2-13B.}
\vspace{-0.6\baselineskip}\end{figure}

\begin{figure}[t]
    \centering
    \includegraphics[width=\linewidth]{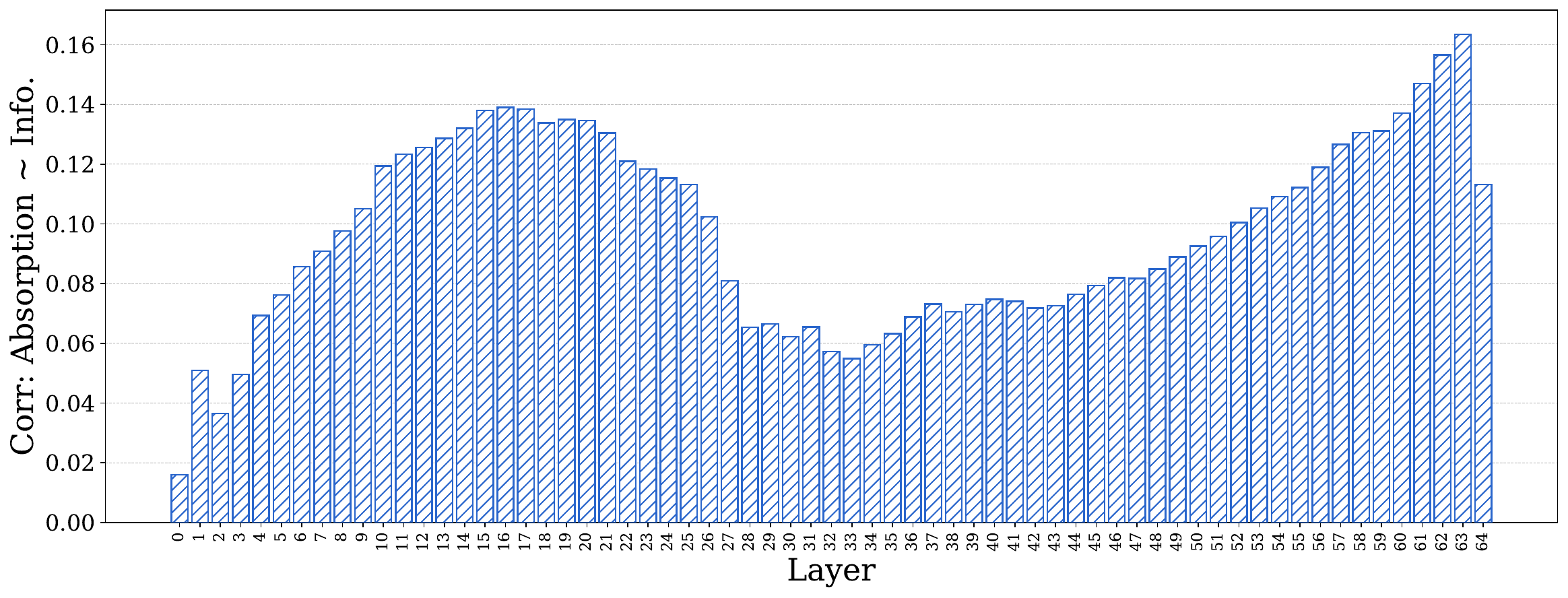}
    \vspace{-2\baselineskip}\caption{Full layer correlation (absorption is a negative measurement) of Fig.~\ref{fig:donor_and_receptor_correlation} on Olmo 3-32B.}
\vspace{-0.6\baselineskip}\end{figure}

\begin{figure}[t]
    \centering
    \begin{minipage}[t]{0.52\linewidth}
    \includegraphics[width=\linewidth]{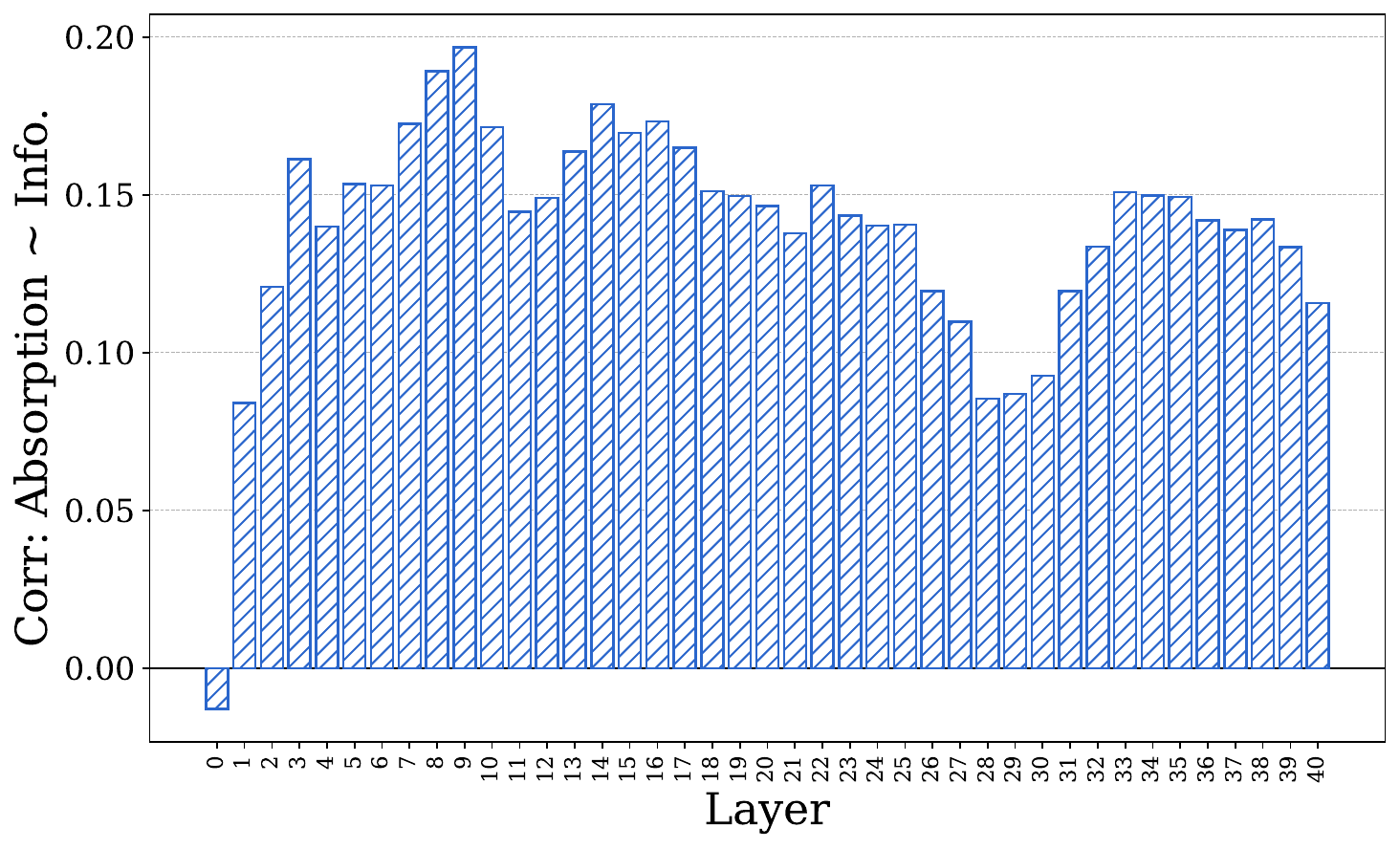}
    \vspace{-2\baselineskip}\caption{Full layer correlation (absorption is a negative measurement) of Fig.~\ref{fig:donor_and_receptor_correlation} on Qwen 3-14B.}
    \end{minipage}\hfill
    \begin{minipage}[t]{0.47\linewidth}
    \includegraphics[width=\linewidth]{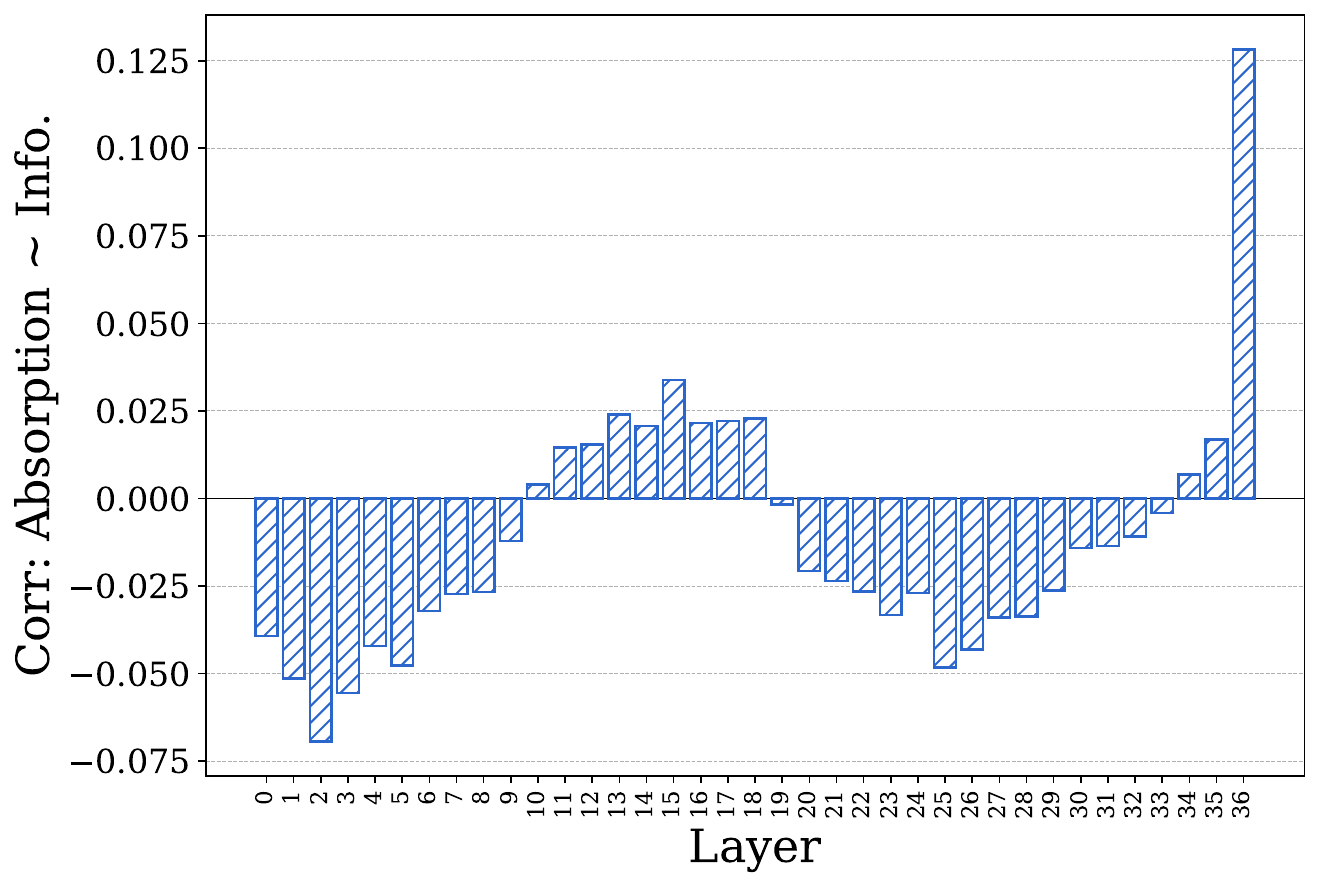}
    \vspace{-2\baselineskip}\caption{Full layer correlation (absorption is a negative measurement) of Fig.~\ref{fig:donor_and_receptor_correlation} on Qwen 3-8B.}
    \end{minipage}
\vspace{-0.6\baselineskip}\end{figure}

\begin{figure}[t]
    \centering
    \includegraphics[width=\linewidth]{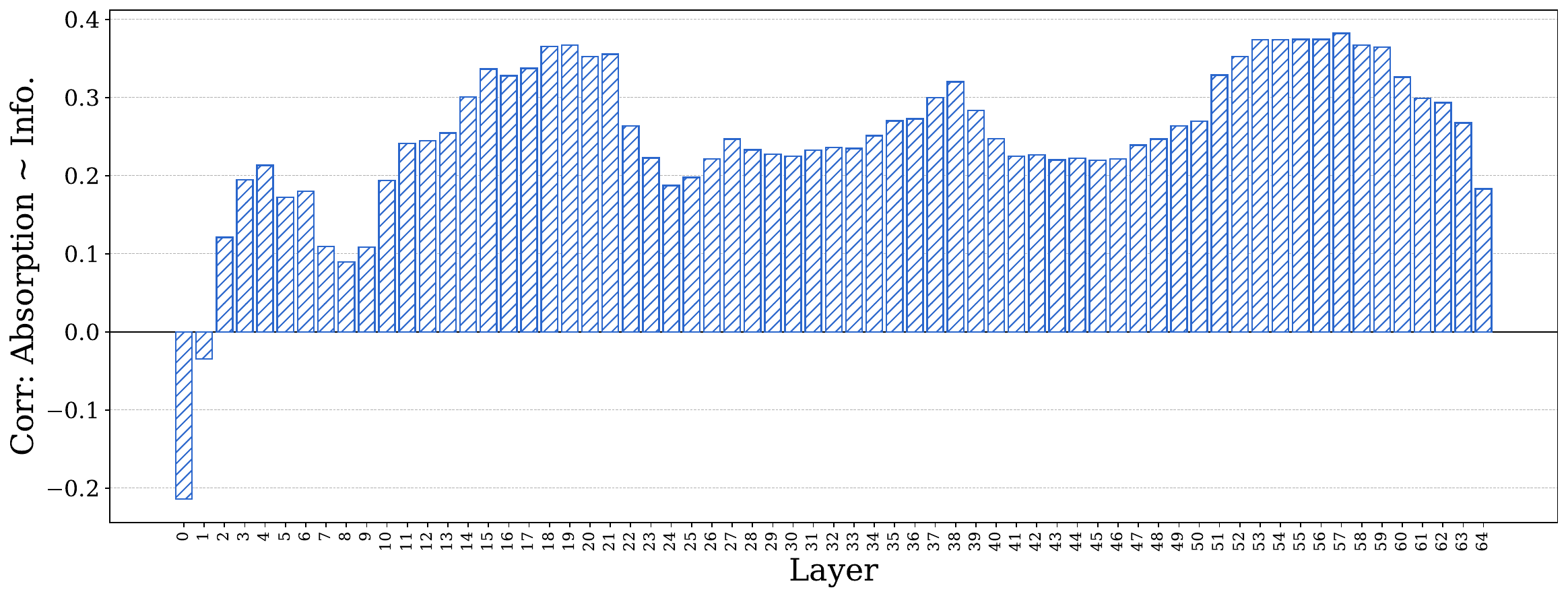}
    \vspace{-2\baselineskip}\caption{Full layer correlation (absorption is a negative measurement) of Fig.~\ref{fig:donor_and_receptor_correlation} on Qwen 3.6-27B.}
\vspace{-0.6\baselineskip}\end{figure}

\begin{figure}[t]
    \centering
    \begin{minipage}[t]{0.49\linewidth}
    \includegraphics[width=\linewidth]{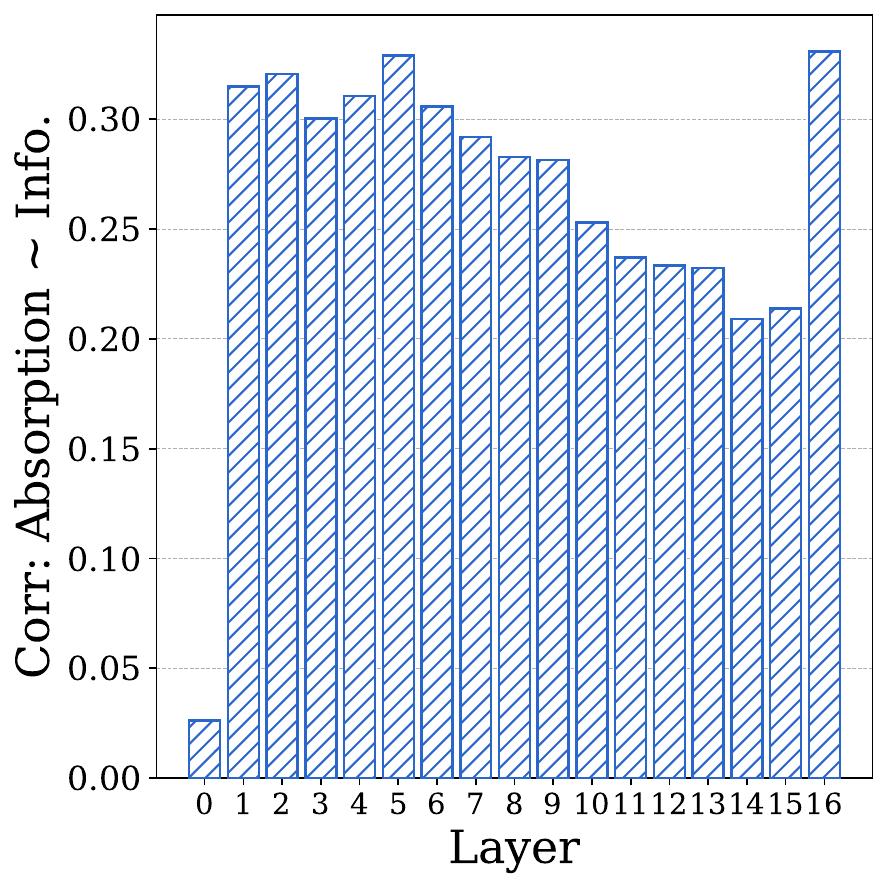}
    \vspace{-2\baselineskip}\caption{Full layer correlation (absorption is a negative measurement) of Fig.~\ref{fig:donor_and_receptor_correlation} on Llama 3.2-1B.}
    \end{minipage}\hfill
    \begin{minipage}[t]{0.49\linewidth}
    \includegraphics[width=\linewidth]{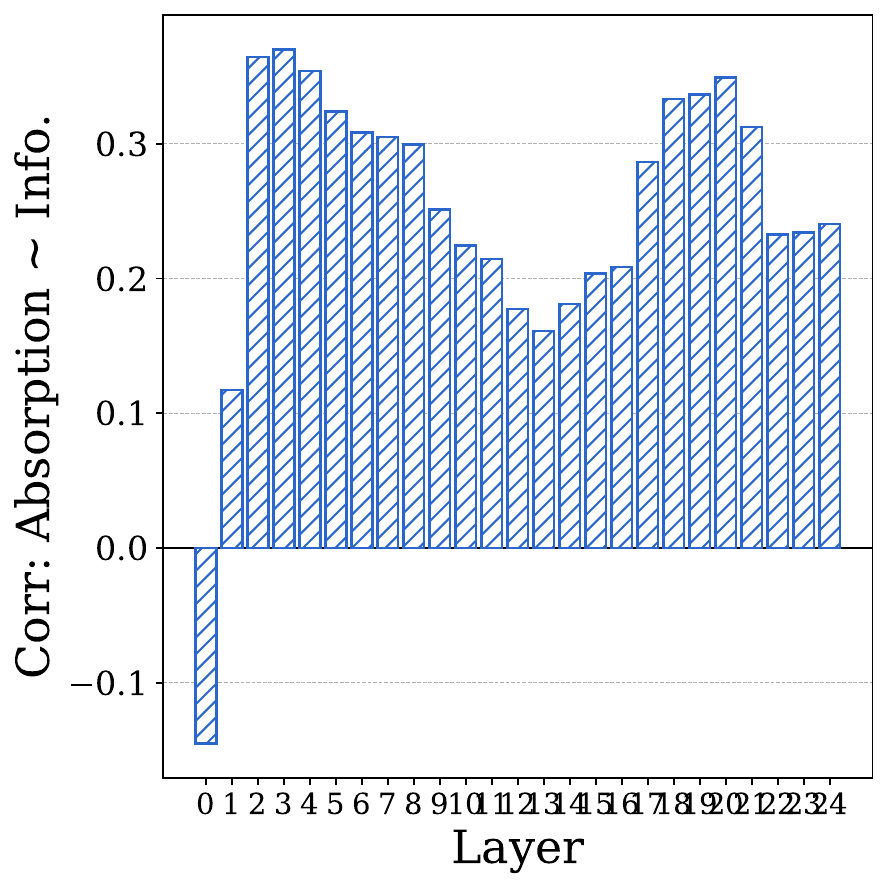}
    \vspace{-2\baselineskip}\caption{Full layer correlation (absorption is a negative measurement) of Fig.~\ref{fig:donor_and_receptor_correlation} on XLM-RoBERTa-Large.}
    \label{fig:more_exp3_end}
    \end{minipage}\hfill
\vspace{-0.6\baselineskip}\end{figure}
\begin{figure}[t]
    \centering
    \begin{minipage}[t]{0.49\linewidth}
    \includegraphics[width=\linewidth]{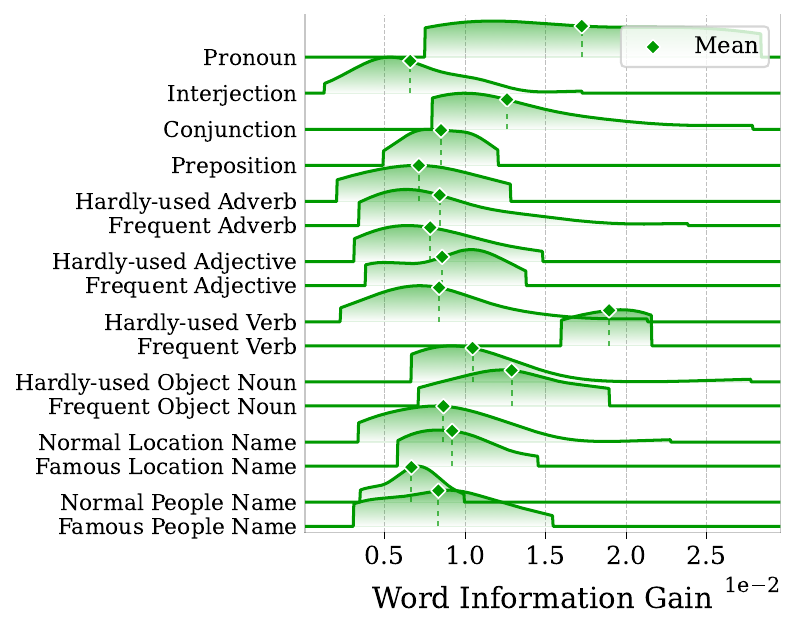}
    \vspace{-2\baselineskip}\caption{Experiment result with token length 2 of Fig.~\ref{fig:static_information} on Granite 4.1-30B.}
    \label{fig:more_exp1_len2_begin}
    \end{minipage}\hfill
    \begin{minipage}[t]{0.49\linewidth}
    \includegraphics[width=\linewidth]{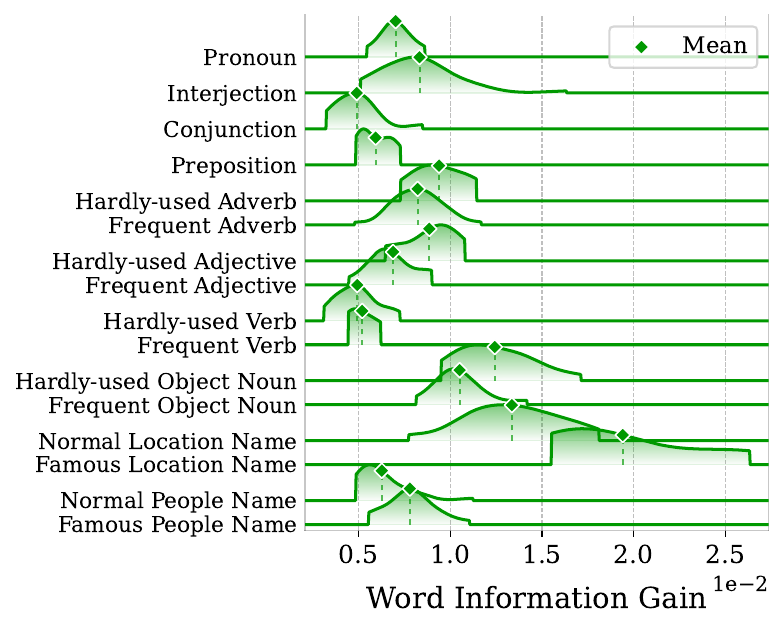}
    \vspace{-2\baselineskip}\caption{Experiment result with token length 2 of Fig.~\ref{fig:static_information} on Llama 3-8B.}
    \end{minipage}
\vspace{-2\baselineskip}\end{figure}

\begin{figure}[t]
    \centering
    \begin{minipage}[t]{0.49\linewidth}
    \includegraphics[width=\linewidth]{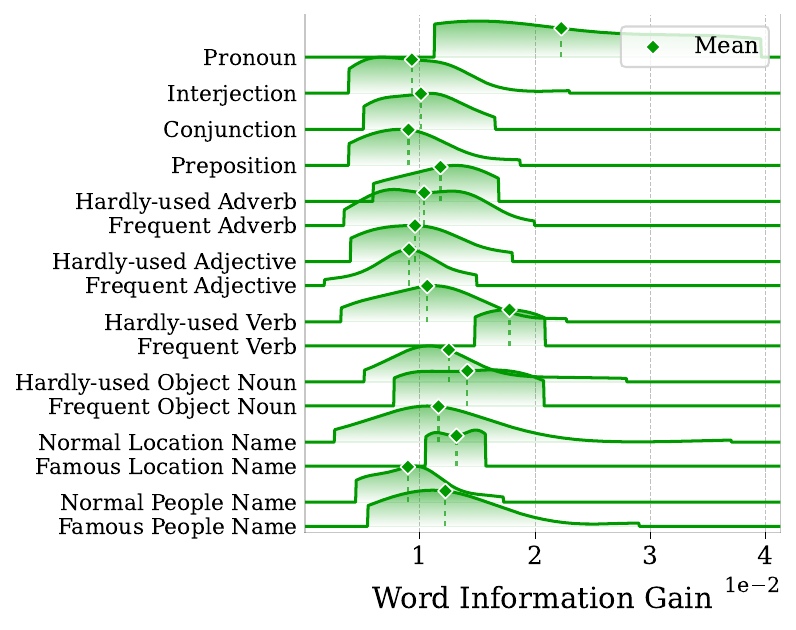}
    \vspace{-2\baselineskip}\caption{Experiment result with token length 2 of Fig.~\ref{fig:static_information} on Olmo 3-32B.}
    \end{minipage}\hfill
    \begin{minipage}[t]{0.49\linewidth}
    \includegraphics[width=\linewidth]{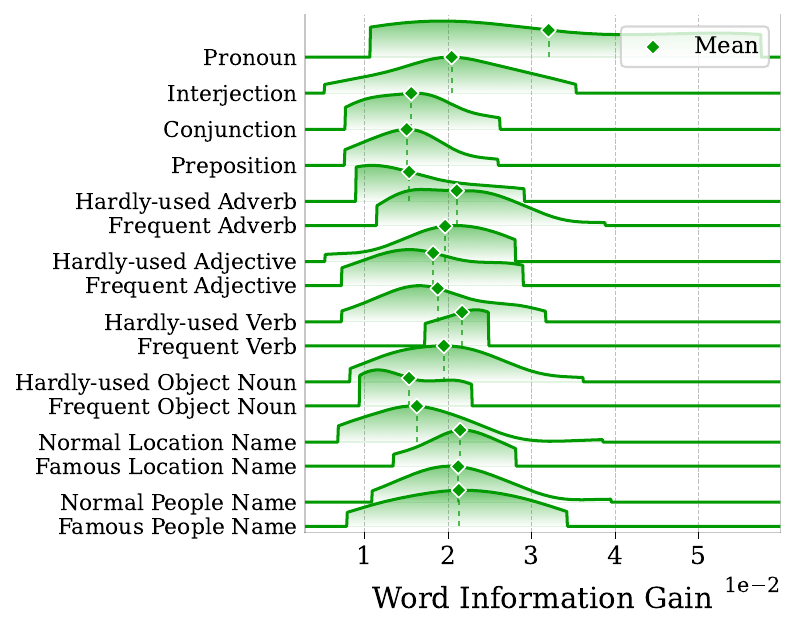}
    \vspace{-2\baselineskip}\caption{Experiment result with token length 2 of Fig.~\ref{fig:static_information} on Qwen 3-8B.}
    \end{minipage}
\vspace{-2\baselineskip}\end{figure}

\begin{figure}[t]
    \centering
    \begin{minipage}[t]{0.49\linewidth}
    \includegraphics[width=\linewidth]{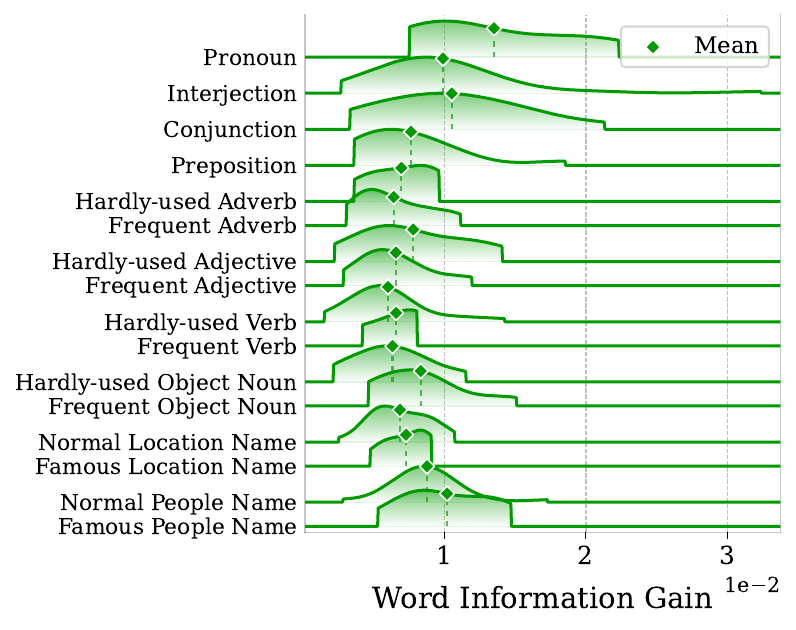}
    \vspace{-2\baselineskip}\caption{Experiment result with token length 2 of Fig.~\ref{fig:static_information} on Qwen 3-14B.}
    \end{minipage}\hfill
    \begin{minipage}[t]{0.49\linewidth}
    \includegraphics[width=\linewidth]{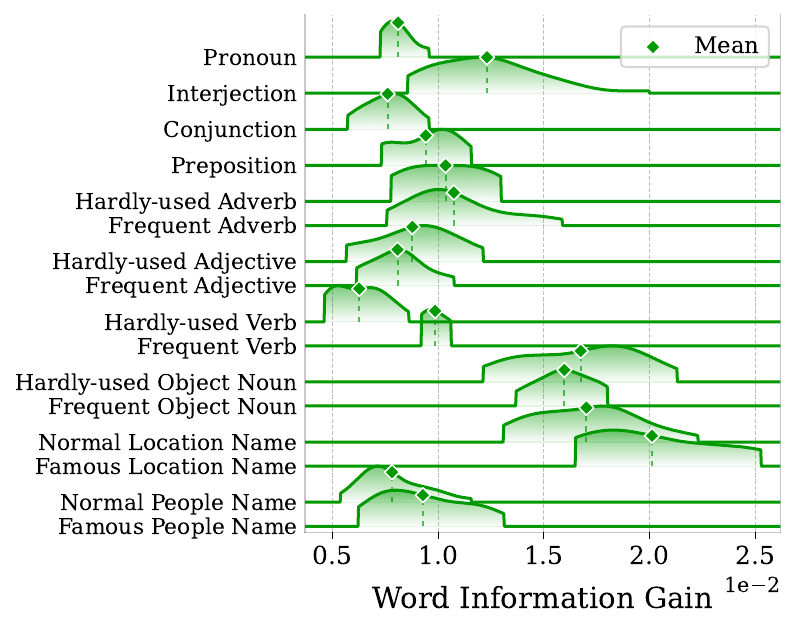}
    \vspace{-2\baselineskip}\caption{Experiment result with token length 2 of Fig.~\ref{fig:static_information} on Qwen 3.6-27B.}
    \end{minipage}
\vspace{-2\baselineskip}\end{figure}

\begin{figure}[t]
    \centering
    \begin{minipage}[t]{0.49\linewidth}
    \includegraphics[width=\linewidth]{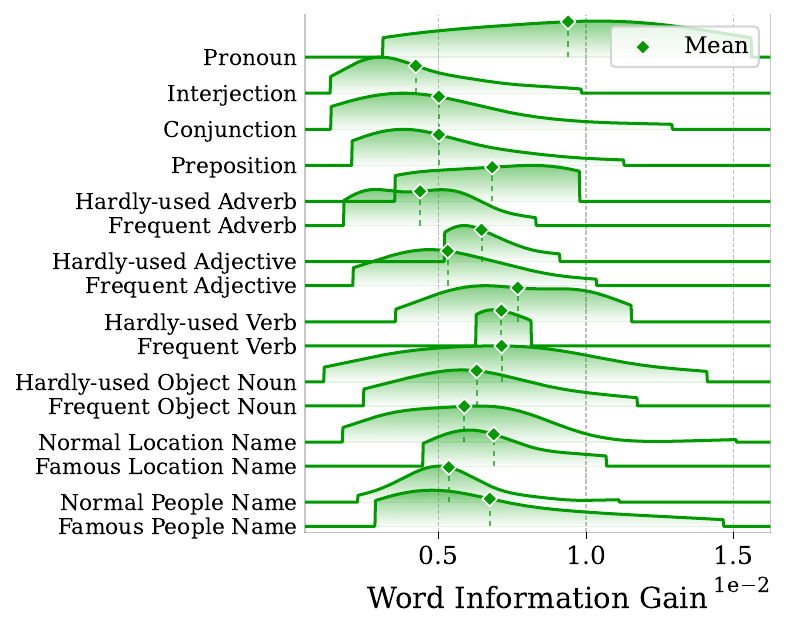}
    \vspace{-2\baselineskip}\caption{Experiment result with token length 2 of Fig.~\ref{fig:static_information} on Llama 3.2-1B.}
    \end{minipage}\hfill
    \begin{minipage}[t]{0.49\linewidth}
    \includegraphics[width=\linewidth]{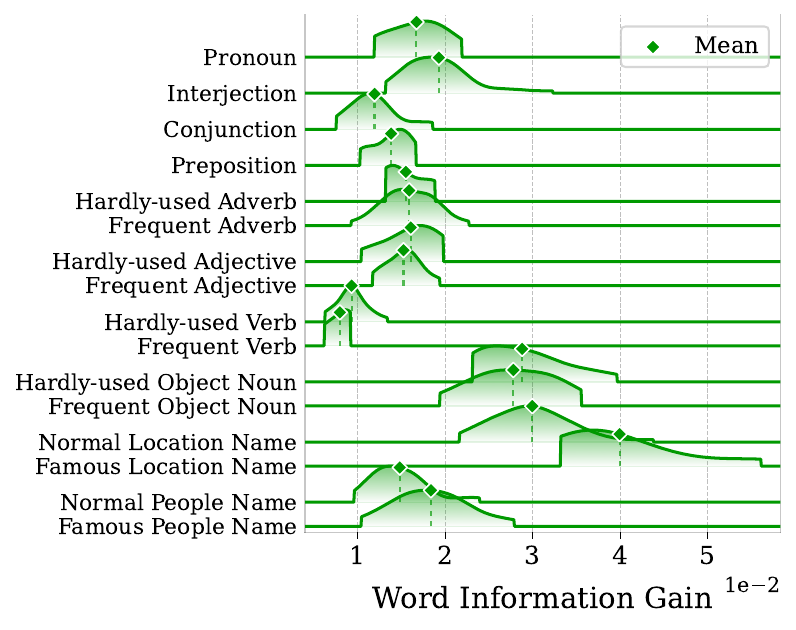}
    \vspace{-2\baselineskip}\caption{Experiment result with token length 2 of Fig.~\ref{fig:static_information} on Llama 2-13B.}
    \label{fig:more_exp1_len2_end}
    \end{minipage}\hfill
\vspace{-2\baselineskip}\end{figure}
\begin{figure}[t]
    \centering
    \begin{minipage}[t]{0.49\linewidth}
    \includegraphics[width=\linewidth]{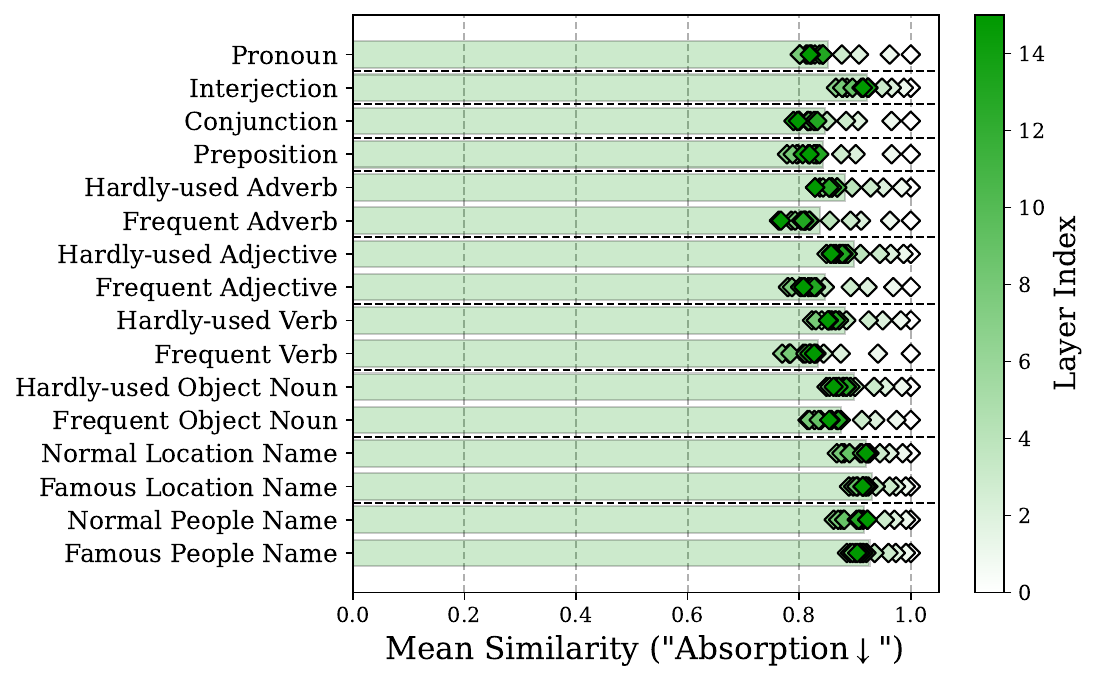}
    \vspace{-2\baselineskip}\caption{Experiment result with token length 2 of Fig.~\ref{fig:donor_and_receptor} on Llama 3.2-1B.}
    \label{fig:more_exp2_len2_begin}
    \end{minipage}\hfill
    \centering
    \begin{minipage}[t]{0.49\linewidth}
    \includegraphics[width=\linewidth]{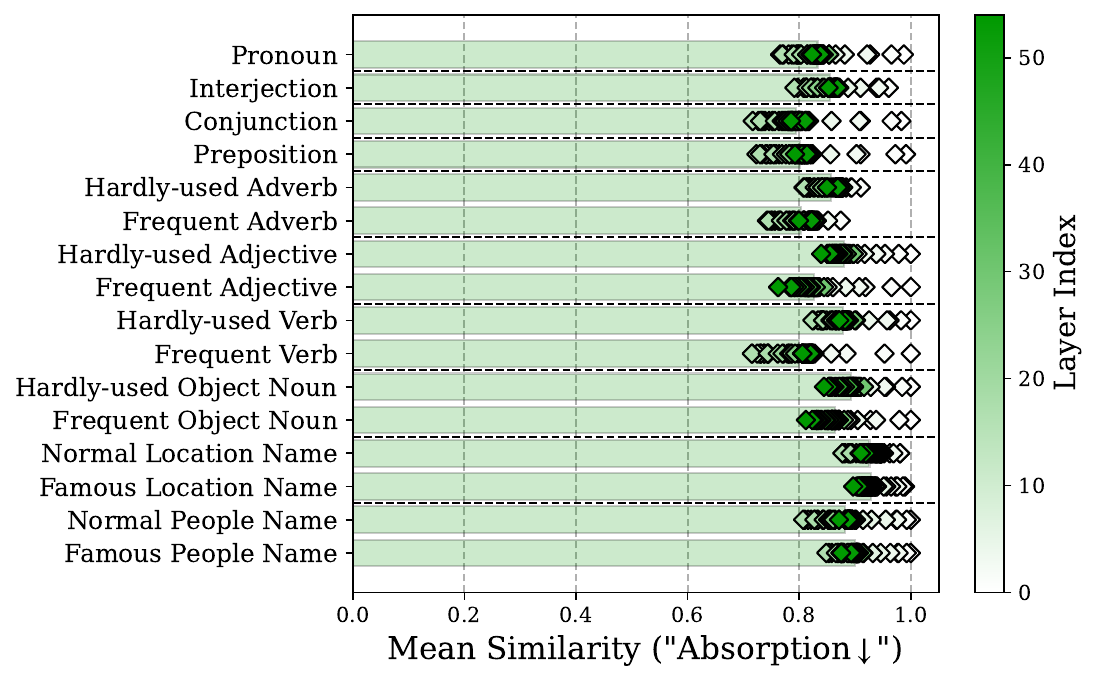}
    \vspace{-2\baselineskip}\caption{Experiment result with token length 2 of Fig.~\ref{fig:donor_and_receptor} on Llama 2-13B.}
    \end{minipage}\hfill
\vspace{-0.6\baselineskip}\end{figure}

\begin{figure}[t]
    \centering
    \begin{minipage}[t]{0.49\linewidth}
    \includegraphics[width=\linewidth]{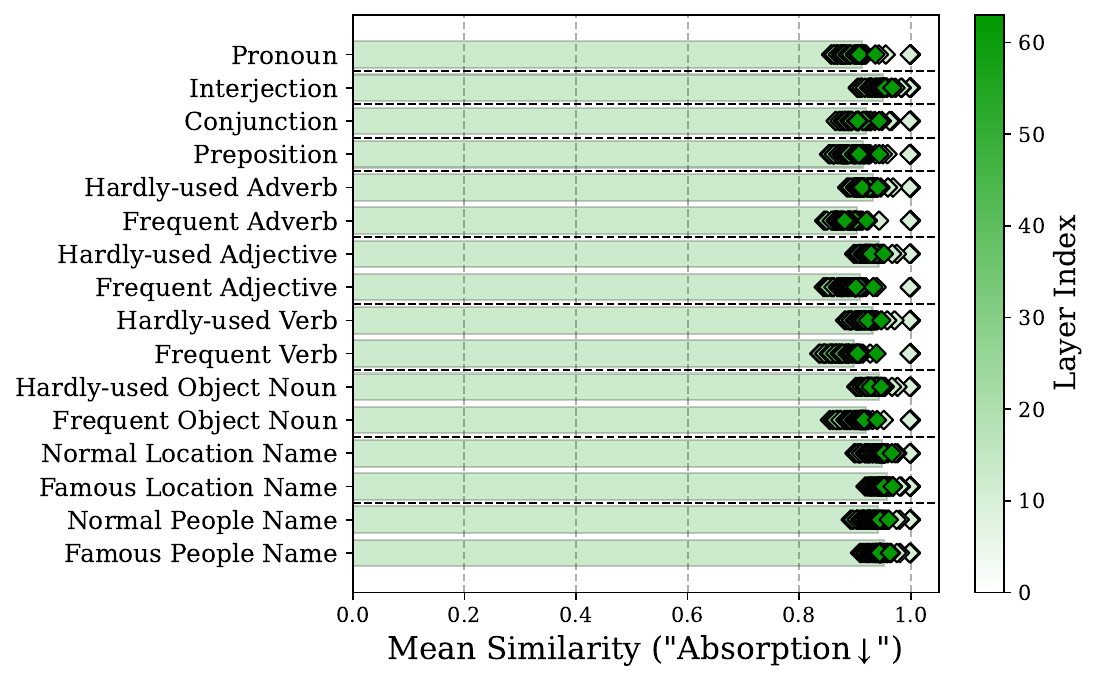}
    \vspace{-2\baselineskip}\caption{Experiment result with token length 2 of Fig.~\ref{fig:donor_and_receptor} on Granite 4.1-30B.}
    \end{minipage}\hfill
    \begin{minipage}[t]{0.49\linewidth}
    \includegraphics[width=\linewidth]{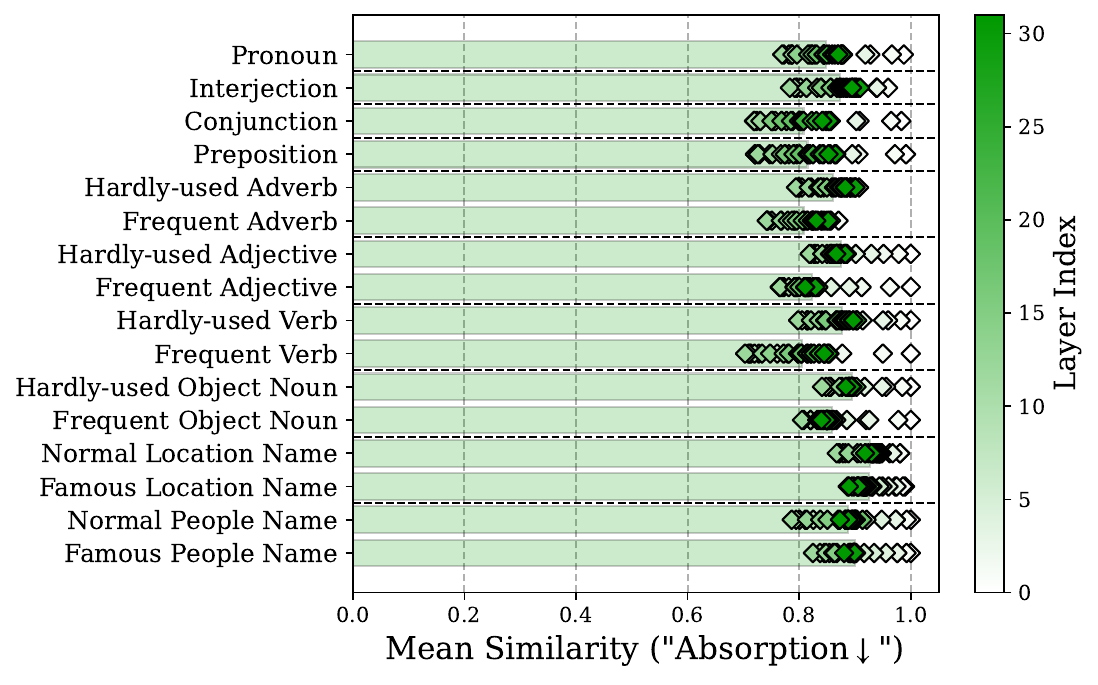}
    \vspace{-2\baselineskip}\caption{Experiment result with token length 2 of Fig.~\ref{fig:donor_and_receptor} on Llama 3-8B.}
    \end{minipage}
\vspace{-0.6\baselineskip}\end{figure}

\begin{figure}[t]
    \centering
    \begin{minipage}[t]{0.49\linewidth}
    \includegraphics[width=\linewidth]{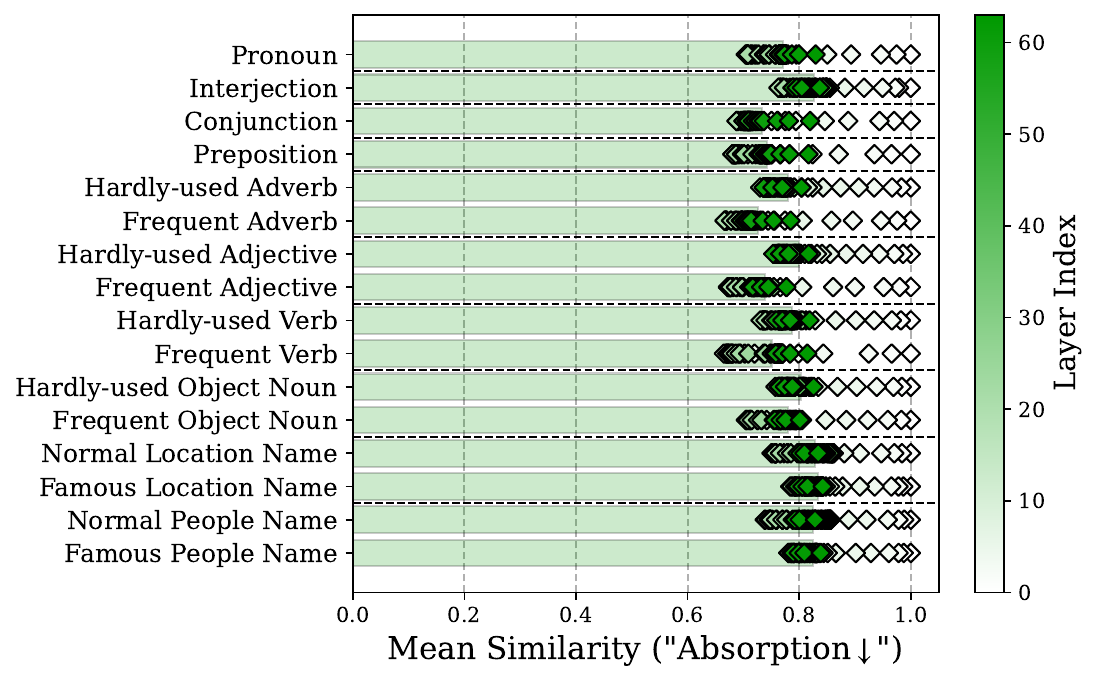}
    \vspace{-2\baselineskip}\caption{Experiment result with token length 2 of Fig.~\ref{fig:donor_and_receptor} on Olmo 3-32B.}
    \end{minipage}\hfill
    \begin{minipage}[t]{0.49\linewidth}
    \includegraphics[width=\linewidth]{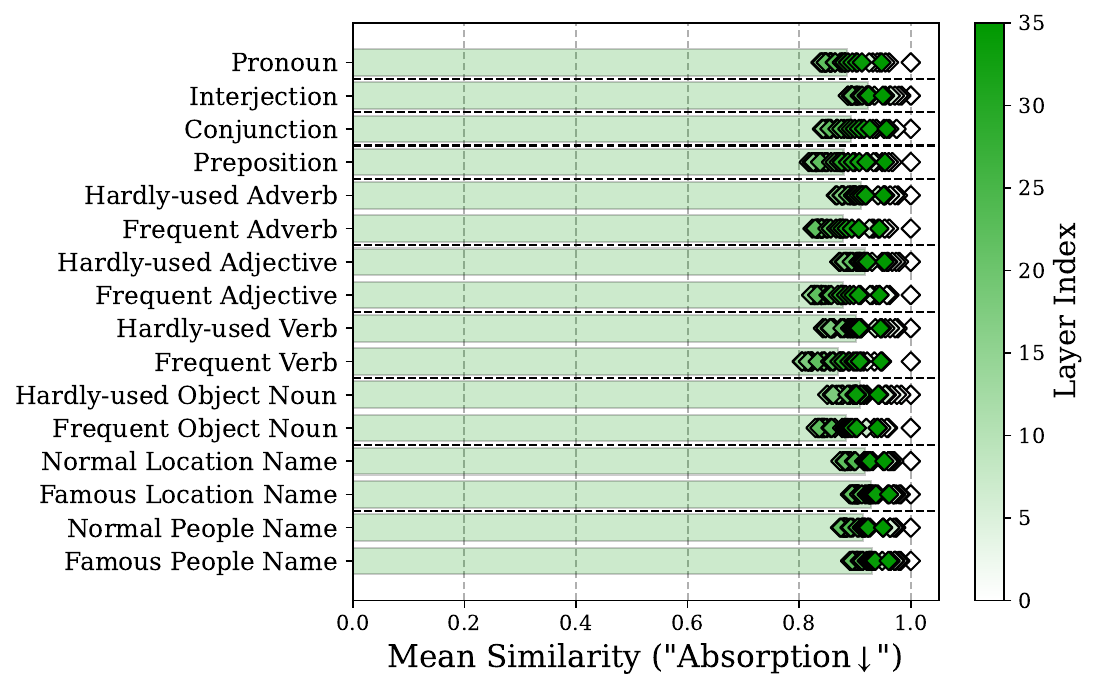}
    \vspace{-2\baselineskip}\caption{Experiment result with token length 2 of Fig.~\ref{fig:donor_and_receptor} on Qwen 3-8B.}
    \end{minipage}
\vspace{-0.6\baselineskip}\end{figure}

\begin{figure}[t]
    \centering
    \begin{minipage}[t]{0.49\linewidth}
    \includegraphics[width=\linewidth]{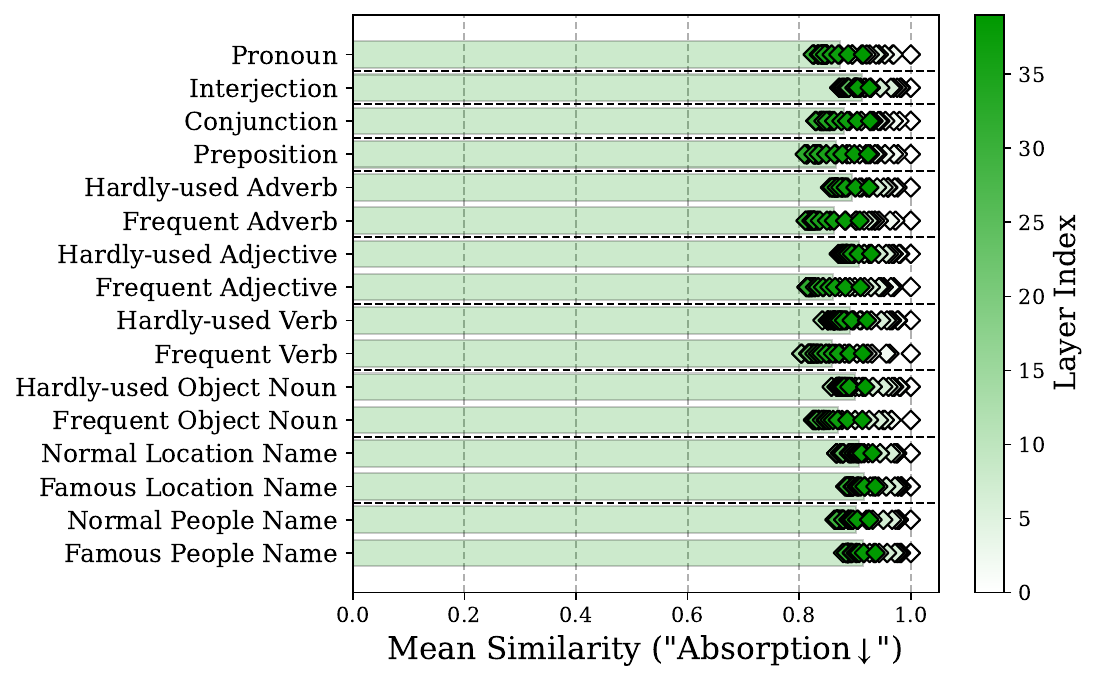}
    \vspace{-2\baselineskip}\caption{Experiment result with token length 2 of Fig.~\ref{fig:donor_and_receptor} on Qwen 3-14B.}
    \end{minipage}\hfill
    \begin{minipage}[t]{0.49\linewidth}
    \includegraphics[width=\linewidth]{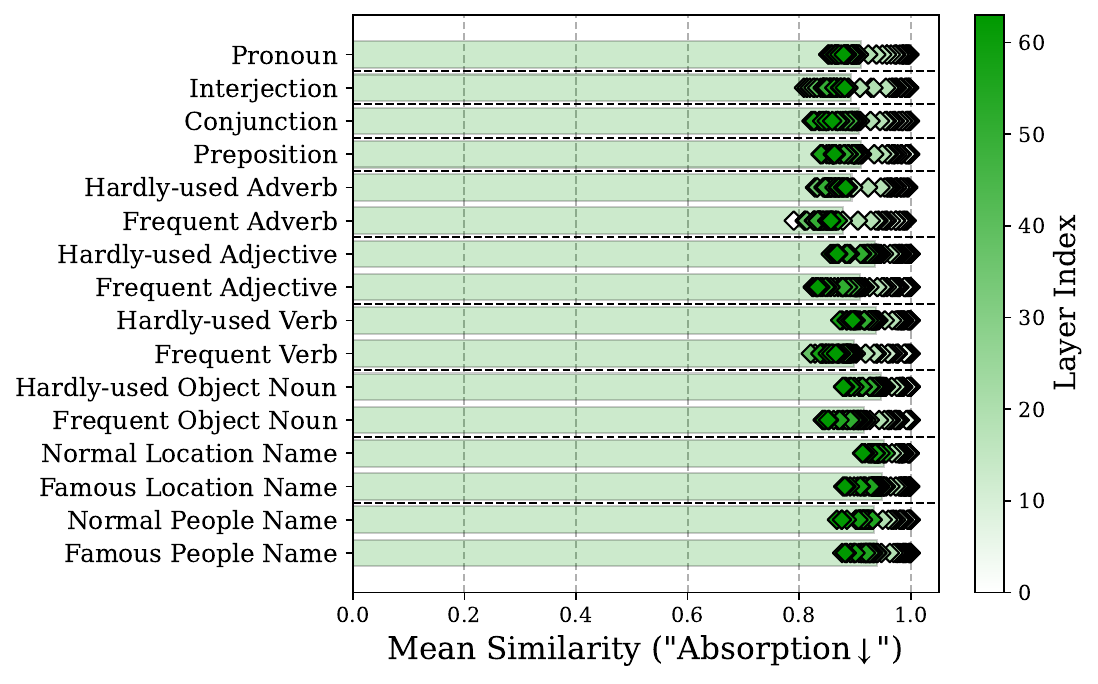}
    \vspace{-2\baselineskip}\caption{Experiment result with token length 2 of Fig.~\ref{fig:donor_and_receptor} on Qwen 3.6-27B.}
    \label{fig:more_exp2_len2_end}
    \end{minipage}
\vspace{-0.6\baselineskip}\end{figure}
\begin{figure}[t]
    \centering
    \begin{minipage}[t]{0.415\linewidth}
    \includegraphics[width=\linewidth]{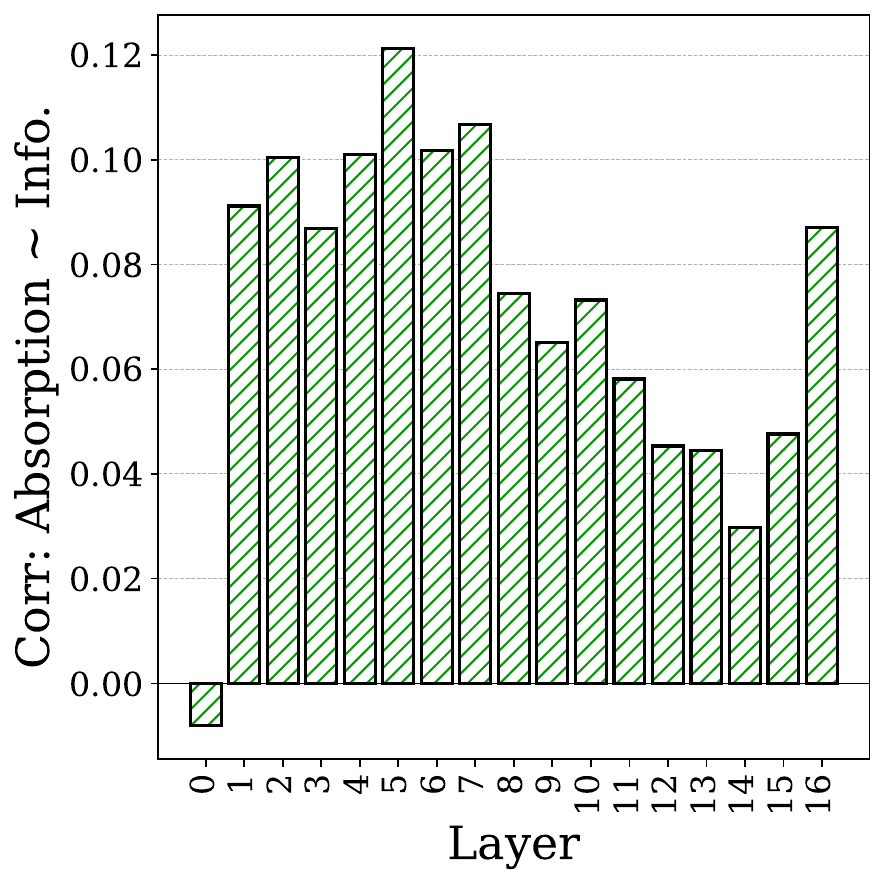}
    \vspace{-2\baselineskip}\caption{Full layer correlation with token length 2 (absorption is a negative measurement) of Fig.~\ref{fig:donor_and_receptor_correlation} on Llama 3.2-1B.}
    \label{fig:more_exp3_len2_begin}
    \end{minipage}\hfill
    \begin{minipage}[t]{0.56\linewidth}
    \includegraphics[width=\linewidth]{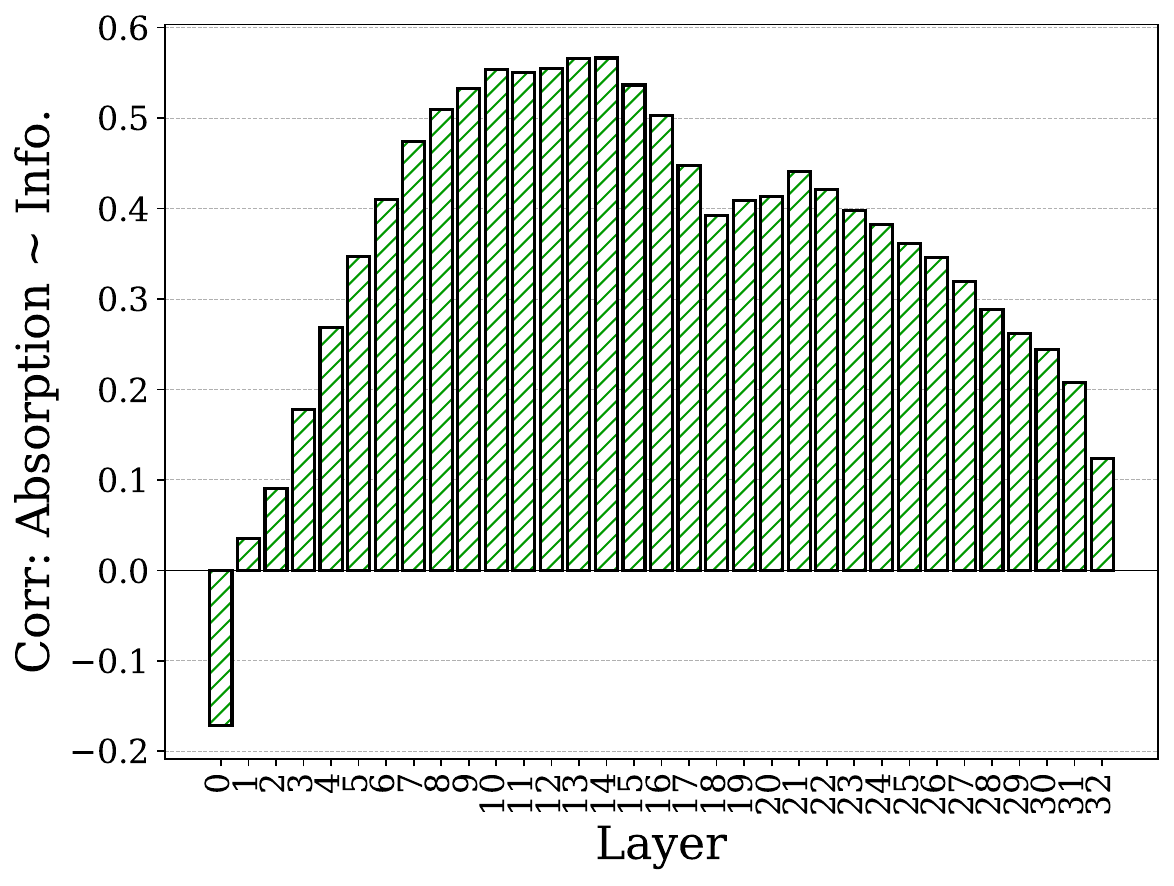}
    \vspace{-2\baselineskip}\caption{Full layer correlation with token length 2 (absorption is a negative measurement) of Fig.~\ref{fig:donor_and_receptor_correlation} on Llama 3-8B.}
    \end{minipage}
\vspace{-0.6\baselineskip}\end{figure}

\begin{figure}[t]
    \centering
    \includegraphics[width=\linewidth]{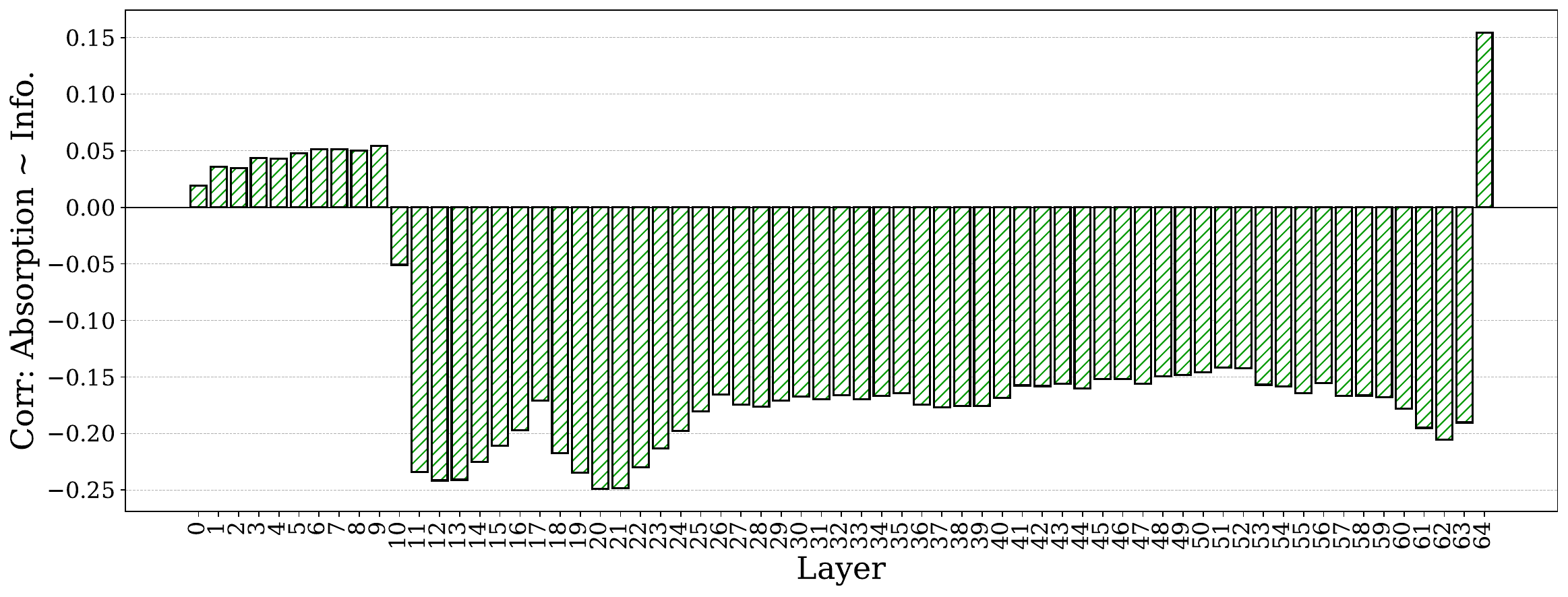}
    \vspace{-2\baselineskip}\caption{Full layer correlation with token length 2 (absorption is a negative measurement) of Fig.~\ref{fig:donor_and_receptor_correlation} on Granite 4.1-30B.}
\vspace{-0.6\baselineskip}\end{figure}

\begin{figure}[t]
    \centering
    \includegraphics[width=\linewidth]{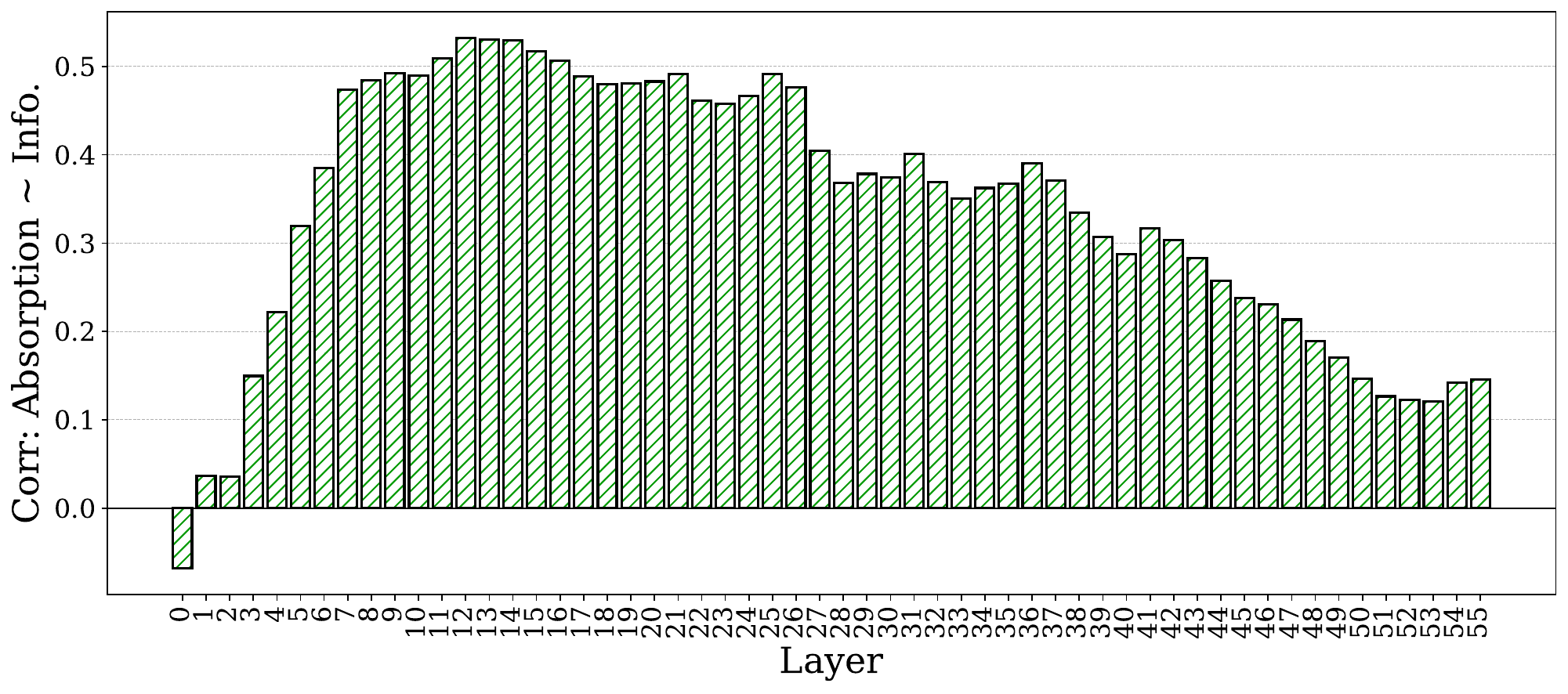}
    \vspace{-2\baselineskip}\caption{Full layer correlation with token length 2 (absorption is a negative measurement) of Fig.~\ref{fig:donor_and_receptor_correlation} on Llama 2-13B.}
\vspace{-0.6\baselineskip}\end{figure}

\begin{figure}[t]
    \centering
    \includegraphics[width=\linewidth]{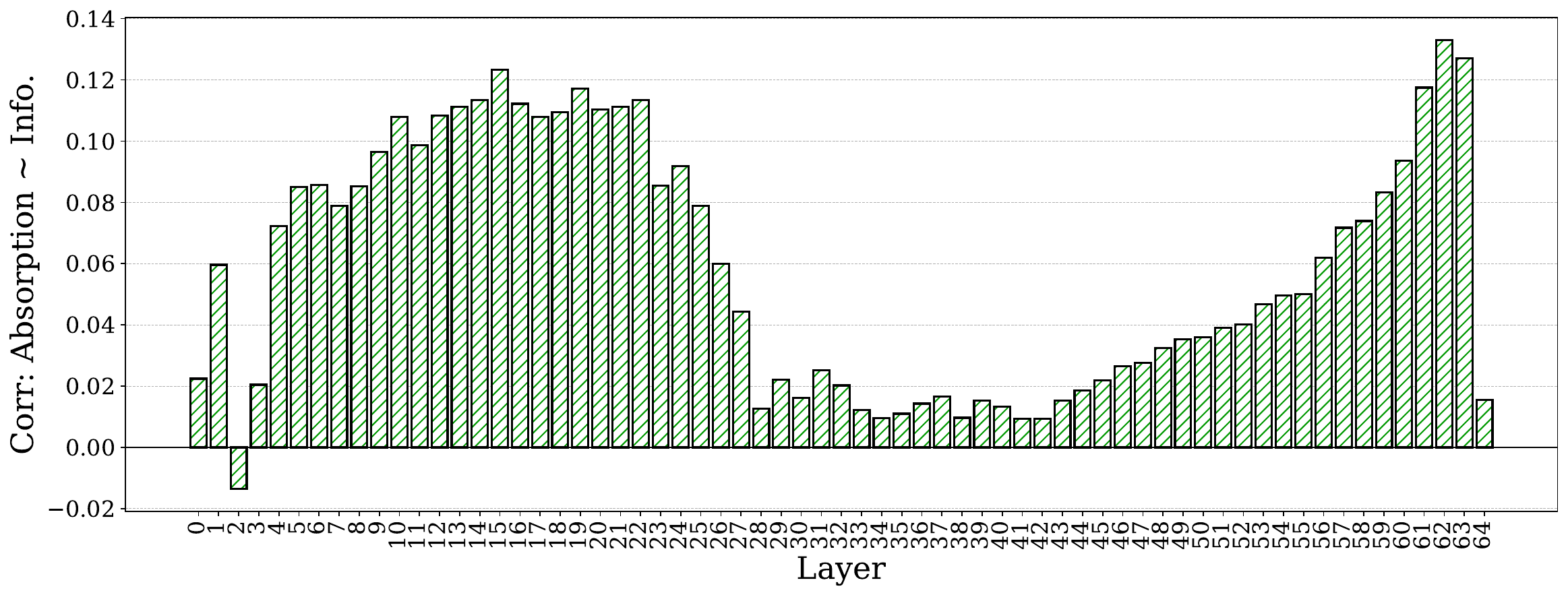}
    \vspace{-2\baselineskip}\caption{Full layer correlation with token length 2 (absorption is a negative measurement) of Fig.~\ref{fig:donor_and_receptor_correlation} on Olmo 3-32B.}
\vspace{-0.6\baselineskip}\end{figure}

\begin{figure}[t]
    \centering
    \includegraphics[width=\linewidth]{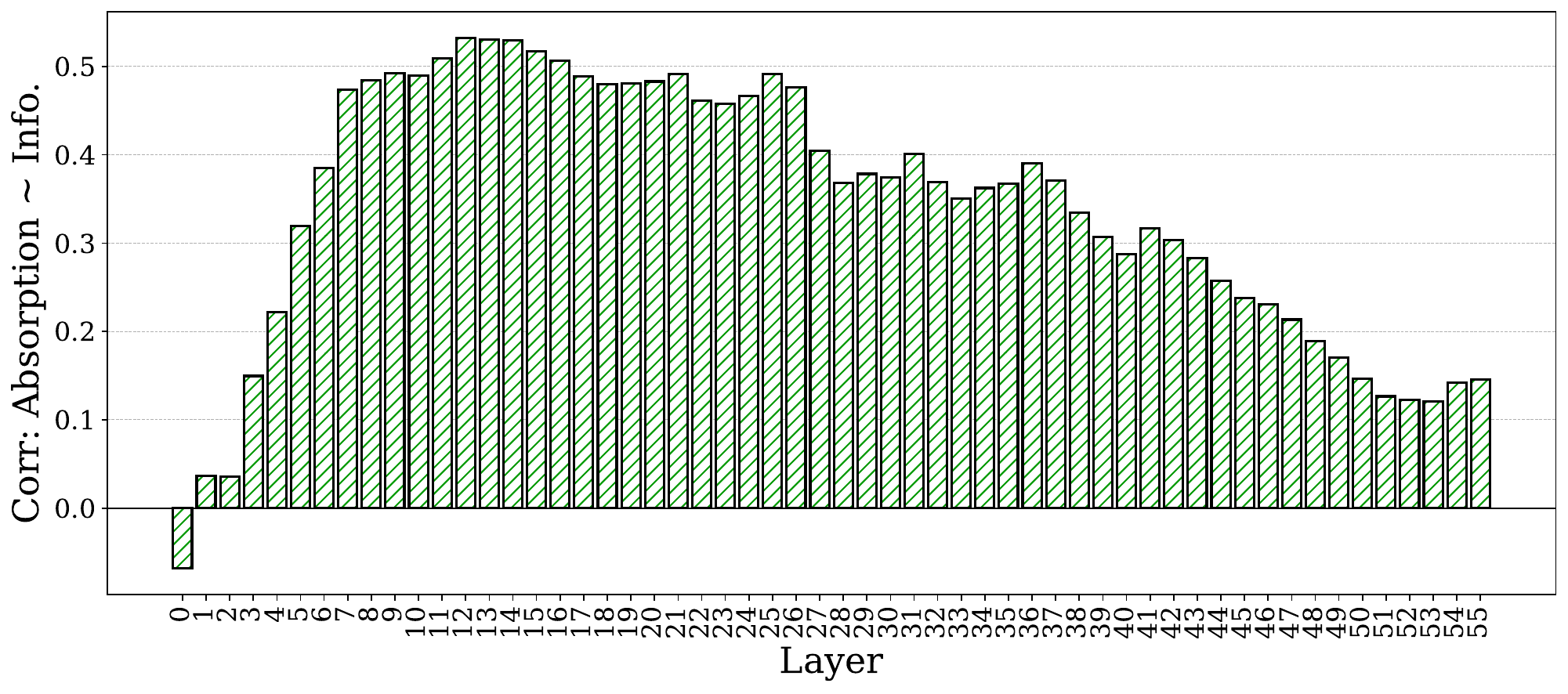}
    \vspace{-2\baselineskip}\caption{Full layer correlation with token length 2 (absorption is a negative measurement) of Fig.~\ref{fig:donor_and_receptor_correlation} on Qwen 3-14B.}
\vspace{-0.6\baselineskip}\end{figure}

\begin{figure}[t]
    \centering
    \begin{minipage}[t]{0.7\linewidth}
    \includegraphics[width=\linewidth]{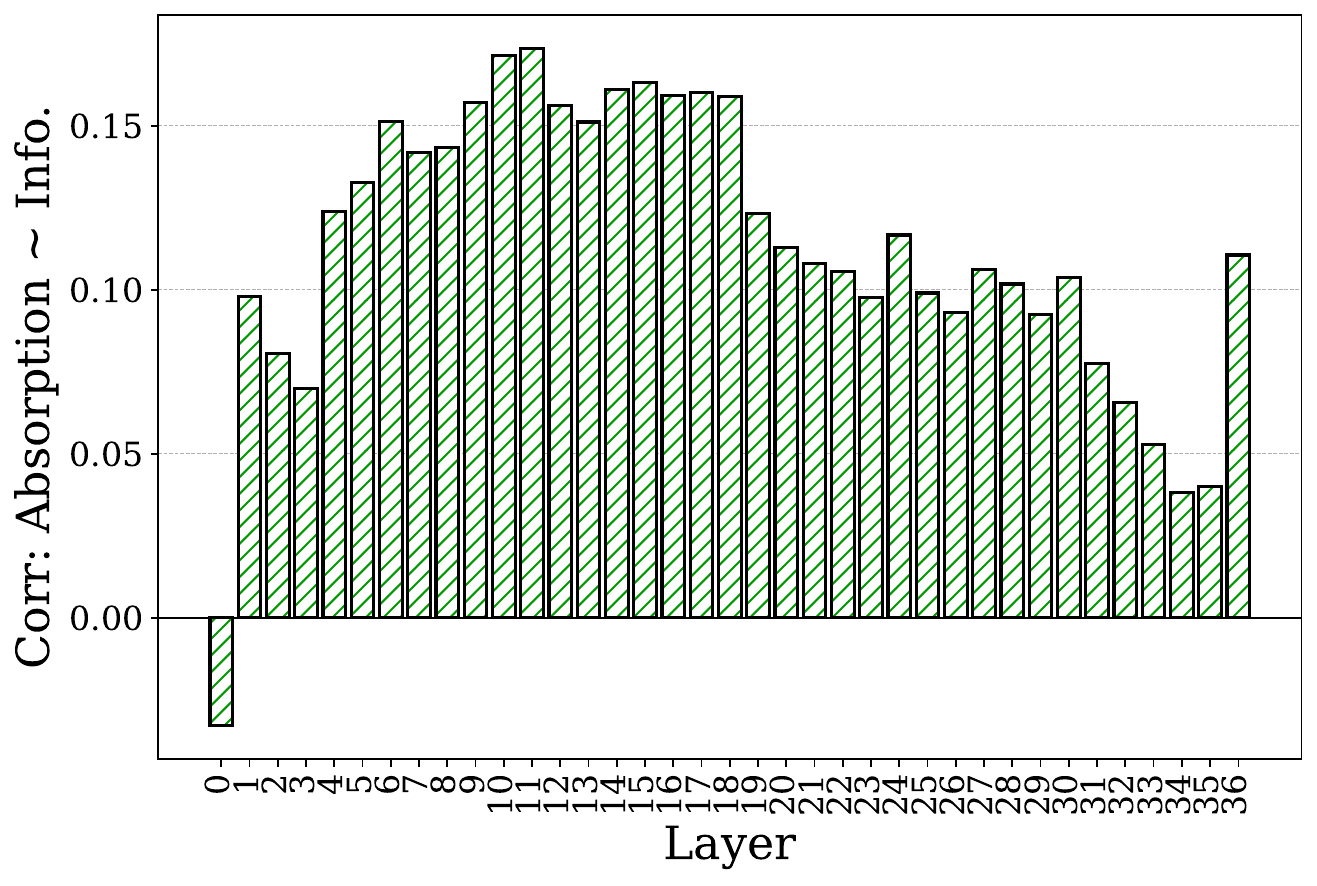}
    \vspace{-2\baselineskip}\caption{Full layer correlation with token length 2 (absorption is a negative measurement) of Fig.~\ref{fig:donor_and_receptor_correlation} on Qwen 3-8B.}
    \end{minipage}
\vspace{-0.6\baselineskip}\end{figure}

\begin{figure}[t]
    \centering
    \includegraphics[width=\linewidth]{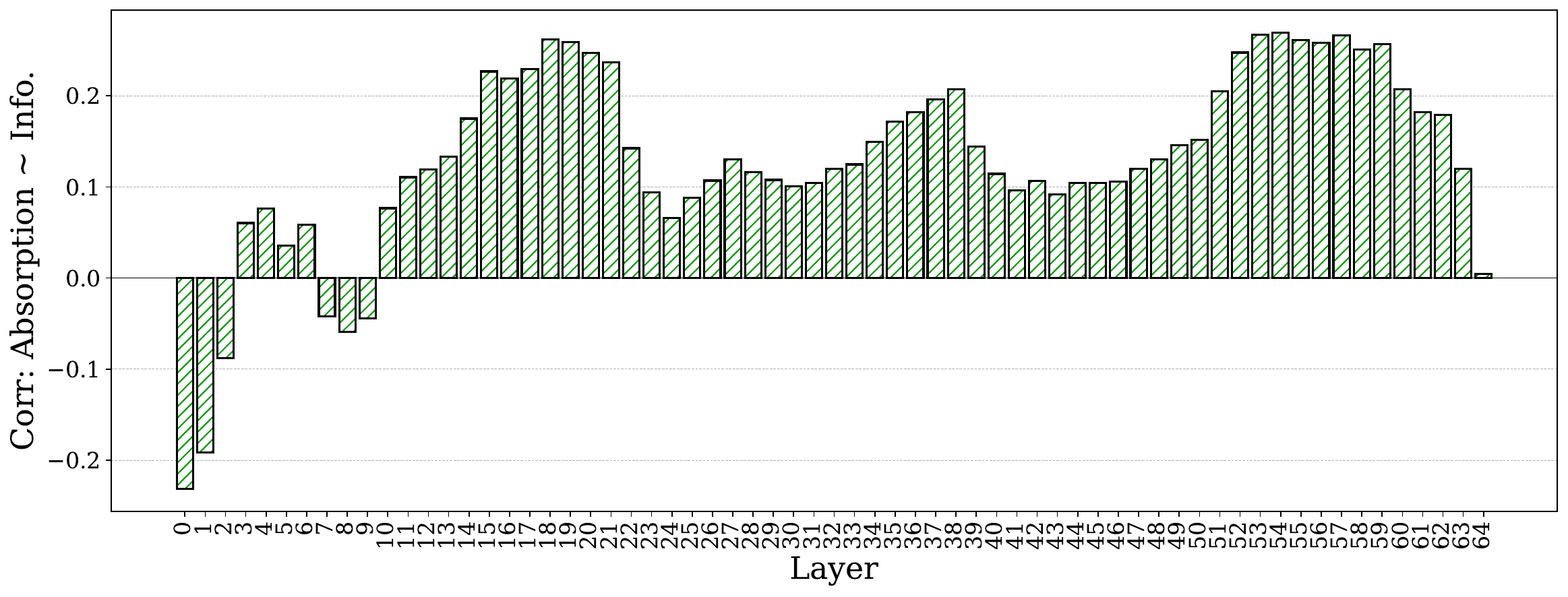}
    \vspace{-2\baselineskip}\caption{Full layer correlation with token length 2 (absorption is a negative measurement) of Fig.~\ref{fig:donor_and_receptor_correlation} on Qwen 3.6-27B.}
    \label{fig:more_exp3_len2_end}
\vspace{-0.6\baselineskip}\end{figure}

\clearpage

\setlength{\LTpre}{0pt}
\setlength{\LTpost}{1.5em}
\setlength{\tabcolsep}{4pt}
\renewcommand{\arraystretch}{1.12}
{\small\begin{longtable}
{@{}
>{\raggedright\arraybackslash}p{0.20\textwidth}
>{\centering\arraybackslash}p{0.07\textwidth}
>{\centering\arraybackslash}p{0.10\textwidth}
>{\raggedright\arraybackslash}p{0.57\textwidth}
@{}}
\caption{Sampled words as the experimental datasets. Frequencies are on the scale of $10^{-6}$.}\label{tab:dataset-vocabulary} \\
\toprule
\textbf{Category} & \textbf{Count} & \textbf{Avg. Freq.} & \textbf{Words} \\
\midrule
\endfirsthead
\caption[]{Sampled words as the experimental datasets (continued).} \\ 
\toprule
\textbf{Category} & \textbf{Count} & \textbf{Avg. Freq.} & \textbf{Words} \\
\midrule
\endhead
\midrule
\multicolumn{4}{r}{Continued on next page.} \\ 
\endfoot
\bottomrule
\endlastfoot
Famous People Name & 47 & 5.36 & Einstein, Shakespeare, Oprah, Picasso, Mandela, Cleopatra, Churchill, Tesla, Elvis, Marie Curie, Gandhi, Aristotle, Monroe, Beethoven, Newton, Amelia Earhart, Frida Kahlo, Rothschild, Darrow, Hemingway, Frost, Twain, Jobs, Job, Obama, J.K. Rowling, Rihanna, Zuckerberg, Mother Teresa, Winston Churchill, Bill Gates, Marlon Brando, Charlie Chaplin, John Lennon, Bruce Lee, Michael Jackson, Madonna, Coco Chanel, Nikola Tesla, Audre Lorde, George Washington, Agatha Christie, Nelson Mandela, Thomas Edison, Steve Jobs, Vincent van Gogh, Serena Williams \\ \midrule
normal people name & 52 & 11.80 & Alice, Bob, Charlie, Emma, David, Sarah, Michael, Jessica, Daniel, Laura, James, Lily, Andrew, Zoe, Matthew, Olivia, Ethan, Mia, Joshua, Ava, Ryan, Grace, Benjamin, Ella, Samuel, Chloe, Lucas, Scarlett, Jacob, Victoria, Henry, Natalie, Jonathan, Aria, Chris, Ruby, Adam, Hailey, William, Leah, Kevin, Julia, Jason, Alexa, Thomas, Claire, Jordan, Sophie, Nathan, Hazel, Isaac, Stella \\ \midrule
famous location name & 45 & 0.22 & Eiffel Tower, Great Wall of China, Taj Mahal, Statue of Liberty, Stonehenge, Machu Picchu, Colosseum, Big Ben, Pyramids of Giza, Golden Gate Bridge, Christ the Redeemer, Sydney Opera House, Acropolis, Niagara Falls, Leaning Tower of Pisa, Burj Khalifa, Mount Fuji, Angkor Wat, Petra, Mount Rushmore, Alhambra, Buckingham Palace, Louvre Museum, Kremlin, Sagrada Família, Table Mountain, Galápagos Islands, Vatican City, Serengeti National Park, Great Barrier Reef, Chichen Itza, Uluru, Himalayas, Amalfi Coast, Venice Canals, Santorini, Yellowstone National Park, Arches National Park, Ha Long Bay, Bora Bora, Mount Kilimanjaro, Bali, Old Faithful, Iguazu Falls, Victoria Falls \\ \midrule
normal location name & 51 & 21.14 & Park, River, Hill, Street, Corner, Avenue, Square, Bridge, Lake, Forest, Trail, Mountain, Road, Path, Beach, Garden, Plaza, Meadow, Bay, Valley, Pond, Cliff, Harbor, Dock, Canyon, Oasis, Sanctuary, Campground, Crossroads, Highway, Dunes, Knoll, Summit, Glade, Orchard, Wetland, Lighthouse, Quarry, Fort, Estuary, Peninsula, Cove, Bungalow, Resort, Retreat, Ranch, Estate, Villa, Commons, Grove, Wilderness \\ \midrule
frequent object noun & 50 & 51.36 & apple, table, chair, computer, book, phone, pen, car, tree, house, bottle, cup, bag, clock, shoe, key, window, door, mirror, lamp, bed, wallet, magazine, guitar, camera, plate, fork, knife, spoon, basket, painting, flower, umbrella, toy, speaker, suitcase, pillow, shirt, jeans, watch, blanket, candle, recipe, map, remote, brush, comb, photo, couch, dish \\ \midrule
hardly used object noun & 50 & 0.32 & abacus, bauble, carafe, dirndl, epergne, flaxen, gazetteer, hovel, ichthyology, jägermeiste, knickknack, locket, marzipan, nightgown, obelisk, parsol, quagmire, ragamuffin, sextant, tiffin, umbilicus, vial, whittle, xylophone, yarmulke, ziggurat, ankh, balustrade, colander, dalliance, effigy, filigree, goblet, harpsichord, incunabulum, jigsaw, kumquat, lichen, madrigal, nanotechnology, omphalos, pecuniary, quasar, recumbent, sconce, tambourine, ukulele, vorpal, wattle, xenophobia \\ \midrule
frequent verb & 53 & 415.03 & be, have, do, say, get, make, go, know, take, see, come, think, look, want, give, use, find, tell, ask, work, seem, feel, try, leave, call, keep, let, begin, help, talk, turn, start, show, hear, play, run, move, like, live, believe, hold, bring, happen, write, provide, sit, stand, lose, pay, meet, include, continue, set \\ \midrule
hardly used verb & 48 & 0.52 & abjure, acquiesce, agglomerate, apoplexy, balderdash, bereave, cacophony, capitulate, circumscribe, declaim, descry, digress, elucidate, emanate, exacerbate, extemporize, fulminate, galvanize, harken, ignoble, imbroglio, juxtapose, kvetch, limn, malinger, negate, obfuscate, pare, quixotic, recalcitrant, redress, reify, remonstrate, ruminate, salubrious, sanctimonious, scour, sedate, strident, subjugate, surreptitious, tantamount, temper, vacillate, venerate, vitiate, winnow, yoke \\ \midrule
frequent adjective & 53 & 86.74 & happy, sad, beautiful, ugly, fast, slow, bright, dark, big, small, loud, quiet, rich, poor, easy, difficult, hot, cold, young, old, clean, dirty, new, old-fashioned, sharp, dull, smooth, rough, heavy, light, sweet, sour, strong, weak, interesting, boring, busy, lazy, friendly, unfriendly, safe, dangerous, soft, hard, clear, cloudy, expensive, cheap, elegant, simple, modern, traditional, powerful \\ \midrule
hardly used adjective & 47 & 0.23 & serendipitous, lurid, ineffable, eldritch, scintillant, nebulous, quixotic, arcane, ephemeral, egregious, cacophonous, insidious, mellifluous, obfuscating, petrichor, verdant, incongruous, apoplectic, sesquipedalian, aphotic, vicissitudinous, phantasmagoric, tenebrous, titillating, diaphanous, uxorious, quizzical, nondescript, sycophantic, salubrious, recondite, turgid, alliterative, jejune, bucolic, apocalyptic, numinous, beguiling, insouciant, unctuous, cacophonic, insipid, resplendent, abysmal, chimerical, inchoate, saturnine \\ \midrule
frequent adverb & 50 & 124.32 & very, quickly, silently, often, completely, beautifully, recently, always, truly, suddenly, carefully, easily, however, usually, almost, perfectly, particularly, immediately, frequently, exactly, simply, rarely, barely, naturally, apparently, gradually, separately, loudly, gently, eventually, deeply, directly, mostly, happily, wildly, temporarily, intensely, thoughtfully, so, already, yet, now, instead, similarly, especially, thus, moreover, consequently, ultimately, nevertheless \\
hardly used adverb & 48 & 0.83 & abjectly, ostensibly, euphemistically, quizzically, inexorably, veritably, haphazardly, assiduously, nefariously, ardently, sporadically, inexplicably, simultaneously, quintessentially, unceremoniously, cryptically, tacitly, doggedly, capriciously, somberly, fortuitously, amenably, perfunctorily, eruditely, inexpensively, abominably, surreptitiously, ostentatiously, unflinchingly, clandestinely, unequivocally, providentially, maverickly, unfathomably, nonchalantly, antithetically, perspicuously, enigmatically, iconoclastically, unyieldingly, preposterously, circuitously, succinctly, agonizingly, indelibly, dimly, axiologically, extemporaneously \\ \midrule
preposition & 50 & 1755.75 & in, on, at, by, for, with, about, against, between, under, over, through, during, before, after, above, below, beside, beyond, among, across, without, within, along, up, down, to, from, toward, near, since, until, via, despite, like, as, in front of, behind, next to, ahead of, inside, outside, according to, in spite of, instead of, because of, due to, for the sake of, out of, in addition to \\ \midrule
conjunction & 49 & 1127.52 & and, but, or, nor, for, so, yet, although, because, since, unless, while, whereas, either, neither, whether, as, if, when, though, even though, in case, after, before, until, despite, in spite of, provided that, as long as, as soon as, according to, like, just as, the same as, not only, but also, both, the other, instead, rather, on the one hand, on the other hand, whether or not, either/or, neither/nor, rather than, in order to, so that, in case that \\ \midrule
interjection & 50 & 3.07 & Wow, Oh, Yay, Oops, Hooray, Eek, Aha, Yikes, Bravo, Whew, Yippee, Ugh, Phew, Whoa, Huh, Eh, Hey, Ouch, Alas, Gosh, Yow, Huzzah, Darn, Dang, Aww, Whoops, Oh dear, Eww, Yummy, Hallelujah, Hooroo, Blimey, Wowza, Yip, Ho, Boo, Right on, Whoopee, Zing, Oof, Ta-da, Jeepers, Crikey, Ewwww, Yessiree, Yay-hoo, Ooh, Awwww, Tsk, Mercy \\ \midrule
pronoun & 53 & 1150.90 & I, you, he, she, it, we, they, me, him, her, us, them, my, your, his, its, our, their, mine, yours, hers, ours, theirs, whom, whoever, whomever, this, that, these, those, such, one, each, many, few, all, some, none, both, either, neither, what, which, who, whose, someone, anyone, everyone, no one, something, anything, everything, nothing \\ 
\end{longtable}
}
{\small\begin{longtable}{@{}>{\raggedleft\arraybackslash}p{0.05\textwidth}>{\raggedright\arraybackslash}p{0.43\textwidth}>{\raggedright\arraybackslash}p{0.46\textwidth}@{}}
\caption{Sampled queries and their output candidates.}\label{tab:dataset-questions} \\ 
\toprule
\textbf{No.} & \textbf{Question} & \textbf{Output Candidates} \\
\midrule
\endfirsthead
\caption[]{Sampled queries and their answer candidates (continued).} \\ 
\toprule
\textbf{No.} & Question & \textbf{Output Candidates} \\
\midrule
\endhead
\midrule
\multicolumn{3}{r}{Continued on next page.} \\ 
\endfoot
\bottomrule
\endlastfoot
1 & What grammatical gender (if any) is typically associated with this people-word in languages that mark gender? & 1) masculine; 2) feminine; 3) common/epicene; 4) no grammatical gender \\
2 & What age group does this word primarily denote? & 1) child; 2) adolescent; 3) adult; 4) elderly; 5) age-neutral \\
3 & Does this word denote a familial kinship relation? If so, which basic role? & 1) parent; 2) child; 3) sibling; 4) grandparent; 5) spouse/partner; 6) other kin; 7) not a kinship term \\
4 & What occupational sector or field does this word most directly refer to? & 1) healthcare; 2) education; 3) government/public service; 4) business/finance; 5) arts/entertainment; 6) trades/industry; 7) science/technology; 8) agriculture; 9) service/hospitality; 10) not an occupation \\
5 & Does this word primarily indicate a social or legal status? If so, which? & 1) citizen; 2) immigrant/resident; 3) refugee/asylum-seeker; 4) stateless; 5) legal/official title (e.g., holder of an office); 6) not a status term \\
6 & Is this word typically used as a formal title or honorific? & 1) honorific/title (e.g., Sir, Dr.); 2) professional title (e.g., Judge, Professor); 3) noble/royal title (e.g., King, Duchess); 4) informal/colloquial address; 5) not a title \\
7 & What evaluative stance or connotation does this word most commonly carry? & 1) neutral/descriptive; 2) positive/respectful; 3) negative/pejorative; 4) affectionate/familiar \\
8 & Does this word indicate membership in a demographic category? If so, which type? & 1) nationality; 2) ethnicity; 3) religion; 4) gender identity; 5) sexual orientation; 6) socioeconomic class; 7) age group; 8) not a demographic term \\
9 & Does this word more often refer to an individual person or to a collective/group? & 1) individual; 2) collective/group; 3) both (can denote either) \\
10 & Can this word be applied metaphorically to non-humans (e.g., animals, organizations) or is it restricted to humans? & 1) humans only; 2) humans and anthropomorphized non-humans; 3) metaphorical use common \\
11 & Is the role or position denoted by this word typically obtained by election/appointment, inherited, or earned through training? & 1) elected/appointed; 2) inherited/hereditary; 3) earned via training/professional qualification; 4) acquired by social/birth identity; 5) varies \\
12 & Is this word primarily a proper name (referring to a specific person) or a common noun (a class label)? & 1) proper name (specific individual); 2) common noun (category/class); 3) both (used as name and class) \\
13 & What level of social power or authority does the role implied by this word typically carry? & 1) high authority/leadership; 2) moderate authority/supervisory; 3) peer-level/independent; 4) low authority/subordinate; 5) varies/not applicable \\
14 & Does this word describe a typically temporary role/state or a more permanent identity? & 1) temporary role/state (e.g., visitor, patient); 2) permanent identity/status (e.g., native, widower); 3) can be either \\
15 & Does this word usually imply a required formal qualification or certification? & 1) requires formal certification/licensing; 2) requires informal training/experience; 3) no formal qualification required; 4) varies \\
16 & What register or level of formality best fits this word in normal usage? & 1) formal; 2) neutral/common; 3) colloquial/informal; 4) slang/pejorative \\
17 & What type of location is this? & 1) country; 2) state/province/region; 3) city; 4) town/village; 5) neighborhood; 6) street/road; 7) landmark/monument; 8) building/structure; 9) park/green space; 10) body of water; 11) mountain/peak; 12) island; 13) airport; 14) railway station; 15) shopping center/mall; 16) other \\
18 & What is the administrative level of this word? & 1) sovereign country; 2) first-level administrative division (state/province/region); 3) second-level administrative division (county/district); 4) municipality/city; 5) submunicipal division (borough/ward/neighborhood); 6) non-administrative geographic feature \\
19 & Is this word primarily natural or man-made? & 1) natural feature; 2) man-made/constructed feature; 3) mixed (natural with significant human alteration) \\
20 & Is this word primarily indoor or outdoor? & 1) indoor; 2) outdoor; 3) both indoor and outdoor (integrated) \\
21 & What is the typical public accessibility of this word? & 1) publicly accessible (open to all); 2) restricted access (requires permission/ticket); 3) private (not open to public); 4) controlled/secured (e.g., military, high security) \\
22 & What is the primary function or land use of this word? & 1) residential; 2) commercial/retail; 3) industrial; 4) transportation hub; 5) administrative/government; 6) recreational/park; 7) religious; 8) historical/heritage site; 9) educational; 10) mixed-use; 11) natural habitat/ecosystem \\
23 & What is the typical spatial scale of this word? & 1) global/transnational (spans multiple countries); 2) national; 3) regional (state/province level); 4) city-level; 5) neighborhood-level; 6) site-level (single building/park); 7) micro (room/platform/terminal) \\
24 & Is this word an inhabited settlement? & 1) yes — permanently inhabited; 2) seasonally inhabited (e.g., seasonal village, resort); 3) no — not a settlement (e.g., mountain, industrial site) \\
25 & How would you classify the urbanization level of this word? & 1) urban; 2) suburban; 3) rural; 4) peri-urban; 5) remote/isolated \\
26 & Does this word contain other named locations or subdivisions? & 1) contains multiple subordinate administrative units; 2) contains subordinate sites/venues (e.g., buildings in a campus); 3) contains a single subordinate site; 4) does not contain other named locations \\
27 & Is this word associated with transportation infrastructure? & 1) major transportation hub (international airport/major seaport/central rail hub); 2) local transit facility (bus stop/metro station/local rail); 3) on a major travel route (highway/rail corridor); 4) not associated with transportation infrastructure \\
28 & Which climate or environmental zone best describes this word? & 1) tropical; 2) subtropical; 3) temperate; 4) continental; 5) arid/desert; 6) polar/alpine; 7) Mediterranean; 8) marine/coastal \\
29 & What is the relationship of this word to significant bodies of water? & 1) coastal/shoreline; 2) on a river/stream; 3) adjacent to a lake/pond; 4) island/surrounded by water; 5) wetland/marsh; 6) inland/no significant nearby waterbody \\
30 & What is the typical temporal permanence of this word? & 1) permanent/long-term; 2) seasonal/annual; 3) temporary/pop-up \\
31 & What legal designation or protected status applies to this word? & 1) sovereign territory; 2) nationally protected area (e.g., national park); 3) internationally designated protected site (e.g., UNESCO World Heritage); 4) zoned for specific use (residential/commercial/industrial); 5) private property; 6) military/secured zone \\
32 & On which continent is this word located? & 1) Africa; 2) Antarctica; 3) Asia; 4) Europe; 5) North America; 6) South America; 7) Oceania \\
33 & What is the typical material composition of this word? & 1) metal; 2) plastic; 3) wood; 4) glass; 5) paper; 6) fabric/textile; 7) ceramic/porcelain; 8) stone; 9) composite (mixed materials) \\
34 & What is the primary function or use category of this word? & 1) containment/storage; 2) transportation/mobility; 3) seating/support; 4) illumination/light source; 5) cooking/food preparation; 6) cleaning/hygiene; 7) communication/information; 8) measurement/instrumentation; 9) construction/structural; 10) entertainment/recreation; 11) decoration/ornamentation; 12) protection/safety; 13) medical/healthcare; 14) personal grooming; 15) clothing/ wearable; 16) other functional equipment \\
35 & What is the typical size class of this word relative to a human? & 1) very small (fits in palm); 2) small (handheld); 3) medium (requires both hands); 4) large (requires more than one person to move); 5) room-sized or larger \\
36 & What is the usual shape category of this word? & 1) flat/planar; 2) elongated/rod-like; 3) spherical/round; 4) cylindrical/tubular; 5) box-like/cuboid; 6) sheet/film; 7) irregular/organic \\
37 & Is this word typically a count noun or a mass noun? & 1) countable (discrete units); 2) mass/uncountable (measured by amount or mass) \\
38 & What is the object's typical mobility or portability? & 1) handheld/portable (single person); 2) portable but heavy (requires equipment); 3) fixed/installed (permanent); 4) self-mobile/vehicle-like (self-propelled) \\
39 & Does this word usually require a power source to operate? & 1) requires electrical/battery power; 2) powered mechanically (manual/crank); 3) no power required (passive) \\
40 & Is this word primarily natural, man-made, or a processed natural material? & 1) natural (occurs in nature); 2) man-made/artifact; 3) processed natural material (e.g., lumber, metal ingot) \\
41 & What is the typical environment or location where this word is found? & 1) indoors (home/office); 2) outdoors (parks/streets); 3) both indoors and outdoors; 4) industrial/warehouse; 5) vehicle/interior \\
42 & What is the object's usual lifespan or durability category? & 1) disposable/single-use (hours–days); 2) short-lived (weeks–months); 3) medium durability (months–years); 4) long-lasting/durable (years–decades); 5) heritage/permanent (decades–centuries) \\
43 & Who or what typically uses/operates this word? & 1) used directly by humans; 2) used by animals; 3) operated by machines/robots; 4) used by both humans and machines \\
44 & Is the object primarily functional, primarily decorative, or both? & 1) primarily functional; 2) primarily decorative; 3) both functional and decorative \\
45 & Is this word typically standalone, part of a set, or part of a larger system/assembly? & 1) typically standalone; 2) typically sold/used in sets (pairs/sets); 3) typically part of a larger system or assembly \\
46 & What is the typical user demographic for this word? & 1) children; 2) adolescents/teens; 3) adults; 4) elderly; 5) all ages; 6) specialized professionals \\
47 & What commercial/category classification best fits this word? & 1) consumer/grocery item; 2) household goods/furniture; 3) personal accessory/clothing; 4) electronics/appliances; 5) tools/equipment; 6) industrial/machinery; 7) raw material; 8) toy/recreational \\
48 & What safety risk level is typically associated with normal use of this word? & 1) low risk (no protective measures needed); 2) moderate risk (care recommended); 3) high risk (requires training or PPE); 4) hazardous (toxic, explosive, or highly dangerous) \\
49 & What is the transitivity of this word (does it take objects and how many)? & 1) intransitive (no direct object); 2) monotransitive (one direct object); 3) ditransitive (two objects: e.g., recipient and theme); 4) ambitransitive (can be used with or without an object) \\
50 & What is the typical valency (number of core arguments) of this word? & 1) monovalent (1 core argument); 2) divalent (2 core arguments); 3) trivalent (3 core arguments); 4) variable valency across senses \\
51 & Which Aktionsart / aspectual class best describes this word? & 1) state; 2) activity; 3) accomplishment; 4) achievement; 5) semelfactive (punctual iterative) \\
52 & Is this word telic (has an inherent endpoint) or atelic? & 1) telic (inherent endpoint); 2) atelic (no inherent endpoint); 3) both depending on context or object \\
53 & Is this word stative or dynamic (eventive)? & 1) stative; 2) dynamic (eventive); 3) both depending on sense \\
54 & Is this word punctual/instantaneous or durative/extended? & 1) punctual / instantaneous; 2) durative / extended; 3) both depending on sense \\
55 & Does this word have inherent causative meaning or readily form causatives? & 1) inherently causative or regularly causativized; 2) non-causative and resists causativization; 3) some senses causative, others non-causative \\
56 & Does this word permit causative/anticausative (ergative) alternation (e.g., 'Someone broke the glass' vs. 'The glass broke')? & 1) permits causative/anticausative alternation; 2) does not permit causative/anticausative alternation \\
57 & Can this word form a passive (patient promotion / agent suppression)? & 1) passivizable (regular passive formation possible); 2) non-passivizable (resists passive formation) \\
58 & Can this word be used reflexively (subject acts on itself)? & 1) allows reflexive use; 2) disallows reflexive use \\
59 & Can this word express reciprocal action (mutual action between participants)? & 1) allows reciprocal use; 2) disallows reciprocal use \\
60 & Which types of clausal complements can this word take? & 1) finite that-clause (e.g., that …); 2) to-infinitive (control); 3) to-infinitive (raising); 4) bare infinitive (catenative/bare VP complement); 5) gerund-participial (-ing) clause; 6) none of the above (no clausal complement) \\
61 & If this word takes an infinitival complement, does it pattern as a control verb or a raising verb? & 1) subject-control verb (matrix subject controls infinitival subject); 2) object-control verb (object controls infinitival subject); 3) raising verb (matrix subject raises to subject position); 4) does not take infinitival complements \\
62 & What semantic role does a canonical subject of this word typically realize? & 1) agentive (causer / intentional agent); 2) experiencer (feeler / perceiver); 3) theme/patient (entity undergoing change or being affected); 4) location/temporal or other non-agentive role \\
63 & Does this word strongly require animate (especially human) arguments in subject or object positions? & 1) requires animate (human) argument; 2) prefers animate but allows inanimate; 3) prefers inanimate; 4) no strong animacy preference \\
64 & Is this word compatible with the progressive aspect (e.g., 'is \_\_\_ing')? & 1) compatible with progressive aspect; 2) incompatible with progressive aspect; 3) compatibility depends on particular sense or use \\
65 & What is the evaluative polarity of this word (when used in ordinary predicative or attributive contexts)? & 1) positive (favorable); 2) negative (unfavorable); 3) neutral/descriptive \\
66 & Is this word gradable? & 1) gradable (allows degrees: e.g., very/more); 2) non-gradable/absolute (does not allow degree readings) \\
67 & Which degree morphology does this word use? & 1) regular gradable (uses -er/-est or more/most); 2) irregular comparative/superlative (e.g., good/better/best); 3) does not form comparative or superlative \\
68 & Which intersectivity class best describes this word when it modifies nouns? & 1) intersective (e.g., 'red' as in 'red car'); 2) subsective (e.g., 'skillful' as in 'skillful surgeon'); 3) privative (e.g., 'fake' as in 'fake painting'); 4) intensional/non-intersective (e.g., 'alleged') \\
69 & What type of measurement scale does this word invoke? & 1) closed/absolute scale with endpoints (e.g., 'full', 'empty'); 2) open/relative scale without fixed endpoints (e.g., 'tall'); 3) binary/dichotomous (no degrees, e.g., 'pregnant') \\
70 & Is the adjective primarily relational (denoting a relation to some other noun) or attributive (describing an intrinsic property)? & 1) primarily relational (e.g., 'medical', 'parental'); 2) primarily attributive/descriptive (e.g., 'blue', 'heavy'); 3) both relational and attributive \\
71 & Which semantic dimension does this word primarily express? & 1) size; 2) age; 3) colour; 4) shape; 5) material; 6) temperature; 7) speed; 8) weight; 9) evaluative/value; 10) origin/nationality; 11) texture; 12) quantity/amount; 13) temporal \\
72 & What is the degree of subjectivity of this word? & 1) subjective (opinion-based); 2) moderately subjective (context-sensitive); 3) objective (observable/measurable) \\
73 & What syntactic distribution does this word allow? & 1) attributive only (appears before a noun); 2) predicative only (appears after a copula); 3) both attributive and predicative; 4) post-nominal only (follows the noun) \\
74 & How does this word accept degree modifiers? & 1) compatible with scalar modifiers like 'very'/'quite' (gradable); 2) compatible mainly with absolute modifiers like 'completely'/'absolutely' (absolute adjectives); 3) incompatible with standard degree modifiers \\
75 & What kind of antonymy relation does this word participate in? & 1) complementary antonymy (binary opposites, e.g., 'alive'/'dead'); 2) gradable antonymy (opposites on a scale, e.g., 'hot'/'cold'); 3) relational antonymy (role-based opposites, e.g., 'parental'/'filial'); 4) converse/reciprocal antonymy \\
76 & Does this word denote a stable/classificatory property or a temporary/episodic state? & 1) stable/classificatory property (inherent or long-term); 2) temporary/stateful (episodic condition); 3) context-dependent/can be either \\
77 & Is this word epistemic/modal (expressing uncertainty, evidentiality or modality) or descriptive? & 1) epistemic/modal (e.g., 'probable', 'alleged'); 2) descriptive/property-denoting; 3) both epistemic/modal and descriptive \\
78 & Can the adjective be interpreted as resultative (indicating the result of an event) rather than merely stative? & 1) resultative (denotes result state after an event); 2) stative/non-resultative (simple property); 3) both resultative and stative readings available \\
79 & Is this word typically compatible with explicit numeric measurement expressions? & 1) measurable with numeric scales (e.g., '5 cm tall'); 2) comparative but not directly numeric (relative measure only); 3) non-measurable/binary (no numeric measurement) \\
80 & What kind of noun arguments does this word typically modify? & 1) primarily animate/human nouns; 2) primarily inanimate nouns; 3) both animate and inanimate; 4) human-specific (applies mainly to people) \\
81 & Which primary semantic category does this word belong to? & 1) manner; 2) time/temporal; 3) place/locative; 4) frequency; 5) degree/intensity; 6) epistemic/modality; 7) causal; 8) purpose; 9) discourse/connective; 10) attitude/evaluation; 11) directional; 12) aspectual \\
82 & What temporal relation relative to the described event does this word indicate? & 1) before; 2) after; 3) during; 4) simultaneous; 5) habitual/regular; 6) immediate \\
83 & What frequency value does this word express? & 1) always; 2) usually; 3) often; 4) sometimes; 5) rarely; 6) never \\
84 & What degree or intensity does this word encode? & 1) minimal/slight; 2) moderate; 3) strong/high; 4) extreme/absolute; 5) binary/all-or-none \\
85 & What epistemic stance or level of certainty does this word convey? & 1) certain/definite; 2) probable/likely; 3) possible; 4) doubtful/unlikely; 5) counterfactual/contrary-to-fact \\
86 & What discourse relation between clauses does this word signal? & 1) addition; 2) contrast; 3) concession; 4) cause/result; 5) sequence/temporal; 6) explanation/clarification \\
87 & Does this word express an evaluative attitude, and if so what is its valence? & 1) positive/appreciative; 2) negative/critical; 3) neutral/objective \\
88 & What semantic scope does this word take in the sentence? & 1) verb phrase / event-level; 2) entire clause / sentence-level; 3) adjective or adverb phrase; 4) prepositional phrase; 5) quantifier/determiner scope \\
89 & Is this word gradable (compatible with degree modification like 'very')? & 1) gradable (compatible with degree modifiers); 2) non-gradable (incompatible with degree modifiers) \\
90 & Does this word express directional meaning for motion, and if so which direction? & 1) toward/forward; 2) away/backward; 3) upward; 4) downward; 5) around/along; 6) no directional meaning \\
91 & What aspectual contribution does this word make to the event structure? & 1) punctual/instantaneous; 2) durative/continuous; 3) iterative/repetitive; 4) inchoative/beginning; 5) terminative/ending; 6) no aspectual contribution \\
92 & Does this word function as a focus or contrast marker? & 1) focus/contrastive (emphatic); 2) non-contrastive/neutral \\
93 & How does this word affect clause polarity or negation? & 1) explicit negation (e.g., 'not'); 2) downward-entailing/minimizer (e.g., 'hardly'); 3) affirmative/reinforcing (e.g., 'indeed'); 4) neutral/no polarity effect \\
94 & What register or formality level is this word typically associated with? & 1) formal; 2) informal/colloquial; 3) neutral/common; 4) archaic/literary \\
95 & Does this word encode evidentiality (source of information), and if so which type? & 1) direct/firsthand; 2) reported/hearsay; 3) inference/deduction; 4) assumed/general knowledge; 5) no evidential meaning \\
96 & Does this word quantify over events or participants, and if so what quantificational pattern does it express? & 1) universal (all/every); 2) existential/some (at least once); 3) proportional/most; 4) distributive/each; 5) no quantification \\
97 & What is the primary semantic domain of the preposition this word? & 1) spatial; 2) temporal; 3) causal/reason; 4) manner; 5) instrumental; 6) beneficiary/recipient; 7) comparative; 8) purpose; 9) topic/aboutness; 10) other \\
98 & If this word is spatial, which spatial role does it most often express? & 1) location (static position); 2) direction/path; 3) source/origin; 4) goal/endpoint; 5) containment/inside; 6) contact/adjacency; 7) mixed/other \\
99 & If this word is temporal, which temporal relation does it most often express? & 1) point in time (at/on); 2) duration/period (during/for); 3) starting point (since/from); 4) frequency (per); 5) temporal order/sequence (after/before); 6) mixed/other \\
100 & Which participant-role semantics does this word typically encode when introducing an argument? & 1) instrument (with/by); 2) beneficiary/recipient (for/to); 3) agent/causer (by); 4) experiencer (to/for); 5) possessor/ownership (of); 6) co-participant/accompaniment (with); 7) none/role not encoded \\
101 & What complement type does this word primarily select? & 1) noun phrase (NP) only; 2) clausal complement (CP) only; 3) gerund/verbal noun (V-ing) primarily; 4) adjective phrase (AP) or nominalizer; 5) accepts multiple complement types \\
102 & Does this word show a preference for animate versus inanimate complements? & 1) typically requires an animate complement; 2) typically requires an inanimate complement; 3) accepts either animate or inanimate \\
103 & Which spatial frame of reference does this word primarily use for locative interpretations? & 1) relative/egocentric (e.g., left/right of speaker); 2) intrinsic/object-centered (e.g., in front of the car); 3) absolute/geographic (e.g., north of); 4) mixed \\
104 & Is the meaning of this word typically compositional when it occurs in fixed expressions/phrases? & 1) usually compositional (literal meaning preserved); 2) often idiomatic/non-compositional; 3) mixed depending on expression \\
105 & How polysemous is this word across distinct senses? & 1) monosemous (one core sense); 2) few distinct senses (2–3); 3) many distinct senses (4 or more) \\
106 & When combined with verbs of motion, does this word mainly mark path, endpoint, or both? & 1) path/trajectory (through/along); 2) endpoint/goal (to/into/onto); 3) both path and endpoint; 4) neither (not used with motion verbs) \\
107 & Can this word be stranded (left behind) in English-style wh-movement or passivization? & 1) can be stranded (commonly); 2) cannot be stranded (ungrammatical); 3) stranding possible but register-dependent \\
108 & Can this word function productively as a verb-particle (phrasal verb particle) rather than only as a preposition? & 1) yes, commonly used as particle; 2) sometimes/limitedly used as particle; 3) no, not used as particle \\
109 & Does this word commonly encode measure/quantity/degree relations? & 1) measure/extent (e.g., by, to indicate amount); 2) ratio/proportion (e.g., per); 3) degree/intensity (e.g., by in 'by far'); 4) no measure/quantity semantics \\
110 & How prone is this word to attachment/scope ambiguity (e.g., ambiguous whether it modifies a verb phrase vs. noun phrase)? & 1) high propensity for attachment ambiguity; 2) moderate propensity; 3) low propensity (typically unambiguous) \\
111 & Does this word more often function as an obligatory argument (complement) of predicates or as an optional adjunct? & 1) typically an obligatory complement/argument; 2) typically an optional adjunct; 3) both depending on the predicate \\
112 & Does this word commonly express comparison or contrast relations (e.g., comparative marker or instead-of meaning)? & 1) comparison (e.g., than-like use); 2) contrast/substitution (e.g., instead of); 3) both comparison and contrast uses; 4) neither \\
113 & What primary semantic relation does this word express between the connected units? & 1) additive (addition/and); 2) adversative/contrast (but/however); 3) alternative/disjunction (or/either); 4) causal (because/so); 5) temporal (when/while/after); 6) conditional (if/unless); 7) concessive (although/though); 8) purposeful (so that/in order that); 9) resultative (therefore/hence); 10) explanatory (that is/i.e.) \\
114 & Does this word operate as a coordinator (joins equals) or as a subordinating marker (introduces a dependent relation)? & 1) coordinator (coordination); 2) subordinator (subordination); 3) both (can coordinate or subordinate) \\
115 & Which temporal relation does this word primarily encode when linking events? & 1) precedence (event A before event B); 2) succession/result (event A after/causes event B); 3) simultaneity (events occur together); 4) no specific temporal relation \\
116 & Does this word signal exclusive alternatives or allow inclusive interpretation of alternatives? & 1) inclusive (one or more alternatives may be true); 2) exclusive (exactly one alternative true); 3) context-dependent (can be interpreted either way) \\
117 & Is the semantic relation introduced by this word symmetric between the conjuncts or asymmetric (one conjunct relates to or depends on the other)? & 1) symmetric (both conjuncts on equal footing); 2) asymmetric (one conjunct subordinate or dependent); 3) either depending on usage \\
118 & Does this word primarily convey causal/explanatory force (connects premise to explanation) or just coordination without inference? & 1) causal/explanatory (signals reason or explanation); 2) non-causal coordination (no inference implied); 3) both depending on context \\
119 & What kind of discourse relation does this word most often signal at the level of sentence-to-sentence connectivity? & 1) addition/continuation; 2) contrast/opposition; 3) cause/result; 4) sequence/temporal progression; 5) condition/hypothesis; 6) concession/contrastive concession; 7) explanation/elaboration \\
120 & Does this word introduce a presupposition about the truth or existence of one of the conjuncts? & 1) presuppositional (assumes truth/existence of a conjunct); 2) assertive (asserts the content of conjuncts); 3) neutral (neither presupposes nor specifically asserts) \\
121 & Does this word tend to trigger scalar or quantity implicatures (e.g., suggesting exclusivity, emphasis, or ordering) in typical use? & 1) scalar implicature (implies scale or stronger vs weaker); 2) quantity implicature (suggests amount/number implications); 3) contrastive implicature (implies opposition); 4) no typical implicature \\
122 & Does this word license semantic coordination across different syntactic categories (e.g., can link a noun phrase to a clause) or is it restricted to same-category linking? & 1) can link different categories (cross-category coordination); 2) restricted to same syntactic category; 3) restricted but with limited exceptions \\
123 & How does negation typically scope with this word: over the whole coordinated structure, over individual conjuncts, or ambiguous? & 1) negation scopes the whole coordination (negates entire conjunction); 2) negation scopes individual conjunct(s) (each conjunct can be negated separately); 3) ambiguous depending on context \\
124 & Does this word carry a preference for affirmative vs. negative polarity contexts (i.e., is it polarity-sensitive)? & 1) prefers affirmative contexts; 2) prefers negative contexts; 3) neutral to polarity (works in both) \\
125 & Does this word commonly mark contrastive focus on one conjunct (highlighting one part as the salient/contrasted element)? & 1) yes (marks contrastive focus); 2) no (does not mark contrastive focus) \\
126 & Is the semantic contribution of this word typically formal/academic, informal/colloquial, or neutral/register-flexible? & 1) formal (used in formal/written registers); 2) informal (colloquial/spoken); 3) neutral/register-flexible (used across registers) \\
127 & Does this word introduce a direction of reasoning (e.g., premise -> conclusion vs. conclusion -> premise) in inferential contexts? & 1) premise-to-conclusion (forward inference); 2) conclusion-from-premise (backward/summary inference); 3) no specific inferential direction \\
128 & Does this word permit ellipsis (gapping/stripping) of repeated material across conjuncts without loss of interpretation? & 1) permits ellipsis across conjuncts (supports gapping/stripping); 2) does not permit ellipsis (requires full repetition); 3) limited support for ellipsis depending on structure \\
129 & What is the primary illocutionary force (speech-act) of this word? & 1) expressive/emotive (shows feeling); 2) phatic/attention-getting; 3) directive/requesting/commanding; 4) vocative/calling; 5) filler/hesitation; 6) backchannel/acknowledgement; 7) onomatopoetic/sound-imitative; 8) performative (e.g., thanks, apology) \\
130 & What primary emotion or affect does this word convey? & 1) joy/amusement; 2) surprise; 3) anger/annoyance; 4) disgust; 5) fear/alarm; 6) sadness/grief; 7) pain/physical discomfort; 8) approval/pleasure; 9) uncertainty/hesitation; 10) neutral/attention-only \\
131 & What is the valence of this word? & 1) positive; 2) negative; 3) neutral; 4) ambiguous/context-dependent \\
132 & What is the typical intensity or strength of this word? & 1) low/subtle; 2) moderate; 3) high/strong; 4) variable/context-dependent \\
133 & What is the typical politeness or formality level of this word? & 1) formal/polite; 2) neutral/standard; 3) informal/casual; 4) slang/vulgar/rude \\
134 & What is the typical syntactic position of this word within an utterance? & 1) standalone as a full utterance; 2) utterance-initial; 3) utterance-final; 4) inserted parenthetical/medial \\
135 & Is this word primarily addressed to the listener, to the speaker themself, or neither? & 1) addressed to listener(s); 2) self-directed (speaker to self); 3) neither (about external event/situation); 4) both/to listener and about situation \\
136 & Does this word typically license or expect a conventionalized response? & 1) requires a conventional response (e.g., greeting expects greeting); 2) often invites a response but not required; 3) no specific response expected \\
137 & Is this word culturally or dialectally restricted? & 1) widely cross-cultural/universal; 2) language- or culture-specific; 3) regional/dialect-specific; 4) age-group or subculture-specific \\
138 & Is this word etymologically or semantically onomatopoeic (sound-imitative)? & 1) yes — primarily onomatopoeic; 2) no — conventional lexical interjection; 3) mixed/partly onomatopoeic \\
139 & Does this word function as a discourse marker or organizer of conversation? & 1) yes — functions as backchannel/acknowledgement; 2) yes — organizes discourse (topic shift, floor management); 3) no — purely expressive/exclamatory \\
140 & Is this word conventionally specialized only as an interjection, or is it also used productively in other word classes? & 1) exclusive interjection (not used as other word classes); 2) also used as a lexical word (noun/verb/adjective) in common usage; 3) both uses are common depending on register \\
141 & What temporal orientation does this word primarily have? & 1) immediate reactive (response to event); 2) anticipatory/warning (signals expected event); 3) preparatory/attention-getting (prepares listener); 4) continuous/backchannel (sustains interaction) \\
142 & Which prosodic feature typically carries this word's meaning? & 1) rising pitch (question-like or expectant); 2) falling pitch (finality/exclamation); 3) level/neutral pitch; 4) short burst/abrupt; 5) prolonged/lengthened \\
143 & Is this word primarily used to express affirmation/acceptance or rejection/negation? & 1) affirmation/approval; 2) rejection/negation; 3) both (context-dependent); 4) neither (no acceptance/rejection function) \\
144 & In which modality is this word mainly used? & 1) primarily spoken; 2) commonly written (represented in orthography, e.g., in writing or comics); 3) frequently both spoken and written \\
145 & What grammatical person does this word encode? & 1) first; 2) second; 3) third \\
146 & What grammatical number does this word primarily express? & 1) singular; 2) plural; 3) dual \\
147 & What grammatical gender does this word encode? & 1) masculine; 2) feminine; 3) neuter; 4) common; 5) gender-neutral \\
148 & What animacy class does this word select? & 1) human; 2) animate non-human; 3) inanimate \\
149 & Which morphosyntactic case or role does this word form represent? & 1) nominative (subject); 2) accusative/object; 3) genitive/possessive; 4) dative/indirect object; 5) oblique/other \\
150 & Is this word a possessive determiner (attributive) or a possessive pronoun (standalone)? & 1) possessive determiner (attributive); 2) possessive pronoun (standalone) \\
151 & Can this word be used reflexively? & 1) reflexive only; 2) non-reflexive only; 3) both reflexive and non-reflexive \\
152 & For first-person plural reference, does this word distinguish inclusivity? & 1) inclusive (includes addressee); 2) exclusive (excludes addressee); 3) no inclusive/exclusive distinction \\
153 & What deictic distance does this word signal? & 1) proximal (near speaker); 2) medial (near addressee); 3) distal (far from speaker and addressee); 4) non-deictic \\
154 & Does this word encode a politeness or formality distinction? & 1) honorific/formal; 2) informal/colloquial; 3) politeness-neutral \\
155 & Is this word interpretable as a bound-variable under quantification or strictly referential? & 1) bound-variable (can be bound by quantifiers); 2) referential only; 3) both bound-variable and referential \\
156 & When plural, does this word impose a collective vs. distributive interpretation? & 1) collective; 2) distributive; 3) ambiguous/both \\
157 & Is this word used for emphasis or contrastive focus? & 1) emphatic/contrastive; 2) neutral/non-emphatic \\
158 & Does this word appear as a clitic (phonologically dependent) or a free independent word? & 1) clitic (phonologically attached); 2) independent (free word) \\
159 & In discourse, can this word be used anaphorically, cataphorically, or both? & 1) anaphoric (refers back); 2) cataphoric (refers forward); 3) both anaphoric and cataphoric \\
160 & Which typical semantic role(s) does this word most often denote in predicate argument structure? & 1) agent; 2) patient/theme; 3) experiencer; 4) possessor/beneficiary; 5) role-neutral \\
\end{longtable}
}
{\small\begin{longtable}{@{}>{\raggedleft\arraybackslash}p{0.06\textwidth}>{\raggedright\arraybackslash}p{0.88\textwidth}@{}}
\caption{All sampled templates. [MASK] is the position to be filled with words.}\label{tab:dataset-templates-any} \\ 
\toprule
\textbf{No.} & \textbf{Template} \\
\midrule
\endfirsthead
\caption[]{All sampled templates. [MASK] is the position to be filled with words (continued).} \\ 
\toprule
\textbf{No.} & \textbf{Template} \\
\midrule
\endhead
\midrule
\multicolumn{2}{r}{Continued on next page.} \\ 
\endfoot
\bottomrule
\endlastfoot
1 & During the morning meeting the team asked whether the project could [MASK] before the client presentation next Tuesday. \\
2 & The gardener noticed that the leaves on the oldest tree had turned a deep [MASK], signaling the change of season. \\
3 & When packing for the long trip she insisted that each suitcase contain at least one warm [MASK] for cold nights. \\
4 & He smiled and gave a brief [MASK] when the host introduced him to the guests at the dinner. \\
5 & If the software continues to [MASK] under heavy load we will need to optimize the database queries and caching. \\
6 & The little boy ran across the yard and shouted [MASK] as he chased after the colorful butterfly in sunlight. \\
7 & Her explanation included a clear [MASK] that allowed everyone to follow the complex steps without confusion or delay. \\
8 & You should always check the label to confirm whether the medication needs to be taken before or after [MASK]. \\
9 & The poet chose to [MASK] a single word repeatedly to build rhythm and emphasize the central emotion of the piece. \\
10 & As the storm approached the crew secured the equipment and moved the boats closer together to keep them [MASK]. \\
11 & I can't decide which flavor to pick because this menu lists too many tempting [MASK] that all sound delicious. \\
12 & To solve the puzzle she needed to align the colored tiles so that each row would [MASK] the pattern shown. \\
13 & The instructor explained that practicing a little every day would gradually [MASK] your ability to perform under pressure. \\
14 & At the finish line the runner collapsed and smiled despite the pain, proud of how far she had [MASK]. \\
15 & During the ceremony the speaker paused to acknowledge the generous donation made by a [MASK] community member. \\
16 & The recipe instructs you to mix the dry ingredients thoroughly until they are evenly [MASK] before adding wet components. \\
17 & Please send me a reminder if anything important is scheduled to [MASK] while I'm traveling for the conference. \\
18 & The teacher asked the students to fold the paper carefully so the airplane would [MASK] farther across the gym. \\
19 & We negotiated for hours and finally reached an agreement that seemed fair to every [MASK] involved in the negotiation. \\
20 & On cold winter mornings she liked to drink a cup of tea that tasted slightly [MASK] with a hint of cinnamon. \\
21 & Before submitting the report double check that every figure includes a clear [MASK] so readers can understand the data. \\
22 & The ancient map showed a hidden valley where the legend claimed a strange bird would always appear at [MASK]. \\
23 & If you want the plant to thrive place it in indirect light and water it only when the soil feels [MASK]. \\
24 & To reduce errors in the experiment the scientist repeated each measurement three times to ensure the results would [MASK]. \\
25 & She wore a bracelet engraved with a single word that reminded her daily to remain calm and [MASK] under pressure. \\
26 & When assembling the furniture follow the diagram and tighten each screw until the joint feels firmly [MASK] and secure. \\
27 & At the bakery early in the morning the smell of fresh bread seemed to [MASK] everyone passing by the open window. \\
28 & The conference schedule lists several panels, but the keynote address will [MASK] the discussion about sustainable urban planning. \\
29 & Our neighbors adopted a small dog that barks loudly whenever anyone approaches the [MASK] at the front gate. \\
30 & He adjusted the microscope’s focus until the tiny organism appeared clearly, revealing its [MASK] details and movement. \\
31 & The librarian stamped the due date and reminded patrons that overdue books may accrue fines starting the following [MASK]. \\
32 & Her grandmother shared a secret recipe that called for adding a pinch of spice to give the sauce a mild [MASK]. \\
33 & If the lights start to flicker during the storm you should unplug sensitive devices to protect them from [MASK]. \\
34 & To practice better time management schedule short breaks so you can recharge and return to work more [MASK]. \\
35 & The child carefully arranged the blocks in a tall tower then stepped back and admired how perfectly [MASK] it stood. \\
36 & When hiking in unfamiliar terrain always carry a map and compass because electronics might [MASK] without a clear signal. \\
37 & At the art gallery visitors studied the painting's texture and color to interpret the artist's intended [MASK]. \\
38 & She promised to return the borrowed umbrella before evening so that the owner would not [MASK] in the rain. \\
39 & The software update included security patches that aim to protect user data from unauthorized [MASK] attempts. \\
40 & He wrote a short note and slipped it into the book, hoping the next reader would find the kind [MASK] message. \\
41 & To improve flavor let the soup simmer slowly for an hour then taste and adjust the seasoning as [MASK]. \\
42 & The committee debated the proposal until midnight, trying to decide whether the new rule would [MASK] the current policy. \\
43 & During the interview she practiced answering questions clearly and avoided speaking too [MASK] so the panel could follow. \\
44 & The festival advertised live music, food stalls, and a parade where local bands would [MASK] through the crowded streets. \\
45 & Please label each container with the date and contents so you can easily identify what is [MASK] inside later. \\
46 & The mountain trail becomes narrow at times, so hikers should walk in single file and remain [MASK] to each other. \\
47 & He adjusted his calendar reminder to repeat weekly so he would never forget to [MASK] his subscription payment. \\
48 & At the science fair students displayed models and answered questions about how their experiments might [MASK] real world problems. \\
49 & The cat stretched lazily in the sunlight and blinked slowly as if to [MASK] gratitude for the warm spot. \\
50 & Before launching the product they conducted a survey to collect honest feedback and better understand user [MASK]. \\
51 & During the blackout neighbors helped one another and used candles to provide light until electricity could be [MASK]. \\
52 & She kept a small tokenizer for quick notes and liked to capture any sudden [MASK] that inspired creative writing. \\
53 & The repair manual advised that worn belts should be replaced promptly to prevent further damage to the [MASK]. \\
54 & On the final day of rehearsals the director asked each actor to perform the scene as if they were [MASK]. \\
55 & If the water tastes odd after running the tap let it run for a few minutes to clear any [MASK]. \\
56 & The professor encouraged students to ask questions and to challenge ideas respectfully in pursuit of deeper [MASK]. \\
57 & When storing photographs avoid exposure to direct sun and seal them in acid-free sleeves to preserve their [MASK]. \\
58 & Please sign the form at the bottom and return it to the office by Friday to ensure timely [MASK]. \\
59 & On warm evenings families gather in the park where children run freely and neighbors chat until it gets [MASK]. \\
60 & The architect sketched a simple layout showing where walls would be removed to create a more open [MASK] in the living area. \\
61 & Before making a purchase compare features and customer reviews so the product will meet your needs and feel [MASK]. \\
62 & She kept the old postcard because it carried a faded [MASK] from a friend who had traveled to distant places. \\
63 & When programming a loop remember to include a clear exit condition so the code does not [MASK] indefinitely. \\
64 & The festival organizer scheduled additional buses to ensure that attendees would reach the venue on time and [MASK]. \\
65 & Reading a map aloud helped them coordinate the route and agree on where they would [MASK] for lunch tomorrow. \\
66 & In the quiet library corner she found a comfortable chair where she could study and think more [MASK]. \\
67 & Please remember to water the plants twice a week and wipe dust from leaves to keep them looking [MASK]. \\
68 & The historian discovered a letter that provided a fresh [MASK] on daily life in the town a century ago. \\
69 & When the power returns please check the refrigerator first to ensure last night's food did not become [MASK]. \\
70 & She adjusted her camera settings to capture the action clearly, aiming to freeze motion and reduce [MASK] in photos. \\
71 & During the negotiation both sides agreed to discuss only core issues and to postpone any minor [MASK] until later. \\
72 & He tightened the rope slowly while the sailboat drifted, making sure each knot held firm and would not [MASK]. \\
73 & To lighten the mood at the meeting she told a short joke that made the audience [MASK] and relax. \\
74 & After studying the blueprint the builder confirmed the foundation depth and described how the walls would be [MASK]. \\
75 & The child learned to count aloud until she could recite numbers up to one hundred without missing a [MASK]. \\
76 & If the battery indicator remains low plug the charger in and allow the device to [MASK] for at least an hour. \\
77 & The travel guide recommended visiting the museum early in the morning to enjoy the exhibits while crowds were still [MASK]. \\
78 & Please fold the letters neatly and place them in the envelope so they can be delivered to the correct [MASK]. \\
79 & The teacher praised the student for offering a creative idea that showed both courage and thoughtful [MASK]. \\
80 & When cooking pasta taste a strand for doneness and adjust the cooking time until it feels slightly [MASK]. \\
81 & He stood at the train station and checked the schedule to confirm the exact time his next train would [MASK]. \\
82 & The guide warned hikers to stay on the marked path to avoid fragile plants and unexpected [MASK] in the area. \\
83 & She wrapped the fragile vase in soft paper and placed it in a box to protect it from accidental [MASK]. \\
84 & During the rehearsal each musician tuned carefully until their instruments produced a harmonious and pleasing [MASK]. \\
85 & The city planted new trees along the boulevard to create shade and improve the neighborhood's overall [MASK]. \\
86 & If you feel a sudden chill close the window and put on an extra layer so you stay warm and [MASK]. \\
87 & He used simple diagrams to explain the concept so that readers with no background would be able to [MASK]. \\
88 & The festival ended with a colorful fireworks display that lit the sky and filled the crowd with loud [MASK]. \\
89 & When presenting your idea practice a clear opening sentence to capture attention and state the main [MASK]. \\
90 & She replaced the worn shoelaces quickly and checked the fit to ensure her shoes felt secure during the [MASK]. \\
91 & The doctor advised drinking more water and getting adequate rest to support recovery and strengthen the body's [MASK]. \\
92 & On exam day arrive early, relax briefly, and read each question carefully to avoid missing key [MASK]. \\
93 & The volunteer organized donations into labeled boxes so staff could easily find clothing, food, and other [MASK] for distribution. \\
94 & She practiced the breathing technique until she could calm her mind and slow her heart rate in a few [MASK]. \\
95 & He programmed the thermostat to lower the temperature at night and automatically raise it again before [MASK]. \\
96 & The young gardener pruned the rose bushes gently, removing dead stems to encourage new growth and more abundant [MASK]. \\
97 & When planning the picnic check the weather forecast and choose a sheltered spot in case winds become unexpectedly [MASK]. \\
98 & To create a pleasant playlist include a variety of tempos so listeners experience both calm and more [MASK] songs. \\
99 & The coach reminded the team that consistent practice, discipline, and good sportsmanship would lead to long-term [MASK]. \\
100 & She kept a small notebook to jot down names, numbers, and any important [MASK] that came up during daily meetings. \\
101 & At the market he selected ripe produce and gently squeezed each fruit to check for softness and sweet [MASK]. \\
102 & The committee requested volunteers to assist with setup and asked that each person arrive ready to [MASK]. \\
103 & When watching the documentary take brief notes about dates and events to help remember the most relevant [MASK]. \\
104 & The bakery displayed a sign asking customers to form a single line and wait patiently for their [MASK]. \\
105 & She planned a simple evening menu and bought fresh ingredients so cooking dinner would be quick and [MASK]. \\
106 & If you notice a strange noise in the engine stop driving immediately and inspect the vehicle for any [MASK]. \\
107 & He practiced the scales every morning to improve finger agility and gain more confident [MASK] when performing music. \\
108 & During the storm the residents gathered in the community hall to share food, stories, and mutual [MASK]. \\
109 & She placed a fresh doormat by the door and wrote a kind note to welcome guests and set a [MASK]. \\
110 & The technician replaced the worn filter and advised checking it monthly to prevent buildup that can reduce [MASK]. \\
111 & To learn a language practice speaking with native speakers and repeat difficult phrases until pronunciation becomes more [MASK]. \\
112 & Before turning in your assignment confirm that all pages are attached and that your name appears clearly on the [MASK]. \\
113 & The cyclist slowed at the intersection and signaled clearly with a hand motion to indicate his next [MASK]. \\
114 & At the campground they stacked firewood neatly and stored extra supplies in a dry container to keep them [MASK]. \\
115 & He set a timer while baking so he would remove the cookies at the perfect moment when edges turned slightly [MASK]. \\
116 & The director cut the scene short because it didn't advance the plot and made the pacing feel unnecessarily [MASK]. \\
117 & When organizing receipts group them by date and category to simplify bookkeeping and make tax preparation less [MASK]. \\
118 & She drew a quick sketch of the room layout to plan furniture positions and test how each piece might [MASK]. \\
119 & The campers checked the map one last time to confirm their planned route and estimated the distance they would [MASK]. \\
120 & Please avoid using harsh chemicals on the countertop and instead clean with mild soap to maintain its natural [MASK]. \\
121 & After reviewing the proposal the board requested a concise summary highlighting risks, benefits, and the projected [MASK]. \\
122 & She placed a small mirror near the plant to reflect light and encourage growth toward the brighter [MASK]. \\
123 & The student organized flashcards by topic to help review vocabulary and ensure each term would be remembered more [MASK]. \\
124 & During the meeting he jotted down action items and assigned responsibility to ensure each task would be completed on [MASK]. \\
125 & To avoid confusion on the road always use your signals early and check mirrors to keep other drivers [MASK]. \\
126 & The local bakery donates leftover bread to shelters so members of the community can access fresh food [MASK]. \\
127 & She measured twice before cutting the fabric, understanding that careful preparation helps projects turn out both neat and [MASK]. \\
128 & If you encounter a locked gate leave it closed and note the location so someone can return later to [MASK]. \\
\end{longtable}
}

\end{document}